\documentclass[a4paper]{article}

\usepackage[american]{babel}
\usepackage{booktabs} 
\usepackage{amsmath, amsthm, amssymb, amsfonts}
\usepackage{mathrsfs}
\usepackage{graphicx}
\usepackage{tikz} 
\usepackage{algorithmic,algorithm2e}
\usepackage{array}
\usepackage{hyperref}
\newtheorem{ex}{Example}

\begin{document}

\title{SPOCC: Scalable POssibilistic Classifier Combination - toward robust aggregation of classifiers}

\author{Mahmoud~Albardan\textsuperscript{1} \and
John~Klein\textsuperscript{1} \and 
Olivier~Colot\textsuperscript{1}}

\date{\textsuperscript{1}Univ. Lille, CNRS, Centrale Lille, UMR 9189 - CRIStAL - Centre de Recherche en Informatique Signal et Automatique de Lille, F-59000 Lille \\
\{mahmoud.albardan,john.klein,olivier.colot\}@univ-lille.fr\\
}
\maketitle              
\begin{abstract}
We investigate a problem in which each member of a group of learners is trained separately to solve the same classification task. Each learner has access to a training dataset (possibly with overlap across learners) but each trained classifier can be evaluated on a validation dataset.

We propose a new approach to aggregate the learner predictions in the possibility theory framework. For each classifier prediction, we build a possibility distribution assessing how likely the classifier prediction is correct using frequentist probabilities estimated on the validation set. The possibility distributions are aggregated using an adaptive t-norm that can accommodate dependency and poor accuracy of the classifier predictions. We prove that the proposed approach possesses a number of desirable classifier combination robustness properties. Moreover, the method is agnostic on the base learners, scales well in the number of aggregated classifiers and is incremental as a new classifier can be appended to the ensemble by building upon previously computed parameters and structures. A python implementation can be downloaded at \href{https://github.com/john-klein/SPOCC}{this link}. 

\textbf{keywords}: robust classifier combination, agnostic aggregation, information fusion, classification, possibility theory
\end{abstract}

\section{Introduction}\label{sec:intro}
Classification is a supervised machine learning task consisting of assigning objects (inputs) to discrete categories (classes). When several predictors have been trained to solve the same classification task, a second level of algorithmic procedure is necessary to reconcile the classifier predictions and deliver a single one. Such a procedure is known as classifier combination, fusion or aggregation. When each individual classifier is trained using the same training algorithm (but under different circumstances) the aggregation procedure is referred to as an ensemble method. When each classifier may be generated by different training algorithms,  the aggregation procedure is referred to as a multiple classifier system. In both cases, the set of individual classifiers is called a classifier ensemble.

Classifier combination comes either from a choice of the programmer or is imposed by context. In the first case, combination is meant to increase classification performances by either increasing the learning capacity or mitigating overfitting. For instance, boosting \cite{schapire1990strength} and bagging \cite{breiman1996bagging} can be regarded as such approaches. 
In the second case, it is not possible to learn a single classifier. A typical situation of this kind occurs when 
the dataset is dispatched on several machines in a network and sequential learning (such as mini-batch gradient descent) is not possible either to preserve network load or for some privacy or intellectual property reasons. In this decentralized learning setting, 
a set of classifiers are trained locally and the ensemble is later aggregated by a meta-learner. 

In this article, we address classifier aggregation in a perspective that is in line with the decentralized setting assuming that the meta-learner has access to evaluations of the individual classifiers on a fraction of each local dataset which is not used for training. 
We do not make any assumption on the base learner models and we do not assume base learners are trained on i.i.d. samples, however it is assumed that the union of the fractions of (local) datasets used by the meta-learner is i.i.d.. 
We introduce a number of desirable robustness properties for the aggregation procedure in this context. We investigate fault tolerance (ability to discard classifiers whose predictions are noise), robustness to adversarial classifiers (ability to thwart classifiers with abnormal error rates) and robustness to redundant information (when classifier predictions are highly dependent).

We introduce an aggregation procedure in the framework of possibility theory. We prove that these robustness properties are verified asymptotically (when the size of the validation set is large) for this new approach. The mechanism governing the aggregation essentially relies on estimates of probabilities of class labels, classifier predictions or classifier correct predictions. There are many related works \cite{huang1995method,kim2012bayesian,lacoste2014agnostic} dealing with classifier combination using similar information. 
We believe we are the first to do so in the framework of possibility theory but more importantly these above referenced work are not proved to possess theoretical robustness guarantees. 
An asymptotic optimality property is verified by an approach from \cite{balakrishnan2015simple}. This property is stronger than most of the properties that we state except for robustness to classifier dependency. Also, two technical conditions are necessary for the property to hold while our results have no such conditions to check. Similar remarks hold w.r.t. \cite{biau2016cobra} which shares some ideas with \cite{balakrishnan2015simple}. Another piece of work with strong properties (oracle inequalities) is exponential weight aggregation \cite{rigollet2012sparse} but the properties are non-exact\footnote{Error rate is proved to convergence to the vicinity of the optimal one not exactly to the optimal one.} and hold in expectation or with high probability while our properties rely on almost sure convergence. Also, exponential weight aggregation is a linear combination model while our method is non-linear.

In addition, the form of aggregation robustness achieved by our method does not jeopardize other important aspects of aggregation such as scalability and incrementalism, two aspects that \cite{balakrishnan2015simple} fails to possess. 
Our approach relies on a parametric model involving a number of parameters that is linear in the number of classifiers. 
Incrementalism is also preserved in the sense that a new classifier can be appended to the ensemble without implying to re-estimate the previously obtained parameter values or structures.


In the next section, we recall the classifier aggregation problem and formally define the robustness properties that we seek. In section \ref{sec:robust_combination_in_the_possibilistic_framework}, we introduce a new aggregation technique in the framework of possibility theory and we show that the desired properties hold asymptotically for this technique. Section \ref{sec:experimental_results} contains numerical experiments illustrating our results.

\section{Problem statement}

\subsection{Classification} 
\label{sub:classification}


Let $\Omega$ denote a set of $\ell$ class labels $\Omega$ = \{$\omega_1,\hdots,\omega_\ell$\}. 
    Let $\mathbf{x}$ denote an input example with $d$ entries. 
Most of the time, $\mathbf{x}$ is a vector and lives in $\mathbb{R}^d$ but sometimes some of its entries are categorical data and $\mathbf{x}$ lives in an abstract space which does not necessarily have a vector space structure. 
Without loss of generality, we suppose that $\mathbf{x}$ is a vector in the rest of this article.
    
A classification task consists in determining a prediction function $c$ that maps any input $\mathbf{x}$ to its actual class $y$ $\in$ $\Omega$. 
This function is obtained from a training set $\mathcal{D}_{\textrm{train}}$ which contains pairs $\left(   \mathbf{x}^{(i)},y^{(i)}  \right)$ where $y^{(i)}$ is the class label of example $\mathbf{x}^{(i)}$.
Given $K$ classifiers (each of them trained by one base learner), the label $y$ assigned by the $k^{\text{th}}$ classifier to the input $\mathbf{x}$ is denoted by $c_k(\text{\textbf{x}})$.

From a statistical point of view, training examples are instances of a random vector $X$ whose distribution is unknown. Likewise, class labels are instances of a random variable $Y$ whose distribution is also unknown. The training set is often alleged to contain i.i.d. samples of the joint distribution of $\left( X,Y \right) $.



\subsection{Classifier performance estimates} 
\label{sub:classifier_performance_estimates}

The ultimate goal of machine learning is to obtain predictors that generalize well (w.r.t. unseen data at training time). Mathematically speaking, this means achieving the lowest possible expected loss between predictions and true values. When misclassification errors do not have different costs, the 0-1 loss function $L$ is the standard choice:
$$L \left( y, c_k \left( \mathbf{x} \right)  \right) = \begin{cases} 0 & \text{if } y=c_k \left( \mathbf{x} \right) \\ 1 & \text{otherwise} \end{cases}. $$
In this case, the expected loss is the misclassification error rate of $c_k$. It is well known, that the error rate minimizer is the Bayes classifier $c_{\star}$:
$$ c_{\star} \left( \mathbf{x} \right) = \underset{y \in \Omega}{\arg\max} \; p \left( Y=y | X=\mathbf{x} \right). $$
Obviously, since the conditional distributions of $Y$ given $X=\mathbf{x}$ are unknown, we must try to find proxys of the Bayes classifier. The error rate (or risk) of classifier $c_k$ is denoted by $r \left[ c_k \right] $.

Although our goal is to achieve the lowest possible error rate, the performances of a classifier are not, in general, constant across true class labels and predicted ones. This finer grained information will be instrumental to elicit our possibilistic ensemble of classifiers. This information is contained in the confusion matrix $\mathbf{M}^{(k)}$. Each entry of this matrix reads 
\begin{equation}
  M^{(k)}_{i,j} = \sum\limits_{ \left( \mathbf{x},y \right) \in \mathcal{D}_{\text{val}} } \mathbb{I} \left\{ y=\omega_i \right\}  \mathbb{I} \left\{ c_k \left( \mathbf{x} \right)  = \omega_j \right\}, \label{eq:mat_conf}
\end{equation}
where $\mathbb{I}$ denotes the indicator function. It is important to compute the confusion matrices using a validation set $\mathcal{D}_{\text{val}}$ disjoint from $\mathcal{D}_{\text{train}}$ otherwise the estimates drawn from the matrix are biased. Actually, if $n_{\text{val}}$ is the size of the validation set, then $\frac{M^{(k)}_{ij}}{n_{\text{val}}}$ is the maximum likelihood estimate of the joint probability $p \left( Y=\omega_i, c_k \left( X \right) = \omega_j \right) $. Also, the sum of the non-diagonal entries of $\mathbf{M}^{(k)}$ over $n_{\text{val}}$ is an unbiased estimate of the error rate of $c_k$. Many other performance criterion estimates can be derived from a confusion matrix.

The classifier combination that we introduce in section \ref{sec:robust_combination_in_the_possibilistic_framework} essentially relies on the information contained in those matrices. Computing those matrices can thus be regarded as the training phase of the combination method. 


\subsection{Agnostic combination of classifiers and position of the problem} 
\label{sub:combining_classifiers}
Let $C$ denote the random vector spanned by plugging $X$ into the ensemble of classifiers: 
$$C = \begin{pmatrix} c_1 \left( X \right) \\ \vdots \\ c_K \left( X \right)    \end{pmatrix}.$$

A realization of this random vector is denoted by $\mathbf{c}$ or $\mathbf{c}\left( \mathbf{x} \right) $ whenever the dependence on inputs must be made explicit. We place ourselves in the context where vectors $\mathbf{c}$ can be pictured as new (learned) representation of inputs and we must be agnostic, i.e. we have no control on the base classifiers. In this context, the best aggregate classifier \cite{balakrishnan2015simple} based on observed data is thus 
\begin{equation}
  c_{\ast} \left( \mathbf{x} \right) = \underset{y \in \Omega}{\arg\max} \; p \left( Y=y | C= \mathbf{c} \left(  \mathbf{x}  \right)  \right). \label{eq:optim_agg}
\end{equation}
Again, the distributions involved in the above definition are unknown. Since $\mathbf{c}$ lives in the discrete space $\Omega^{K}$, it is possible to try to learn these distributions but this leads to very hard inference problems \cite{kim2012bayesian,balakrishnan2015simple,li2019exploiting} and such statistical learning approaches do not scale well w.r.t. either $\ell$ or $K$. 
The smallest memory complexity among these references is achieved by \cite{li2019exploiting} who introduce a mixture model relying on tensor decomposition. If $K'$ denotes the number of components in the decomposition, the number of parameters to learn is linear in $K'\times K$. Linearity in $K$ can thus be claimed if $K' \ll K$ which cannot always be assumed.
In addition, the Bayesian solutions introduced in these references do not allow to obtain an incremental\footnote{Incremental aggregation means that a new classifier can be appended to the ensemble later without having to recompute everything from scratch. } aggregation algorithm, an attribute that we believe is much desirable.

Generally speaking, classifier combination consists in finding a function $f$ capturing the relation between vectors $\mathbf{c}$ and class labels $y$ that achieves the closest possible performances as compared to $c_\ast$. In the approach presented in the next section, we leverage the flexibility of possibility theory to find one such function. The proposed possibilistic approach visits several functions, i.e. aggregation strategies, and select the one maximizing accuracy obtained on the validation dataset. The strategies in question are generalizations of logical rules as opposed to probabilistic approaches which resort to calculus rules. In this regard, the proposed solution both relies on artificial learning and artificial reasoning principles.

Besides, we have chosen to place ourselves in a framework in which each classifier can only deliver one piece of information, i.e. a predicted class label. 
This allows us to be completely agnostic on the nature of the base learners. 
Indeed, depending on the training algorithm and model employed by a learner, this latter may be able to deliver a score vector (usually in the form of a probability distribution). 
In this case, the above analysis no longer applies and the optimal aggregation consists in inferring posterior predictive probabilities of class labels given the scores. 
Examples of approaches in this line of work are reviewed in \cite{tulyakov2008review}. 
It is possible to remain relatively agnostic on the nature of base classifiers by resorting to a statistical calibration step that allows to obtain prediction probabilities from non-probabilistic classifiers as done in \cite{bella2013effect}. 
Calibration will consume a significant portion of the validation set leaving a smaller amount to train the aggregation technique. 
Consequently, score based aggregation is out of the scope of this paper. 
Actually, score based aggregation is a leading follow-up of the approach introduced in the next section as mentioned in the concluding remarks in section \ref{sec:conclusion}. 



\subsection{Desirable properties for classifier combination} 
\label{sub:desirable_properties}

In terms of purely error rate related performances, the most desirable property for some aggregation function $f$ is
\begin{equation}
  r \left[    f \left( \mathbf{c} \right)\right] \longrightarrow r \left[    c_\ast\right] \text{ as } n_{\text{val}} \longrightarrow \infty. \label{eq:oracle}
\end{equation}
The aggregation technique studied in \cite{balakrishnan2015simple} achieves a result of this kind (under two technical assumptions). Indeed this technique, which elaborates on \cite{huang1995method}, amounts to compute maximum likelihood estimates of the probabilities involved in \eqref{eq:optim_agg}. 
But classifier aggregation can also bring other types of guarantees which we refer to as robustness. Robustness is understood here as a form of fault tolerance, i.e. the ability to maintain a good level of predictions in several circumstances involving malfunctioning individual classifiers. There may be different causes behind malfunctioning classifiers, e.g. hardware failure or malicious hacks. 

Among other possibilities, we have identified the following desirable properties in this scope:
\begin{itemize}
   \item[(a)] robustness to random guess: if the error rate of $c_k$ is $\frac{\ell-1}{\ell}$ then $f \left( \mathbf{c} \right) = f \left( \mathbf{c}_{-k} \right)  $ where $\mathbf{c}_{-k} \in \Omega^{K-1}$ is the same vector as $\mathbf{c}$ but with its $k^{\text{th}}$ entry deleted.

   Property (a) means that if the predictions of $c_k$ are in average no better than random guess then $c_k$ has no influence on the aggregated classifier.

   \item [(b)] robustness to adversarial classifiers: if $c_k$ has an error rate larger than random guess, i.e. $r \left[ f \left( \mathbf{c} \right) \right] > \frac{\ell-1}{\ell} $, then there is a classifier $c^{(\text{rec})}_k$ with an error rate lower than random guess such that $ f\left( \mathbf{c} \right) = f \left( \tilde{\mathbf{c}} \right) $ where $\tilde{c}_s = c_s$ for any $s\neq k$ and $\tilde{c}_k = c^{(\text{rec})}_k$.


   Property (b) means that we can somewhat rectify the incorrect predictions of classifier $c_k$ so that the aggregated classifier is identical to the one obtained from a non-adversarial situation.

   \item [(c)] robustness to redundant information: if there are two individual classifiers such that $c_k \left( \mathbf{x} \right) = c_{k'} \left( \mathbf{x} \right)  $ for any $\mathbf{x}$, then $f \left( \mathbf{c} \right) = f \left( \mathbf{c}_{-k} \right)  $.

   Property (c) means that copies of classifiers have no influence on the aggregated classifier.


 \end{itemize} 

In the above, we assume that the aggregation function $f$ is produced by a given algorithmic procedure and that this procedure applies for any $K>1$. So $f \left( \mathbf{c}_{-k} \right) $ is not the restriction of $f \left( \mathbf{c} \right) $ but another function learned from the same algorithm by omitting classifier $c_k$.

Obviously, one can think of other properties or reshape them in different ways. For instance, a soft version of property (c) would be better in the sense that an ensemble contains rarely identical copies of a predictor but it contains very often highly dependent ones. This a first attempt to formalize desirable robustness properties for classifier combination and we hope that more advanced declinations of these will be proposed in the future.

For the time being, our goal in this paper is to introduce an aggregation procedure that is compliant with properties (a) to (c). We will prove that these properties hold for the possibilistic approach that we introduce in the next section at least asymptotically for some of them. Observe that \eqref{eq:oracle} asymptotically implies properties (a) to (c) so the added value of our approach as compared to \cite{balakrishnan2015simple} relies on its incremental aspect as shown in \ref{sub:online_and_incremental_aggregation} and scalability w.r.t. $K$ as numerical experiments will illustrate in section \ref{sec:experimental_results}.


\subsection{Other related works} 
\label{sub:other_related_works}

So far, we have mentioned only closely related works which perfectly fall in the same setting as ours, i.e. performing the same type of aggregation based on the same information. 
We remind that this paper is focused on a classifier aggregation paradigm in which one must be agnostic on the base learners. 
As explained before this immediately rules out a large number of methods such as those relying on classifier scores and ensemble methods. 
Score based combination most often assume that base learners exploit a given training algorithm. 
For instance, \cite{loustau2008aggregation} introduced an aggregation method tailored to SVMs while \cite{guo2019ifusion} applied another one to combine deep nets. By definition, so do ensemble methods such as  \cite{liu2019advancing}. Other score based algorithms will require that training algorithms belong to the same class of models, typically probabilistic learners as in \cite{hoang2019collective}.

It must also be made clear that the addressed paradigm in this paper is not federated learning. In federated learning, there is no base learners. A group of remotely connected clients have access to a local dataset. Clients are meant to compute a parameter update based on their local data and send this update to a meta-learner (\cite{yao2019federated,li2019rsa,ijcai2019-363}) to collaboratively train a model. This means that parameter updates are aggregated and this is thus not a classifier aggregation problem.

Finally, an utmost important and original aspect of the method introduced in this paper is its robustness properties w.r.t. noise, adversaries and information redundancy. To the best of our knowledge, there is no prior art in classifier aggregation that has touched jointly these aspects for agnostic aggregation of predicted class labels. 
Robustness to noise is investigated by \cite{NIPS2019_8728} but again in a very different setting which is deep fusion, i.e. deep learning from multiple inputs. 
Other references in the literature are focused on other aspects than robustness such as security for instance (\cite{ma2019secure}). 
However, robustness is a hot topic in the supervised learning paradigm with important consequences in deep learning \cite{madry2018}. But obtaining robust base learners does not ensure that the aggregation itself is robust. 



\section{Robust combination in the possibilistic framework} 
\label{sec:robust_combination_in_the_possibilistic_framework}

In this section, we introduce a new classifier combination approach in the possibility theory framework. Possibility theory \cite{Zad78,dubois1988possibility} is an uncertainty representation framework. 
It has strong connections with belief functions \cite{Dub82,de1982interpretation}, random sets \cite{goodman1982fuzzy,pei1982treating,sales1982fuzzy} , imprecise probabilities \cite{dubois1992upper,de1999supremum} or propositional logic \cite{benferhat1999possibilistic}. For a concise but thorough overview of possibility theory, the reader is referred to \cite{dubois2015possibility} but \ref{sec:an_overview_of_possibility_theory} already provides deeper insights into this framework and as to why it is particularly relevant for classifier aggregation tasks. 

Possibility theory is a widely used framework in symbolic artificial intelligence. It allowed the derivation of new propositional and/or modal logics in which the level of uncertainty of logical propositions can be assessed (\cite{prade1991possibilistic,dubois2017generalized}). This has applications in logic programming (\cite{alsinet2008logic}), automated reasoning (\cite{dubois1994automated}) or expert systems (\cite{shenoy1992using}). A popular class of non-probabilistic graphical models relying on ordinal condition functions can also be revisited as a possibilistic model, see \cite{amor2018possibilistic} for recent developments in this field. It has also been used in other branches of artificial intelligence such as information fusion (\cite{destercke2008possibilistic}) or machine learning (\cite{serrurier2015entropy,hullermeier2002possibilistic}).

In this paper, we adopt a knowledge based system view of this theory. In this regard classifier predictions are expert knowledge to which a degree of belief is attached in the form of possibility distributions. Following a normative approach, experts are reconciled by designing a conjunctive rule that must obey the desirable properties presented in the previous section.

\subsection{Possibility theory basics} 
\label{sub:possibility_theory_basics}

A possibility distribution $\pi$ maps each element of $\Omega$ to the unit interval $\left[ 0;1 \right] $ whose extreme values correspond respectively to impossible and totally possible epistemic states. If $\pi \left(   y  \right)=1$ then this class label is totally possible (meaning that we have no evidence against $y$) . If $\pi \left(   y  \right)=0$ then $y$ is ruled out as a possible class label.

Given a subset \textit{A} of $\Omega$, a possibility measure $\Pi$ is given by:
  \begin{equation}
  \Pi(A) = \underset{y\in A}{\max} \;  \pi(y)
  \end{equation} 
  which means that the possibility of a subset \textit{A} is equal to the maximum possibility degree in this subset. A possibility measure is thus maxitive: $\Pi(A\cup B) =\max(\Pi(A),\Pi(B))$ as opposed to probability measures which are summative. Observe that this property accounts for the fact that the possibility distribution is enough information to compute the possibility measure of any subset. 


\subsection{From classifier confusion matrices to possibility distributions} 
\label{sub:from_classifier_confusion_matrices_to_possibility_distributions}

If one normalizes the $j^{\text{th}}$ column of the confusion matrix $\mathbf{M}^{(k)}$, then we obtain an estimate of the probability distribution $p \left( Y = \omega_i | c_k = \omega_j \right) $. So, if the $k^{\text{th}}$ classifier predicts $\omega_j$ for some input $\mathbf{x}$, we can adopt these frequentist probabilities as our beliefs on the class label of $\mathbf{x}$. But unless, unrealistic conditional independence assumptions\footnote{These assumptions and the corresponding probabilistic approach are described in section \ref{sec:experimental_results}.} are made, probabilistic calculus rules will not easily allow to combine beliefs arising from several classifiers. 

As an alternative to this approach, we propose to build a possibility distribution from $p \left( Y = \omega_i | c_k = \omega_j \right) $ as information fusion in the possibilistic framework can mitigate dependency issues and does not lead to intractable computations. To cast the problem in the possibilistic framework, we use Dubois and Prade transform (DPT) \cite{Dub82}. 
For some arbitrary probability distribution $p$ on $\Omega$, let $\mathsf{per}$ denote a permutation on $\Omega$ such that probability masses of $p$ are sorted in descending order, i.e. if $p'=p\circ \mathsf{per}$ then $ p'\left( \omega_1 \right) \geq p'\left(  \omega_2 \right) \geq \hdots \geq p'\left(\omega_\ell \right) $.
The (unique) possibility distribution $\pi$ arising from $p$  through DPT is given by
  \begin{align}
    \pi \left( \mathsf{per} \left( \omega_i \right)   \right)  =\begin{cases}
      1 & \text{if } i=1 \\
      \pi\left(  \mathsf{per} \left( \omega_{i-1} \right)   \right)  & \text{if } i >1  \text{ and } p' \left( \omega_i \right)  = p'\left( \omega_{i-1} \right)  \\
      \sum\limits_{q=i}^\ell  p' \left( \omega_q \right) & \text{otherwise}
    \end{cases}.\label{eq:transfo_possib}
  \end{align}
  
If $p_{Y|c_k=\omega_j}$ denotes the distribution of class labels when the $k^{\text{th}}$ classifier predicts $\omega_j$, the corresponding possibility distribution is denoted by $\pi_{k|j} = \text{DPT} \left\{ p_{Y|c_k=\omega_j} \right\}$. For each input $\mathbf{x}$, the $K$ classifier predictions are turned into $K$ expert opinions in the form of possibility distributions $\left( \pi_{k| \mathsf{ind} \left(    c_k \left( \mathbf{x} \right) \right)  } \right)_{k=1}^K $ where $\mathsf{ind} \left( \omega_j \right)=j $.


\subsection{Aggregation of possibility distributions} 
\label{sub:aggregation_of_possibility_distributions}

Formally speaking, any $K$-ary operator on the set of possibility distributions is an admissible combination operator. 
Triangular norms, or t-norms, are instrumental to yield well defined aggregation operators for possibility distributions. A t-norm  $\mathcal{T}:  \left[ 0;1 \right]^2 \longrightarrow [0,1] $ is a commutative and associative mapping therefore it is easy to build a $K$-ary version of it using successive pairwise operations: 

$$ \mathcal{T} \left( a_1,\hdots, a_K \right) = \mathcal{T} \left( a_1 , \mathcal{T } \left( \hdots, \mathcal{T} \left( a_{K-1},a_K \right)  \right)  \right),$$
for any $\left( a_k \right)_{k=1}^K \in \left[ 0;1 \right]^K  $.

Moreover, a t-norm has 1 as neutral element, 0 as absorbing element and possesses the monotonicity property which reads: for any $a,b,c,d \in \left[ 0;1 \right] $ such that $a\leq c$ and $b\leq d$, then $ \mathcal{T} \left(   a,b \right) \leq \mathcal{T} \left(   c,d \right) $. Finally, a t-norm is upper bounded by the minimum of its operands. 

To combine possibility distributions using a t-norm, we can simply apply a t-norm elementwise. For instance, if $\pi_{kk'}$ is the aggregated possibility distribution obtained by applying a t-norm to distributions $\pi_k$ and $\pi_{k'}$, then 

$$\pi_{kk'} \left( y \right) = \mathcal{T} \left(  \pi_k \left( y \right) , \pi_{k'}\left( y \right)   \right), \forall y \in \Omega. $$

We will use the same t-norm symbol to stand for the overall combination of possibility distributions and we will write $\pi_{kk'} = \mathcal{T} \left(  \pi_k , \pi_{k'} \right)$. Examples of t-norms are elementwise multiplication $\mathcal{T}_{\times}$ and elementwise minimum $\mathcal{T}_{\wedge}$.

Decision making based on maximum expected utility is also justified using non-additive measures (capacities) \cite{GILBOA198765} such as possibility measures. Consequently, the possibilistic aggregated classifier, denoted $c_{\text{ens}}$, is given by 
\begin{align}
  c_{\text{ens}} \left( \mathbf{x} \right) &= \underset{y \in \Omega}{\arg\max} \; \pi_{\text{ens}} \left( y \right), \\
  \text{with } \pi_{\text{ens}}  &= \mathcal{T} \left(  \pi_{1|\mathsf{ind} \left(c_1 \left( \mathbf{x} \right)\right) }, \hdots,  \pi_{K|\mathsf{ind} \left(c_K \left(  \mathbf{x} \right) \right) } \right).
\end{align}

Algorithm \ref{spocc-train} explains what computations should be anticipated as part of a training phase and Algorithm \ref{spocc-test} summarizes how an input $\mathbf{x}$ class label is predicted at test time. The procedure corresponding to these algorithms is referred to as Scalable POssibilistic Classifier Combination (\textbf{SPOCC}). Note that there may be several class labels maximizing $\pi_{\text{ens}}$ therefore the aggregated classifier prediction $c_{\text{ens}} \left( \mathbf{x} \right)$ may be set-valued. Working with set-valued predictions is out of the scope of this paper and will be considered in future works. In the advent of a class label tie, and for any probabilistic, possibilistic or deterministic aggregation approach, one of these labels is chosen at random.

    \begin{algorithm}
    \DontPrintSemicolon
     \caption{SPOCC - training phase}
     \KwData{validation set $\mathcal{D}_{\text{val}}$, classifiers $\left( c_k \right)_{k=1}^K$, number of class labels $\ell$.}

     \For{$k\in \left\{ 1,\hdots, K \right\}$}{
     Compute confusion matrix $\mathbf{M}^{(k)}$ as in \eqref{eq:mat_conf}.

      \For{$j \in \Omega$}{

      Compute conditional probability estimates 
        $$ \hat{p}_{\text{mle}} \left( Y = \omega_i | c_k = \omega_j \right) \leftarrow \frac{M^{(k)}_{ij}}{\sum\limits_{i'}M^{(k)}_{i'j} }, \forall i \in \Omega.  $$

      Compute possibility distribution using \eqref{eq:transfo_possib}
      $$\pi_{k|j}\leftarrow \text{DPT} \left\{   \hat{p}_{\text{mle}} \left( Y = \cdotp | c_k = \omega_j \right)\right\}.$$ 
      }
     }
     
     Return possibility distributions $\left( \pi_{k|j} \right)_{\substack{ 1 \leq k \leq K \\ 1 \leq j \leq \ell}}$.
     \label{spocc-train}
     \end{algorithm}

    \begin{algorithm}
    \DontPrintSemicolon
     \caption{SPOCC - test phase}
     \KwData{input $\mathbf{x}$, classifiers $\left( c_k \right)_{k=1}^K$, possibility distributions $\left( \pi_{k|j} \right)_{\substack{ 1 \leq k \leq K \\ 1 \leq j \leq \ell}}$ and t-norm $\mathcal{T}$.}

     \For{$k\in \left\{ 1,\hdots, K \right\}$}{
      Compute individual classifier prediction $j_k \leftarrow \mathsf{ind} \left(    c_k \left( \mathbf{x} \right)\right) $.
     }

     Compute  $\pi_{\text{ens}}  \leftarrow \mathcal{T} \left(  \pi_{1|j_1} , \hdots,  \pi_{K|j_K} \right)$.

     Return $c_{\text{ens}} \left( \mathbf{x} \right) \leftarrow \underset{y \in \Omega}{\arg\max} \; \pi_{\text{ens}} \left( y \right)$.
     
     \label{spocc-test}
     \end{algorithm}     

\subsection{Adaptive aggregation w.r.t. dependency} 
\label{sub:adaptive_aggregation_w_r_t_dependency}

The predictions of an ensemble of individual classifiers are usually significantly dependent because they are trained to capture the same bound between inputs and class labels. So if classifiers are at least weak classifiers, they will often produce identical predictions. More importantly, from an information fusion standpoint, if a majority of the classifiers are highly dependent and have a larger error rate than the remaining ones, they are likely to guide the ensemble toward their level of performances making classifier fusion counter-productive.

In the approach introduced in this paper, it is possible to mitigate dependency negative impact by choosing an idempotent t-norm such as elementwise minimum $\mathcal{T}_\wedge$. Indeed, in the worst dependency case, classifier $c_{k}$ is a copy of classifier $c_{k'}$ therefore they have an unjustified weight in the ensemble predictions. But if two individual classifiers are identical they will also yield identical possibility distributions and if these latter are combined using $\mathcal{T}_\wedge$, then these two classifiers will be counted as one. This is exactly the spirit of property (c).

Two difficulties arise from this quest for robustness w.r.t. classifier redundancy:
\begin{itemize}
  \item[(i)] It is not recommended to systematically use an idempotent combination mechanism because it is also possible that two poorly dependent classifiers yield identical possibility distributions in which case it appears justified that their common prediction impacts on the ensemble aggregated decision.
  \item[(ii)] There are different levels of dependency among subsets of individual classifiers therefore, using a single t-norm to jointly aggregate them is not the best option.
\end{itemize}

To address the first issue, we propose to use the following parametric family $\left( \mathcal{T}_{\lambda} \right)_{\lambda \in [1;+\infty)} $ of t-norms:
\begin{equation}
  \mathcal{T}_{\lambda} \left( a_1,a_2 \right)  =  e^{-(|\log a_1|^\lambda + |\log a_2|^\lambda)^{\frac{1}{\lambda}}}, \forall a_1,a_2 \in \left[ 0;1 \right] .
\end{equation}

This family is known as Aczel-Alsina t-norms and is such that $\mathcal{T}_1 = \mathcal{T}_{\times}$ and $\mathcal{T}_\infty = \mathcal{T}_{\wedge}$. We can thus tune $\lambda$ all the higher as the level of dependence between classifiers is high.

To assess the dependence level $\lambda$ among two classifiers $c_k$  and $c_{k'}$, we use the following definition
\begin{equation}
  \kappa \left( c_k,c_{k'}\right) = 1 - \exp \left( - \frac{1}{n_{\text{val}}} \left\lvert \log \left( \frac{ \mathcal{L}_0 }{\mathcal{L}_1  } \right) \right\rvert \right), \label{eq:lambda}
\end{equation}
where $\frac{ \mathcal{L}_0 }{\mathcal{L}_1  }$ is the likelihood ratio of the independence model over the joint model. These likelihoods are given by 
\begin{align}
 \mathcal{L}_0  &= \prod_{i=1}^{n_{\text{val}}} \hat{p}_{\text{mle}} \left( c_k \left( \mathbf{x}^{(i)} \right)  \right) \hat{p}_{\text{mle}} \left( c_{k'} \left( \mathbf{x}^{(i)} \right)  \right),\\
 \text{and } \mathcal{L}_1  &= \prod_{i=1}^{n_{\text{val}}} \hat{p}_{\text{mle}} \left( c_k \left( \mathbf{x}^{(i)} \right)  , c_{k'} \left( \mathbf{x}^{(i)} \right)  \right).
 \end{align}


These likelihoods are computed using all training examples contained in the validation set $\mathcal{D}_{\text{val}}$. The probabilities involved in the computation of $\mathcal{L}_0$ are the maximum likelihood estimates of the parameters of the multinomial marginal distributions $p \left( c_k \left( X \right)  \right) $ and $p \left( c_{k'} \left( X \right)  \right) $ respectively. The probabilities involved in the computation of $\mathcal{L}_1$ are the maximum likelihood estimates of the parameters of the multinomial joint distribution $p \left( c_k \left( X \right), c_{k'} \left( X \right) \right) $.

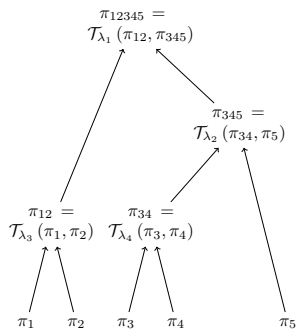
\begin{figure}[h]
\begin{center}
\resizebox{.4\textwidth}{!}{
\begin{tikzpicture}
\node[text width=1.8cm, align=center]  (v1) at (1,0) {$\pi_1$};
\node[text width=1.9cm, align=center]  (v2) at (2,0) {$\pi_2$};
\node[text width=1.8cm, align=center]  (v3) at (3,0) {$\pi_3$};
\node[text width=2.0cm, align=center]  (v4) at (4,0) {$\pi_4$};
\node[text width=1.8cm, align=center]  (v5) at (6.25,0) {$\pi_5$};
\node[text width=1.8cm, align=center]  (v6) at (1.5,2) {$\pi_{12}=\mathcal{T}_{\lambda_{3}} \left( \pi_1,\pi_2 \right) $};
\node[text width=1.8cm, align=center]  (v7) at (3.5,2) {$\pi_{34}=\mathcal{T}_{\lambda_{4}} \left( \pi_3,\pi_4 \right) $};
\node[text width=1.8cm, align=center]  (v8) at (5.25,4) {$\pi_{345}=\mathcal{T}_{\lambda_{2}} \left( \pi_{34},\pi_5 \right) $};
\node[text width=1.8cm, align=center]  (v9) at (3.125,6) {$\pi_{12345}=\mathcal{T}_{\lambda_{1}} \left( \pi_{12},\pi_{345} \right) $};

\draw  (v1) edge[->] (v6);
\draw  (v2) edge[->] (v6);
\draw  (v3) edge[->] (v7);
\draw  (v4) edge[->] (v7);
\draw  (v5) edge[->] (v8);
\draw  (v7) edge[->] (v8);
\draw  (v8) edge[->] (v9);
\draw  (v6) edge[->] (v9);
\end{tikzpicture}}
\caption{Example of a dendrogram for $K=5$. Leaf nodes are at the bottom. For each of the four non-leaf nodes, a specific dependence level $\lambda_{a}$ ($a\in \left\{ 1;2;3;4 \right\}$) must be determined to compute the aggregated possibility distribution.\label{fig:dendro}}
\end{center}
\end{figure}

The definition of the dependence level $\kappa$ can be extended to more than two classifiers but this will turn out to be unnecessary because we will use hierarchical agglomerative clustering \cite{ward1963hierarchical} (HAC) to address issue (ii). HAC will produce a dendrogram $\mathcal{G}$, i.e. a t-norm computation binary tree. Each leaf in this tree is in bijective correspondence with one of the possibility distributions $\pi_k$ induced by a classifier. There are thus $K$ leafs in $\mathcal{G}$. Furthermore, each non-leaf node in the tree stands for a t-norm operation involving two operands only. Consequently, each non-leaf node has exactly two children and there $K-1$ such nodes, one of them being the root node. Figure \ref{fig:dendro} gives an illustrative example of a dependence dendrogram allowing to compute the aggregated possibility distribution.

HAC relies on a classifier dissimilarity matrix $\mathbf{D}$. In our case, entries of this $K \times K$ matrix are simply given by $D_{kk'} = 1-\kappa \left( c_k,c_{k'}\right)$. 

The t-norm based possibility distribution aggregation method described in the above paragraphs is meant to replace the penultimate step of Algorithm \ref{spocc-test} but most of the computations pertaining to this dependency adaptive aggregation can be done at training time. Indeed, the computation of the pairwise dependence levels and the dendrogram do not depend on the unseen example $\mathbf{x}$ that we will try to classify at test time. For a minimal test phase computation time, we need to assign to each non-leaf node $V_a$ of $\mathcal{G}$ the appropriate dependence level $\lambda_a $ (as illustrated in Figure \ref{fig:dendro}) during the training phase. 
The corresponding array is denoted by $\boldsymbol{\lambda} \left[ a \right]   \mapsto \lambda_a$. 
The function $J_{\mathcal{G};\boldsymbol{\lambda}}$ maps the set of possibility distributions $\left( \pi_k \right)_{1\leq k \leq K} $ to the aggregated distribution $\pi_{\text{ens}}$ by executing the computation graph $\mathcal{G}$ and using the dependence levels contained in $\boldsymbol{\lambda}$. 
There are $K-1$ hyperparameters in array $\boldsymbol{\lambda}$ that need to be tuned. They will be automatically set to appropriate values by heuristic search, see \ref{sec:heuristic_search_for_dependency_parameters} for a presentation of this grid search based heuristic. 
It is noteworthy that this heuristic will use HAC to define clusters of classifiers, thereby reducing the number of hyperparameters to tune as this amounts to merge some nodes of the dendrogram and apply some t-norm to more than two possibility distributions.  





\subsection{Adaptive aggregation w.r.t. informational content} 
\label{sub:adaptive_aggregation_w_r_t_informational_content}

When the predictions delivered by classifier $c_k$ are poorer than those of another classifier $c_{k'}$, it is instrumental to reduce the impact of $c_k$ on the decisions issued by the ensemble. Regardless of the formal definition behind what are called "poor predictions", we propose to use the following mechanism to gradually fade classifier $c_k$ out of the ensemble: for a given scalar $\alpha_k \in \left[ 0;1 \right] $, we update all conditional possibility distributions related to $c_k$ as follows:
\begin{equation}
  \pi_{k|j} \leftarrow  \left( 1-\alpha_k \right) \pi_{k|j} + \alpha_k, \forall j \in \Omega. \label{eq:discount}
\end{equation}

This mechanism is equivalent to an operation known as discounting \cite{Sha76}. When $\alpha=0$, then classifier $c_k$ influence on the ensemble is not reduced. When $\alpha_k=1$, classifier $c_k$ is discarded from the ensemble since we obtain constant one possibility distributions which are the neutral element of t-norms and the t-norm based aggregation method introduced in the previous subsection.

Obviously, we need to find a value of the discounting coefficient tailored for each classifier and in line with what poor predictions are meant to be. Again, it is tempting to set these $K$ hyperparameters using grid search but the corresponding complexity calls for a more subtle strategy. Similarly as for dependency hyperparameters, we will resort to a heuristic search. 

Among other possibilities, our solution consists in binding the discounting rates together using the following formula:
\begin{equation}\label{eq:alpha}
  \alpha_k = 1 - \left( \frac{1 - \hat{r}\left[ c_k \right]}{1 - \underset{k'}{\min}\: \hat{r}\left[ c_{k'} \right]} \right)^{\rho},
\end{equation}
where $\hat{r}$ is the estimated error rate on the validation set and $\rho \in \left[ 0;+\infty \right] $ is a hyperparameter to tune by grid search. Using the above equation, the best base classifier is not discounted and we have $\hat{r}\left[ c_k \right] \leq \hat{r}\left[ c_{k'} \right] \Rightarrow \alpha_k \leq \alpha_{k'}$.

\subsection{Fully adaptive aggregation} 
\label{sub:fully_adaptive_aggregation}


The fully adaptive version (w.r.t. both dependence and informative content) of SPOCC is referred to as \textbf{adaSPOCC}. The corresponding training and test phases are described in Algorithm \ref{adaspocc-train} and \ref{adaspocc-test} respectively. 
A python implementation can be downloaded at \href{https://github.com/john-klein/SPOCC}{this link}. 

    \begin{algorithm}
    \DontPrintSemicolon
     \caption{adaSPOCC - training phase}
     \KwData{validation set $\mathcal{D}_{\text{val}}$, classifiers $\left( c_k \right)_{k=1}^K$, number of class labels $\ell$.}

     Execute SPOCC - training phase (algorithm \ref{spocc-train})

     \For{$k\in \left\{ 1,\hdots, K \right\}$}{ 

      \For{$k'\in \left\{ k,\hdots, K \right\}$}{
        Compute the dissimilarity $D_{kk'} \leftarrow 1 - \kappa \left( c_k,c_{k'}\right)$ using \eqref{eq:lambda}.

        Assign $D_{k'k} \leftarrow D_{kk'}$.
      }

     }

     Obtain dendrogram $\mathcal{G}$ by applying HAC to dissimilarity matrix $\mathbf{D} $.

     Apply heuristic to set parameters $ \lambda_a \in \boldsymbol\lambda$ (see \ref{sec:heuristic_search_for_dependency_parameters}).

     Compute parameters $\left( \alpha_k\right)_{k=1}^{K}$ as in \eqref{eq:alpha}.     

     Update all conditional possibility distributions as in \eqref{eq:discount}.

     Return possibility distributions $\left( \pi_{k|j} \right)_{\substack{ 1 \leq k \leq K \\ 1 \leq j \leq \ell}}$, dendrogram $\mathcal{G}$, array $\boldsymbol{\lambda}$.     
     \label{adaspocc-train}
     \end{algorithm}

    \begin{algorithm}
    \DontPrintSemicolon
     \caption{adaSPOCC - test phase}
     \KwData{input $\mathbf{x}$, classifiers $\left( c_k \right)_{k=1}^K$, possibility distributions $\left( \pi_{k|j} \right)_{\substack{ 1 \leq k \leq K \\ 1 \leq j \leq \ell}}$, dendrogram $\mathcal{G}$, array $\boldsymbol{\lambda}$.}

     \For{$k\in \left\{ 1,\hdots, K \right\}$}{
      Compute individual classifier prediction $j_k \leftarrow \mathsf{ind} \left(c_k \left( \mathbf{x} \right) \right)$.

     }

     $\pi_{\text{ens}} \leftarrow J_{\mathcal{G};\boldsymbol{\lambda}} \left( \pi_{1|j_1} , \hdots,  \pi_{K|j_K} \right) $ (computation graph execution).

     Return $c_{\text{ens}} \left( \mathbf{x} \right) \leftarrow \underset{y \in \Omega}{\arg\max} \; \pi_{\text{ens}} \left( y \right)$.
     
     \label{adaspocc-test}
     \end{algorithm}



\subsection{Properties of the possibilistic ensemble} 
\label{sub:properties_of_the_possibilistic_ensemble}

In this paper, we adopt a normative approach for the selection of a classifier decision aggregation mechanism. 
In this subsection, we give sketches of proofs showing that robustness properties (a) to (c) hold for adaSPOCC asymptotically:
\begin{itemize}
  \item Property (a): if $c_k$ is a random classifier then when $n_{\text{val}} \rightarrow \infty$, each conditional distribution $p_{Y | c_k=\omega_j}$ converges to a uniform distribution so DPT turns it into a constant one possibility distribution, which is the neutral element of $\mathcal{T}_\lambda$.

  \item Property (b): let $c_k$ denote an adversorial classifier, i.e. $r \left[ c_k \right] > \frac{\ell-1}{\ell} $. (ada)SPOCC uses the following rectified classifier $c_k^{(\text{rec})} = h \circ c_k$ defined as 
  \begin{equation}
    c_k^{(\text{rec})} \left( \mathbf{x} \right) = \underset{y \in \Omega}{\arg\max} \; p \left( Y=y | c_k \left( \mathbf{x} \right)  \right). \label{eq:rec}
  \end{equation}
  We have 
  \begin{align} \hspace{-0.5cm}
  1 - r \left[ c_k^{(\text{rec})} \right] &= \sum_{y'\in \Omega} p \left( Y=y' | c_k^{(\text{rec})}=y' \right) p \left( c_k^{(\text{rec})}=y' \right). 
  \end{align}
  Moreover, we can write 
  \begin{multline}
  \hspace{-1cm} p \left( Y=y' | c_k^{(\text{rec})}=y' \right) = \sum_{y'' \in \Omega} p \left( Y=y' | c_k^{(\text{rec})}=y' , c_k=y'' \right) \\ p \left( c_k=y'' | c_k^{(\text{rec})}=y' \right).
  \end{multline}
  
  Given the definition of $c_k^{(\text{rec})}$ we know that $p \left( c_k=y'' | c_k^{(\text{rec})}=y' \right)=0$ if $y'' \not\in h^{-1}\left( y' \right)  $. The definition also gives 
  \begin{align}
    p \left( Y=y' | c_k^{(\text{rec})}=y' , c_k =y'' \right) &= \underset{y \in \Omega}{\max} \; p \left( Y=y | c_k = y'' \right).
  \end{align}
  The maximal probability value of a discrete variable is always greater or equal than $\frac{1}{\ell}$ therefore
  \begin{align}
  \hspace{-0.5cm} p \left( Y=y' | c_k^{(\text{rec})} =y' \right) &\geq \frac{1}{\ell} \sum_{y'' \in h^{-1} \left( y' \right)  }  p \left( c_k=y'' | c_k^{(\text{rec})}=y' \right) \\
  & \geq \frac{1}{\ell} p \left( c_k \in h^{-1} \left( y' \right) | c_k^{(\text{rec})}=y' \right).
  \end{align}
  Again, given the definition of $c_k^{(\text{rec})}$ we know that $p \left( c_k \in h^{-1} \left( y' \right) | c_k^{(\text{rec})}=y' \right)=1$. 
  Since $c_k$ is not the random classifier, at least one of the conditional distributions $p_{Y|c_k}$ is not uniform in which case the inequality is strict. We thus obtain $1 - r \left[ c_k^{(\text{rec})} \right] > \frac{1}{\ell}$.

  Finally, when $n_{\text{val}} \rightarrow \infty$, if $c_k \left( \mathbf{x} \right) = y $ and $c_k^{(\text{rec})}\left( \mathbf{x} \right) =y'$, the $y^{\text{th}}$ column of $\mathbf{M}^{(k)}$ will be identical to the $y'^{\text{th}}$ column of the confusion matrix of $c_k^{(\text{rec})}$ so they will be mapped to identical possibility distributions.
  
  \item Property (c): when $n_{\text{val}} \rightarrow \infty$, the likelihood ratio appearing in \eqref{eq:lambda} writes
  \begin{align}
  \frac{ \mathcal{L}_0 }{\mathcal{L}_1} &= \prod_{i=1}^{n_{\text{val}}} \frac{p \left( c_k \left( \mathbf{x}^{(i)} \right)  \right) p \left( c_{k'} \left( \mathbf{x}^{(i)} \right)  \right)}{p \left( c_k \left( \mathbf{x}^{(i)} \right)  , c_{k'} \left( \mathbf{x}^{(i)} \right)  \right)}.
  \end{align}
  If $c_{k'}$ is a copy of $c_k$ then $p \left( c_k \left( \mathbf{x}^{(i)} \right)  , c_{k'} \left( \mathbf{x}^{(i)} \right)  \right) = p \left( c_k \left( \mathbf{x}^{(i)} \right)  \right) = p \left( c_{k'} \left( \mathbf{x}^{(i)} \right)  \right)$ and 
  \begin{align}
  \frac{ \mathcal{L}_0 }{\mathcal{L}_1} &= \prod_{i=1}^{n_{\text{val}}} p \left( c_k \left( \mathbf{x}^{(i)} \right)  \right) .
  \end{align}
  If $c_k$ is not a constant function, then probabilities are smaller than one and $\frac{ \mathcal{L}_0 }{\mathcal{L}_1} \rightarrow 0$. The pair of classifiers $\left( c_k,c_{k'} \right) $ will thus be detected as maximally dependent by HAC and they will be aggregated using $\mathcal{T}_{1} = \mathcal{T}_\wedge$ hence property (c) holds in this case.

  When $c_k$ is a constant function, then both $c_k$ and $c_{k'}$ will yield identical possibility distributions that are a Dirac function. In this case, the output of $\mathcal{T}_{\lambda}$ will also be this Dirac function, meaning that idempotence is always true in these circumstances. Note that, however, the procedure described in \ref{sub:adaptive_aggregation_w_r_t_dependency} will fail to detect the dependency between $c_k$ and $c_{k'}$. There are plenty of ways to thwart this issue as constant classifiers are not difficult to detect. In practice, we will use add-one Laplace smoothing to estimate probabilities $p \left( c_k \right) $ so we will never obtain a Dirac function as possibility distribution.

\end{itemize}

Properties (a) to (c) rely on asymptotic estimates of multinomial distribution parameters which, from the strong law of numbers, converge almost surely to their exact values. Consequently, the properties do not hold only in expectation or with high probability but systematically (when $n_{\text{val}}$ is large).

Although properties (a) to (c) are not as strong as \eqref{eq:oracle}, adaSPOCC is a scalable aggregation technique as the number of parameters it requires to learn from the validation set is in $O \left( \ell^2 K \right) $ while the number of parameters to learn from $\mathcal{D}_{\text{val}}$ in \cite{balakrishnan2015simple} is in $O \left( \ell ^{K+1} \right) $ and is therefore doomed to overfit when $K$ is large. 


\subsection{Incremental aggregation} 
\label{sub:online_and_incremental_aggregation}

When a new classifier $c_{K+1}$ must be appended to the ensemble, it suffices to compute its corresponding confusion matrix $\mathbf{M}^{(K+1)}$ to be able to run SPOCC. 
All previously estimated parameters (confusion matrices and possibility distributions of the other classifiers) can be readily re-used.

Going incremental for adaSPOCC is not as straightforward as for SPOOC. 
A new coefficient $\alpha_{K+1}$ needs to be computed but this latter is deduced from $\mathbf{M}^{(K+1)}$ so this is not an issue. 
However, the matrix $\mathbf{D}$ also needs to be updated by appending a new line and a new column to it, which makes $K-1$ new entries to compute because $\mathbf{D}$ is symmetric and its diagonal elements are irrelevant. 
Then, HAC must be re-run. To increase the level of incrementalism of adaSPOCC in this regard, it is possible to use an incremental clustering algorithm such as \cite{menon2019online}. 
The newly coming classifier will be either appended to an existing cluster or a new cluster that solely contains $c_{K+1}$ will be created. 
In the first case, the hyperparameters $\boldsymbol\lambda$ can be left unchanged. 
In the second case, there is an additional node in the dendogram and one additional hyperparameter must be estimated by grid search. Since we perform grid search for only one such parameter, this is obviously faster than the heuristic search described by Algorithm \ref{heuris_lambda}. 
In conclusion, adaSPOCC is also a incremental aggregation algorithm and all previously estimated parameters are also re-used without needing to be updated. 



\section{Experimental results} 
\label{sec:experimental_results}

In this section, we present a number of experimental results allowing to prove the robustness of SPOCC and adaSPOCC as compared to other aggregation techniques. The section starts with results obtained when the base classifiers are trained on a synthetic dataset and are meant to highlight performances discrepancies in simple situations where robustness is required. Another set of experiments on real datasets are also presented to prove that the method is not only meaningful on toy examples.

\subsection{General setup} 
\label{sub:general_setup}

Designing numerical experiments allowing to compare aggregation methods is not a trivial task. A crucial aspect consists in training a set of base classifiers that achieve a form of diversity \cite{woz14} so that the fusion of their predictions has a significant impact on performances. Among other possibilities \cite{brown2005diversity}, we chose to induce diversity by feeding the base classifiers with different disjoint subsets of data points at training time. The subsets are not chosen at random but instead in a deterministic way allowing each base classifier to focus on some regions of the input space and thus learn significantly different decision frontiers. 

Because the union of the validation sets is an i.i.d. sample of $p \left( X,Y \right) $ and aggregation methods have access to the predictions of each learner on this set union, well designed aggregation methods are able to restore high levels of performances even if base learners are trained from non-i.i.d. samples. 
This allows us to test if aggregation methods are relatively agnostic to the quality of the data used to train the base learners. 

Each aggregation technique is fed only with the predictions of the base classifiers on the validation set in order to tune hyperparameters or learn the combination itself. Consequently, SPOCC and adaSPOCC are only compared to well established methods that use the same level of information. The benchmarked aggregation techniques are :
\begin{itemize}
  \item classifier selection\footnote{Selection can be regarded as a special type of fusion.} based on estimated accuracies of the base classifiers,
  \item weighted vote aggregation based on estimated accuracies of the base classifiers,
  \item exponentially weighted vote aggregation based on estimated accuracies of the base classifiers,
  \item naive Bayes aggregation,
  \item Bayes aggregation,
  \item stacking.

\end{itemize}

In the exponentially weighted vote aggregation, accuracies are not directly used as vote weight (as in standard weighted vote aggregation) but are mapped to weights using a softmax function. This function has a positive temperature hyperparameter that regulates the assignment of weights. When this parameter is zero, then we retrieve unweighted vote aggregation whereas when it is very large, then we retrieve classifier selection.

Bayes aggregation relies on \eqref{eq:optim_agg}. The conditional distributions involved in this equation are learned from the validation set. Naive Bayes aggregation uses conditional independence assumptions that allow to factorize probabilities as

\begin{align}
p \left( y | \mathbf{c} \left( \mathbf{x} \right)  \right) &\propto p \left( y \right)  \prod_{k=1}^K p \left( c_k \left( \mathbf{x} \right) | y  \right). 
\end{align}

The conditional independence assumptions are not realistic but yield a model with far less parameters to learn as compared to Bayes aggregation.

For each of the estimated probabilities involved in the mechanism behind SPOCC, adaSPOCC, Bayes or naive Bayes aggregation, we perform add-one-Laplace smoothing to avoid computational issues related to zero probabilities. The chosen t-norm for SPOCC is $\mathcal{T}_{5}\approx \mathcal{T}_{\wedge}$.

Finally, we also train a softmax regression to map classifier predictions to the true class labels. This approach belongs to a methodology known as stacking \cite{wolpert1992stacked}. 
An $L_2$ regularization term is added to the cross-entropy loss. A positive hyperparameter regulates the relative importance of the regularization term.

All hyperparameters are tuned automatically using a cross-validated grid search on the validation set. For each hyperparameter, the grid contains 100 points. When the hyperparameter is unbounded, a logarithmic scale is used to design the grid. 

The statistical significance of the reported results are given in terms of $95\%$ confidence intervals estimated from bootstrap sampling. When the accuracies of two aggregation methods have overlapping confidence intervals, the performance discrepancy is not significant. A companion python implementation of adaSPOCC and benchmark methods can be downloaded at \href{https://github.com/john-klein/SPOCC}{this link}


\subsection{Synthetic Data} 
\label{sub:synthetic_data}
\begin{figure}
\begin{center}
  \includegraphics[width=.4\textwidth]{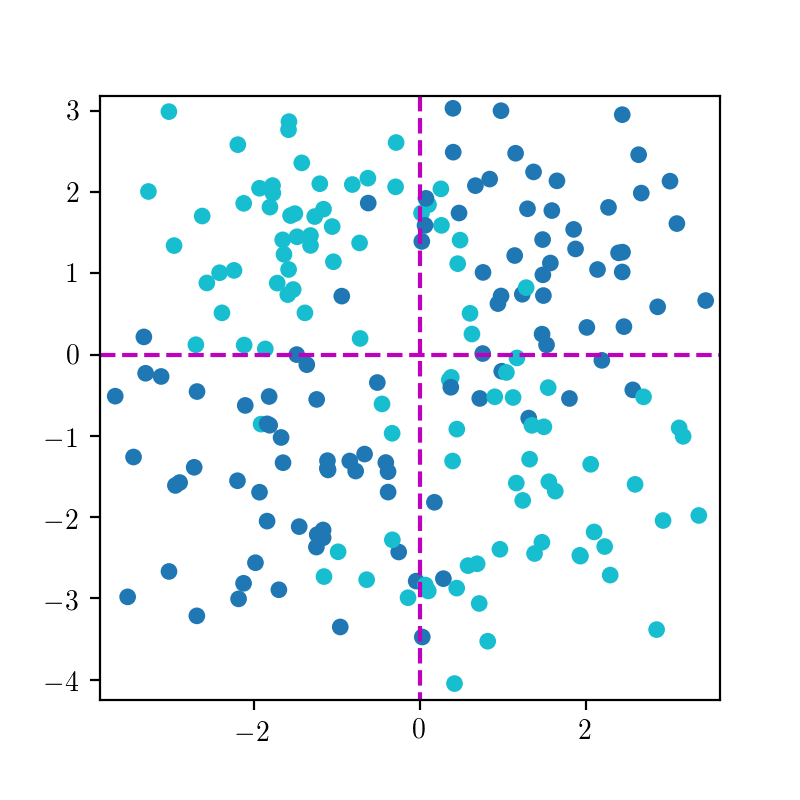}
  \caption{Synthetic dataset obtained from four Gaussian distributions. Examples belonging to class $\omega_1$ are in blue while those belonging to class $\omega_2$ are in cyan. Optimal decision frontiers are in magenta. \label{fig:synth_data}} 
\end{center}
\end{figure}

In this subsection, we use a very simple generating process to obtain example/label pairs. Data points are sampled from four isotropic Gaussian distributions. The centers of these Gaussian distributions are located at each corner of a centered square in a 2D input space ($d=2$). The standard deviations of each of the distribution is 1. There are $\ell=2$ possible class labels: $\Omega= \left\{ \omega_1; \omega_2 \right\}$. Points such that $x_1$ and $x_2$ are both positive belong to $\omega_1$. Points such that $x_1$ and $x_2$ are both negative also belong to $\omega_1$. All the other points belong to $\omega_2$. Figure \ref{fig:synth_data} shows one such dataset obtained from this generating process. 

\begin{figure}
\begin{center}
  \includegraphics[width=.2\textwidth]{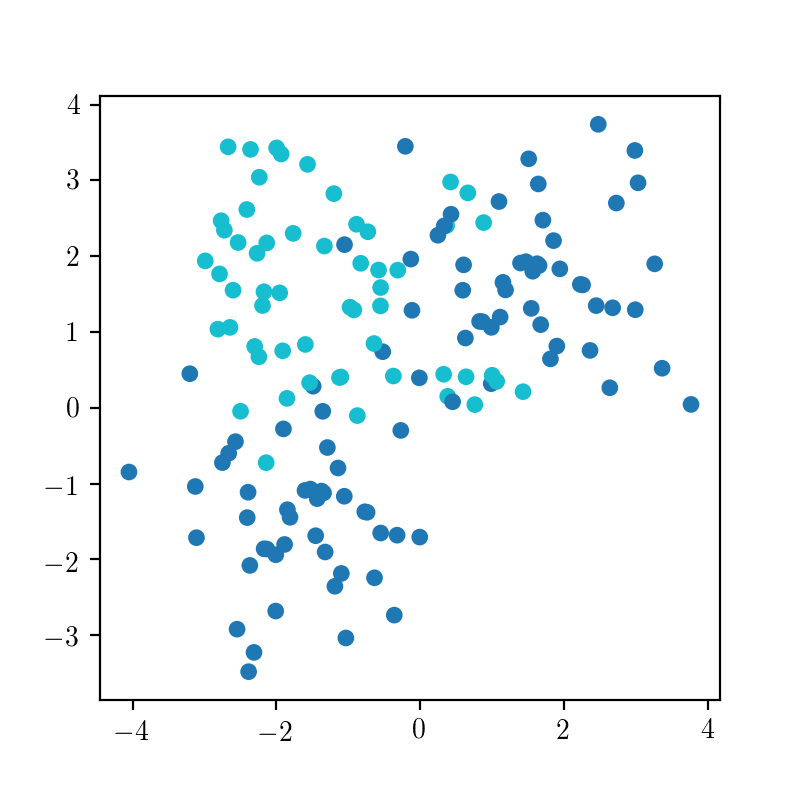} \includegraphics[width=.2\textwidth]{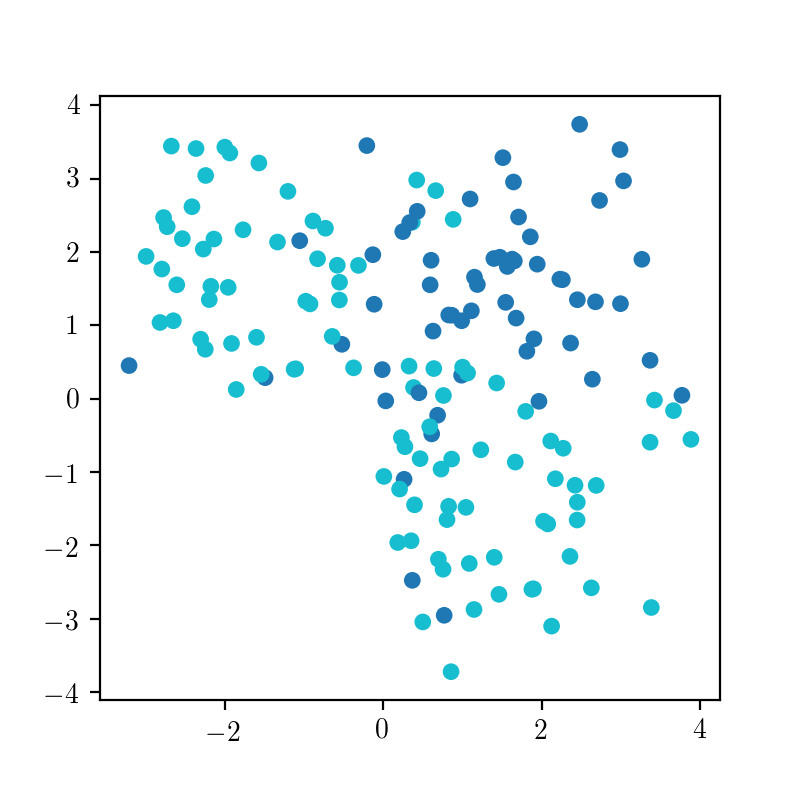} \includegraphics[width=.2\textwidth]{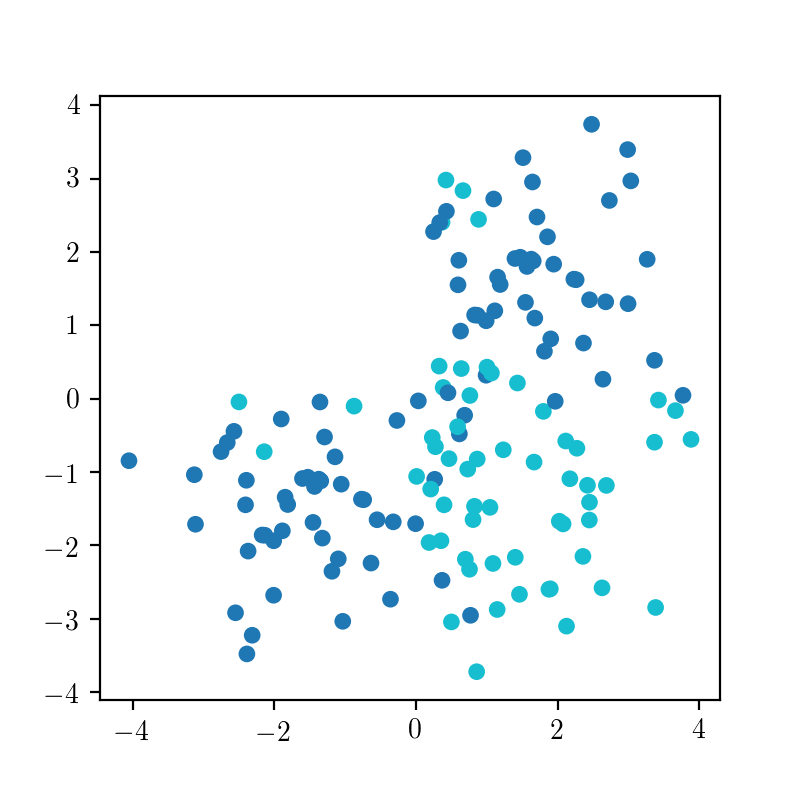} \includegraphics[width=.2\textwidth]{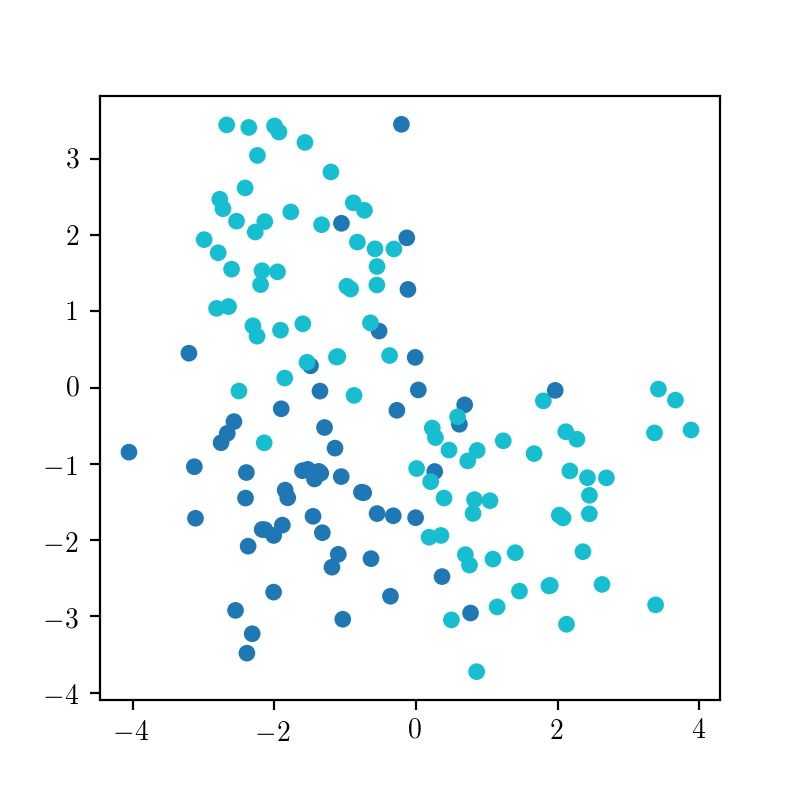}
  \caption{Subsets of the data seen respectively by $c_1$ to $c_4$\label{fig:synth_subsets}} 
\end{center}
\end{figure}

In this series of experiments, the dataset has $n=200$ points and is divided in four overlapping subsets as depicted in Figure \ref{fig:synth_subsets}. Then a randomly selected portion of $80\%$ of each such subset is used to train one of the base classifier. The remaining $20\%$ are used for the validation set. 
 Each base classifier $c_k$ sends the corresponding set of prediction/label pairs $\left\{ \left(    c_k \left( \mathbf{x} \right) ,y \right) \right\}_{\left( \mathbf{x},y \right) \in \mathcal{D}_{\text{val}} }$ to the aggregation method. 

Since we have access to the data generating process, the test set is dynamically created. 
We sample test points until the observed accuracies of all the tested methods are with probability $0.95$ in their respective Clopper-Pearson confidence intervals of half-width $0.2\%$. 
The whole procedure is repeated 100 times. 
The averaged test errors in this case are thus a good approximation of the generalization errors of the tested methods. 

Given the shape of optimal decision frontiers, the base classifiers trained in this subsection are decision trees with a maximal depth of two.

\subsubsection{Robustness w.r.t. adversaries}

Among other possibilities, adversarial predictions are simulated by sampling from a Bernoulli distribution $Z \sim \text{Ber}\left( \theta \right) $. Given $Z=1$, the prediction of a base classifier is replaced with another (arbitrarily selected) class label that will not coincide with the classifier prediction. When $Z=0$, the classifier prediction is unchanged. Consequently, an adversarial classifier built in this way from a base classifier with an error rate lower than random guess will achieve an error rate greater then random guess as $\theta \rightarrow 1$.

\begin{figure}
\begin{center}
  \includegraphics[width=.8\textwidth]{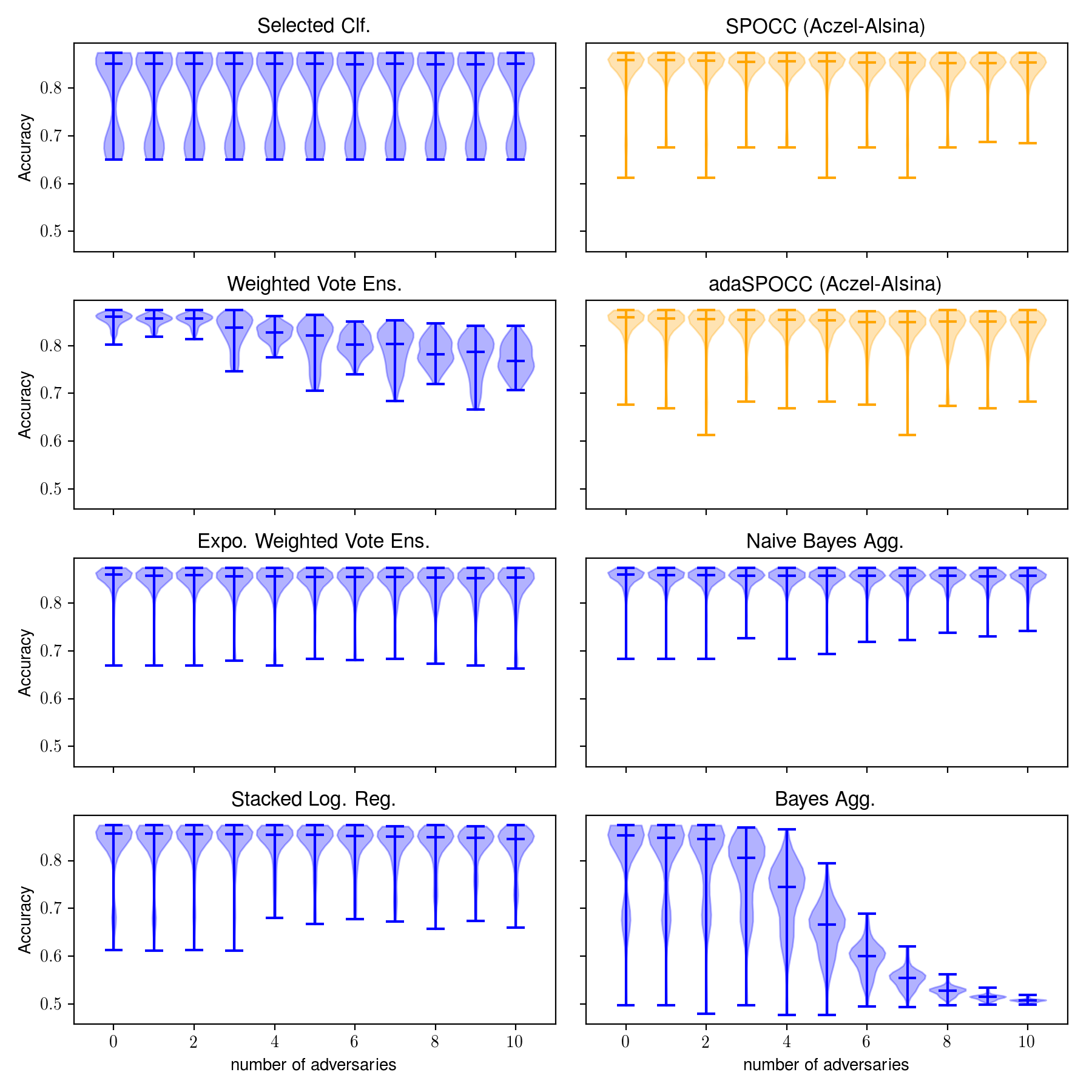}
  \caption{Evolution of accuracy distributions (violin plots) for several aggregation methods w.r.t. the number of adversaries. SPOCC and adaSPOCC are in orange while other methods are in blue.   \label{fig:adv}} 
\end{center}
\end{figure}

The evolution of the classification accuracy of the benchmarked aggregation methods as the number of adversaries grows can be witnessed on Figure \ref{fig:adv}. For simplicity, all adversaries are built from the same base classifier ($c_1$) with $\theta = 0.5$. Two methods cannot maintain the same level of performances when the number of adversaries increases: weighted vote ensemble and Bayes aggregation. For the weighted vote ensemble, this is explained by the fact that the number of misleading classifiers outnumber legitimate classifiers and start to obtain a majority of votes. For Bayes aggregation, the performances are degrading simply because of overfitting. Indeed, Bayes aggregation has a number of parameter to learn that is exponential in $K$ while SPOCC and other methods have a number of parameters at most linear in $K$.

\subsubsection{Robustness w.r.t. faults}

Erroneous predictions are simulated by sampling from a Bernoulli distribution $Z \sim \text{Ber}\left( \theta \right) $. Given $Z=1$, the prediction of a base classifier is replaced with an (arbitrarily selected) class label that will coincide with the classifier prediction with probability $\frac{1}{\ell}$. When $Z=0$, the classifier prediction is unchanged. Consequently, a noisy classifier built in this way from a base classifier will achieve an error rate equal to $\frac{\ell-1}{\ell}$ (random guess) as $\theta \rightarrow 1$.

\begin{figure}
\begin{center}
  \includegraphics[width=.8\textwidth]{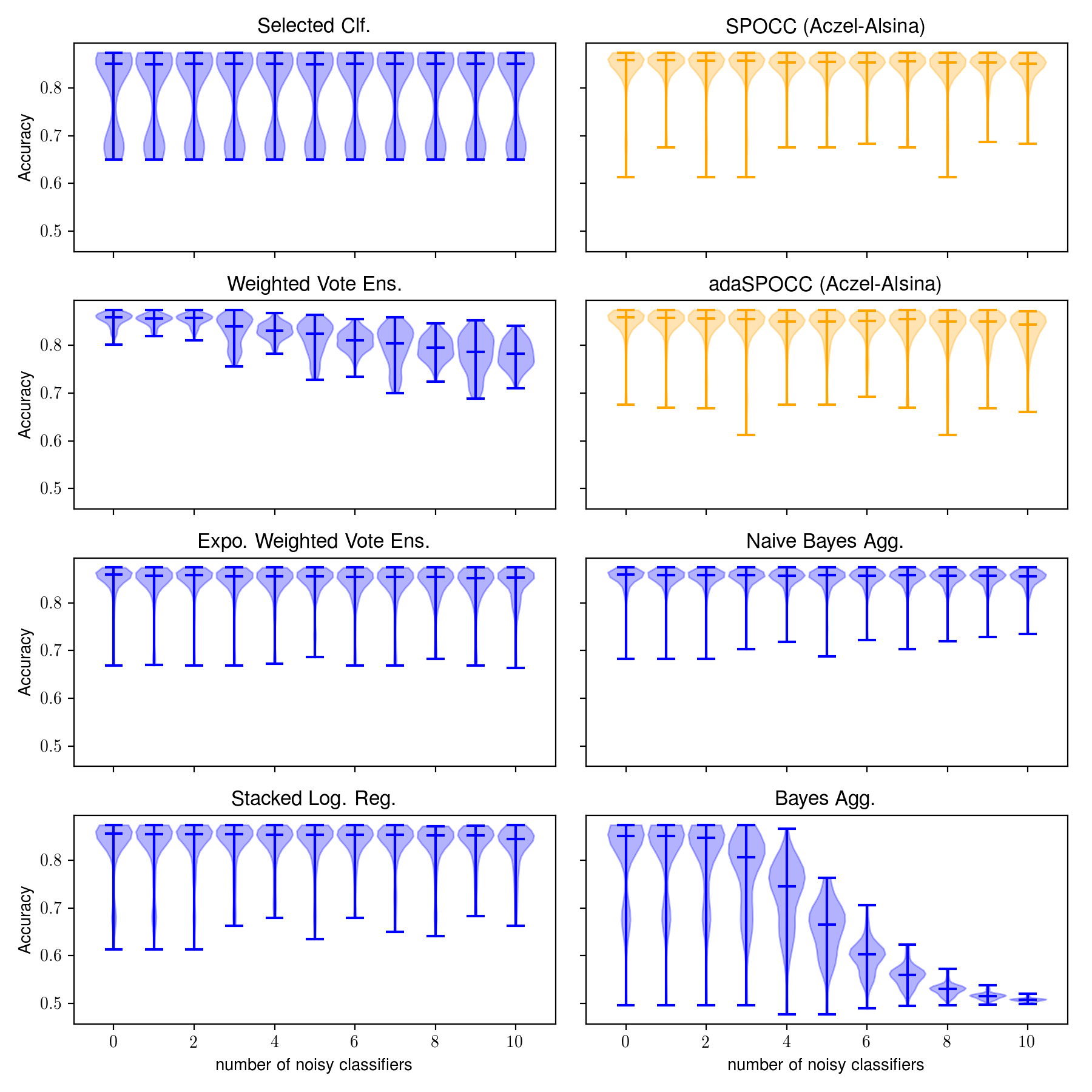}
  \caption{Evolution of accuracy distributions (violin plots) for several aggregation methods w.r.t. the number of noisy classifiers. SPOCC and adaSPOCC are in orange while other methods are in blue.   \label{fig:faults}} 
\end{center}
\end{figure}

The evolution of the classification accuracy of the benchmarked aggregation methods as the number of noisy classifiers grows can be witnessed on Figure \ref{fig:faults}. For simplicity, all noisy classifiers are built from the same base classifier ($c_1$) with $\theta = 0.9$. Similarly as for adversarial classifiers, weighted vote ensemble and Bayes aggregation cannot maintain the same level of performances when the number of perturbed classifiers increases. The same reasons are also behind these performance decays (majority of incorrect classifiers for the weighted vote ensemble and overfitting for Bayes aggregation). 

\subsubsection{Robustness w.r.t. informational redundancy}\label{subsubsec:redundancy}

\begin{figure}
\begin{center}
  \includegraphics[width=.8\textwidth]{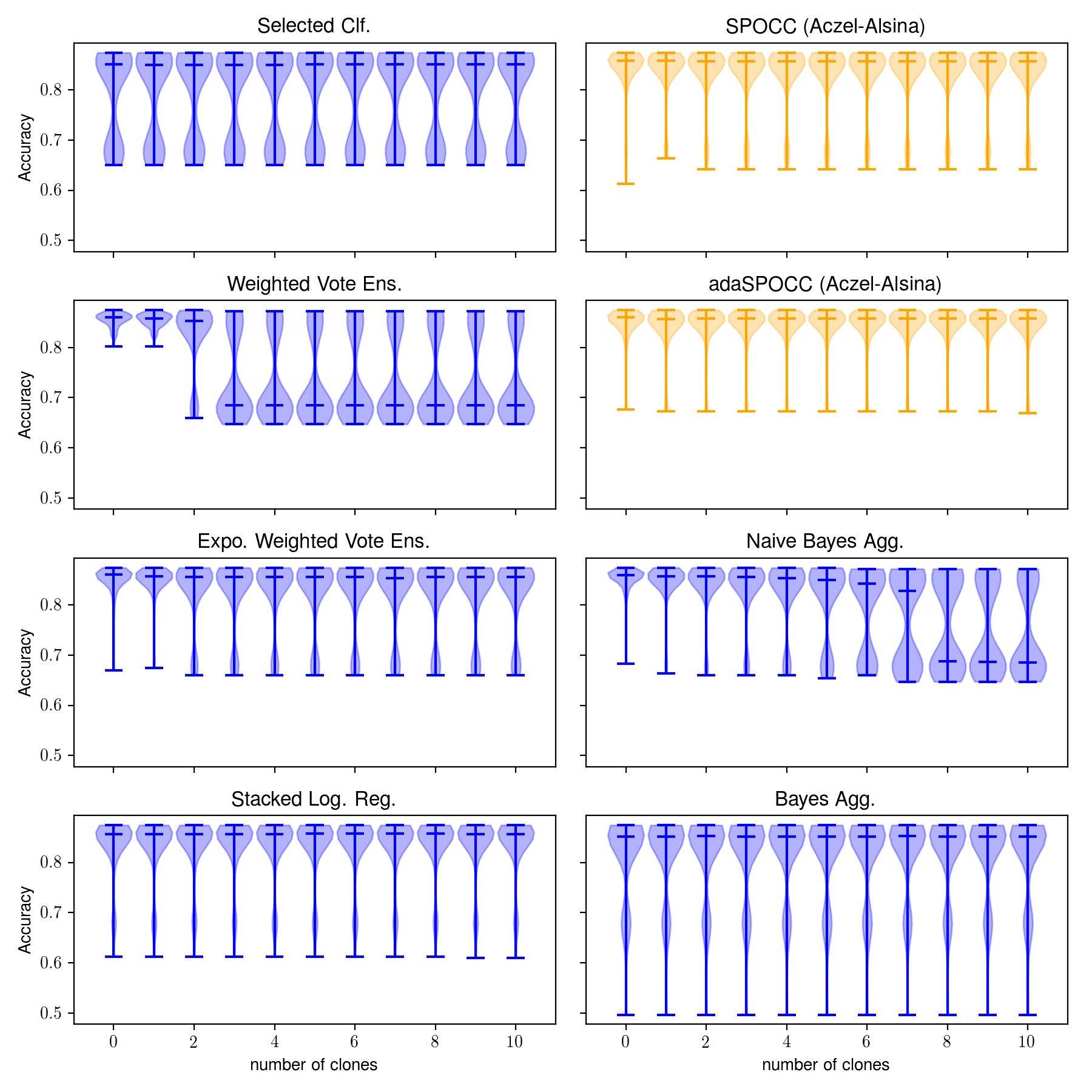}
  \caption{Evolution of accuracy distributions (violin plots) for several aggregation methods w.r.t. the number of copies of $c_1$. SPOCC and adaSPOCC are in orange while other methods are in blue.   \label{fig:redund}} 
\end{center}
\end{figure}

Redundancy in classifier predictions is simulated by adding several copies of one of the base classifiers (classifier $c_1$ in our experiments). As shown in Figure \ref{fig:redund}, this very simple setting allows to observe severe performance decays for the weighted vote ensemble, the exponentially weighted vote ensemble and the naive Bayes aggregation. Vote based ensembles are very sensitive to changes of majority. Naive Bayes aggregation is also sensitive to this phenomenon and suffers from its inability to capture dependency relations between the base classifiers. 

Unlike the previous experiment, it can be noted that Bayes aggregation maintains the same level of performances as the number of clones of $c_1$ increases. Because clones will always produce identical predictions as $c_1$, training the Bayes aggregation in these conditions is equivalent to learn from $K=4$ base classifiers regardless how many copies of $c_1$ are added. However, should these copies be slightly perturbed, then we would observe the same overfitting issues as in the previous experiments.

\subsubsection{Summarizing synthetic data experiment robustness results}
In the previous paragraphs, we have seen which methods are tolerant to adversaries, faults and redundancy and scale well w.r.t. $K$. Only SPOCC, adaSPOCC and stacking seem to be robust w.r.t. each of these forms of difficulties. Beside robustness, their absolute performances also matter. Average\footnote{Averages w.r.t either the number of adversaries, noisy classifiers or clones respectively.} performances are reported in Table \ref{tab:synth_perf} for each experiment as well as the global average on all experiments. 
We also provide as reference the optimal (Bayes) classifier accuracy as well as the performances of the best base classifier $c_k$, i.e. optimal selection.

\begin{figure}
\begin{center}
  \includegraphics[width=.24\textwidth]{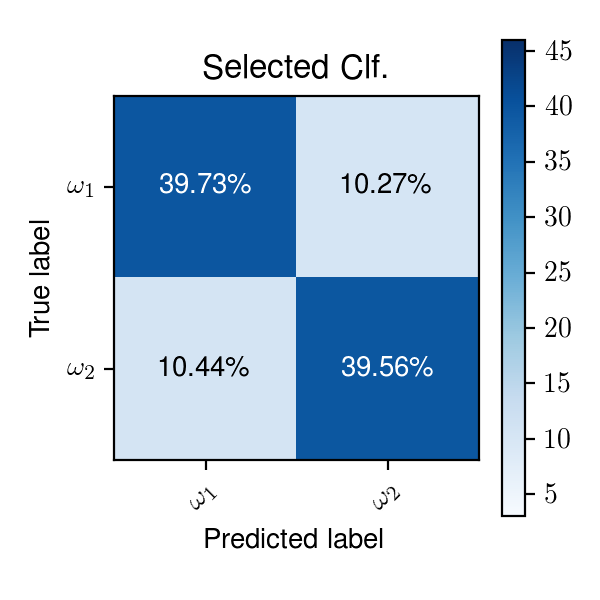} 
  \includegraphics[width=.24\textwidth]{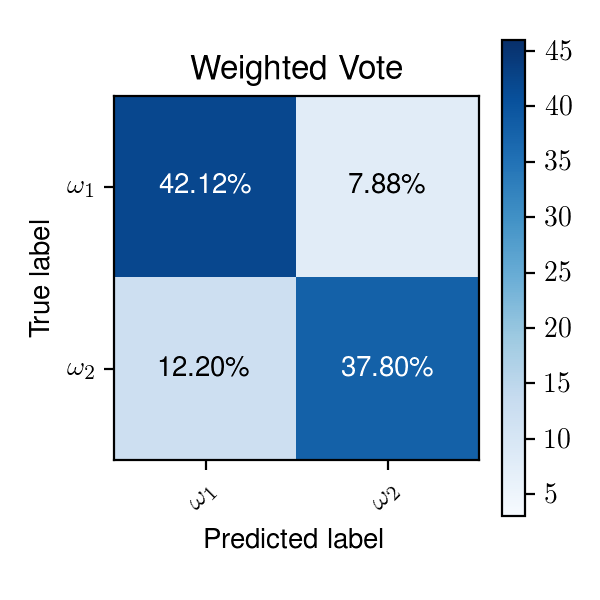} 
  \includegraphics[width=.24\textwidth]{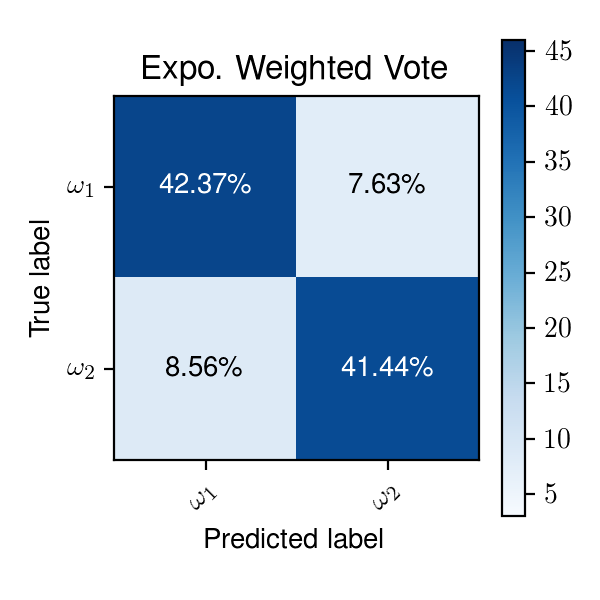} 
  \includegraphics[width=.24\textwidth]{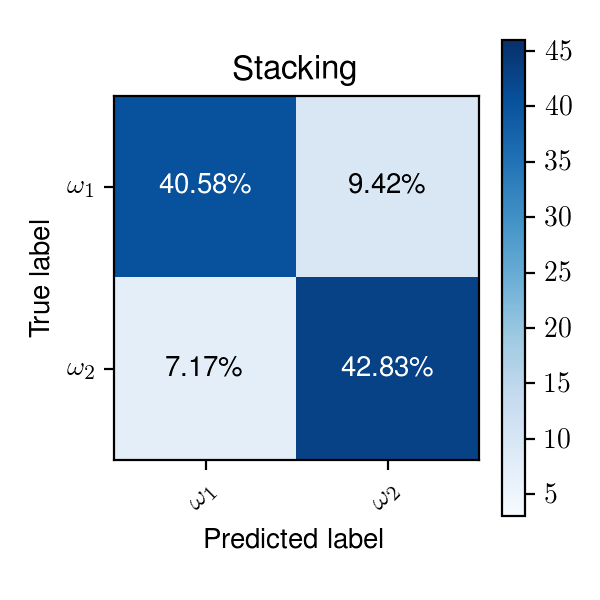}

  \includegraphics[width=.24\textwidth]{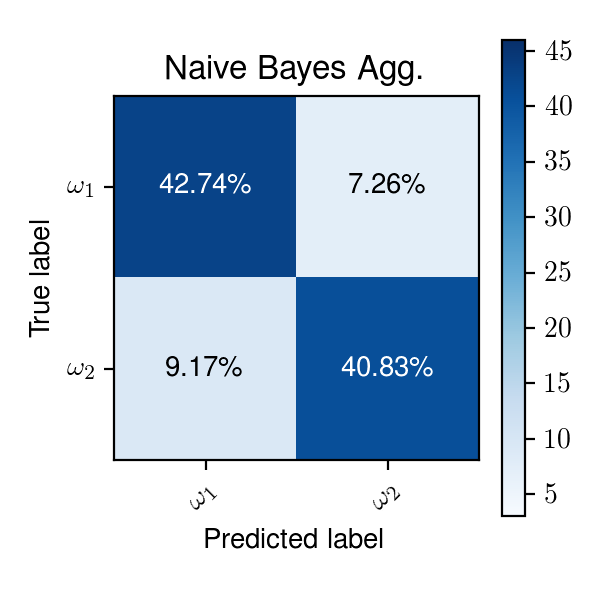} 
  \includegraphics[width=.24\textwidth]{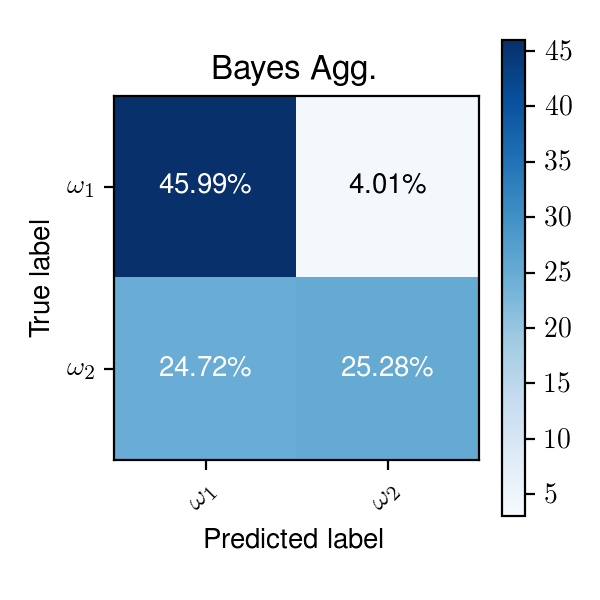} 
  \includegraphics[width=.24\textwidth]{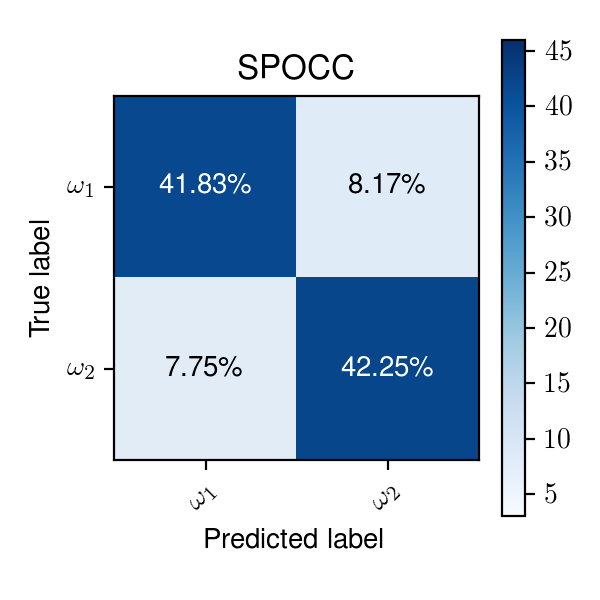} 
  \includegraphics[width=.24\textwidth]{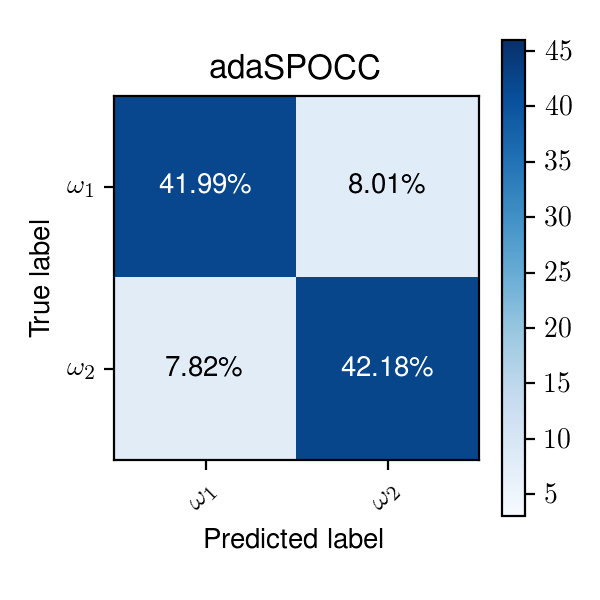} 

  \includegraphics[width=.24\textwidth]{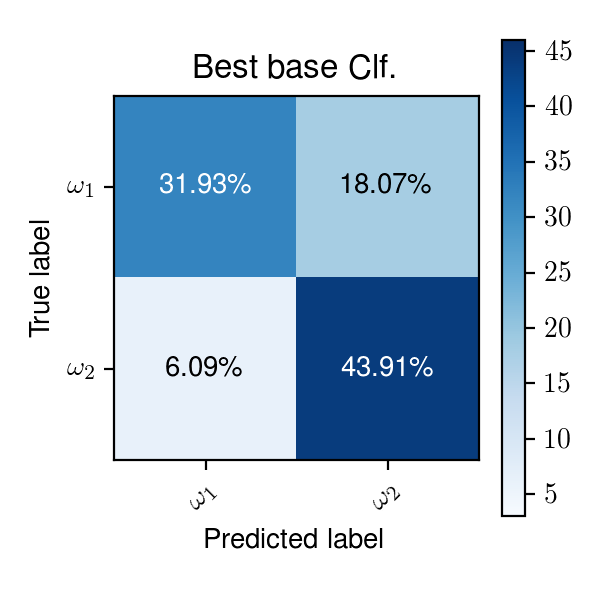} 
  \includegraphics[width=.24\textwidth]{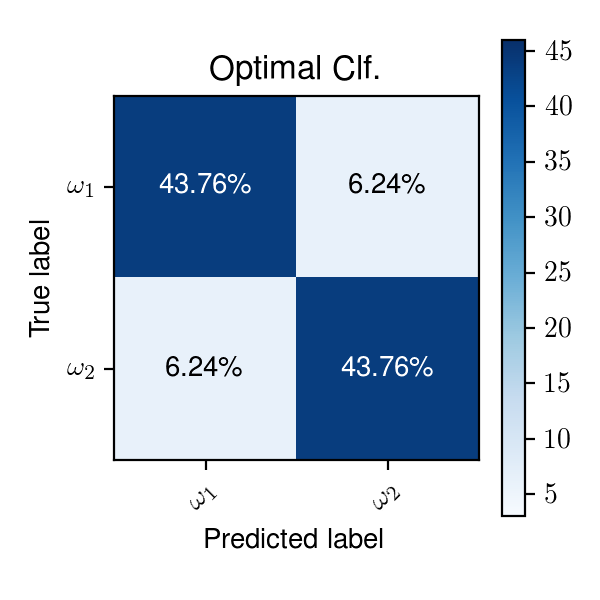} 
  \caption{Normalized averaged confusion matrices for the reported results in the last column of Table \ref{tab:synth_perf}.   \label{fig:conf_mat_synth}} 
\end{center}
\end{figure}

In terms of accuracies, SPOOC or adaSPOOC are always the top 1 or top 2 aggregation approach. While the naive Bayes aggregation is slightly better than SPOOC or adaSPOOC in the two first series of experiments, it performs very significantly worse in the last one and it is outperformed on global average. Stacking and all other methods obtain worse (or sometimes comparable) results as compared to (ada)SPOCC.  Moreover, observe that adaSPOCC achieves the smallest variance, meaning that its performances are more stable across dataset draws. 
Normalized confusion matrices corresponding to the global average on all experiments (last column of Table \ref{tab:synth_perf}) are shown in Figure \ref{fig:conf_mat_synth}. It shows the distribution of error rates in terms of type I and type II errors. Let alone classifier selection (which achieves anyway poor general performances), adaSPOOC is the aggregation method with the smallest type I / type II error discrepancy.

\begin{table}[!t]
\renewcommand{\arraystretch}{0.8}
\caption{Average performances of aggregation methods on the synthetic data. The first figure is the average accuracy followed by the semi-width of the $95\%$ confidence interval width of this latter and the average standard deviation. Best accuracies (or those not statistically significantly different) are in bold characters. }
\label{tab:synth_perf}
\centering
\resizebox{.8\textwidth}{!}{
\begin{tabular}{|c|c|c|c||c|}
\hline
 Method & Adversaries & Faults & Redundancy & Global Average  \\
\hline
Clf. Selection & $79.26\%$ & $79.25\%$ & $79.25\%$ & $79.25\%$  \\
&$\pm 0.52$ & $\pm 0.50$& $\pm 0.51$& $\pm 0.28$\\
& std. $8.56\%$ & std.  $8.56\%$ & std. $8.55\%$ & std. $8.56\%$  \\
\hline

Weighted Vote & $81.47\%$ & $81.89\%$ & $76.46\%$ & $79.94\%$  \\
&$\pm 0.25$ & $\pm 0.23$& $\pm 0.52$& $\pm 0.22$\\
& std. $4.23\%$ & std.  $3.83\%$ & std. $8.86\%$ & std. $6.57\%$  \\
\hline
Exp. Weighted Vote & $84.32\%$ & $84.35\%$ & $82.75\%$ & $83.81\%$  \\
&$\pm 0.23$ & $\pm 0.23$& $\pm 0.39$& $\pm 0.17$\\
& std. $3.86\%$ & std.  $3.77\%$ & std. $6.46\%$ & std. $4.92\%$  \\
\hline
Stacking & $83.43\%$ & $83.50\%$ & $83.38\%$ & $83.44\%$  \\
&$\pm 0.28$ & $\pm 0.29$& $\pm 0.36$& $\pm 0.17$\\
& std. $4.75\%$ & std.  $4.82\%$ & std. $5.89\%$ & std. $5.18\%$  \\
\hline
Naive Bayes Agg. & $\mathbf{85.23}\%$ & $\mathbf{85.22}\%$ & $80.27\%$ & $83.57\%$  \\
&$\pm 0.13$ & $\pm 0.14$& $\pm 0.46$& $\pm 0.19$\\
& std. $2.25\%$ & std.  $2.25\%$ & std. $8.12\%$ & std. $5.55\%$  \\
\hline
Bayes Agg. & $66.20\%$ & $66.36\%$ & $81.16\%$ & $71.24\%$  \\
&$\pm 0.77$ & $\pm 0.79$& $\pm 0.51$& $\pm 0.46$\\
& std. $13.45\%$ & std.  $13.43\%$ & std. $8.40\%$ & std. $13.90\%$  \\
\hline
SPOCC & $84.42\%$ & $84.42\%$ & $83.52\%$ & $\mathbf{84.10}\%$  \\
&$\pm 0.23$ & $\pm 0.23$& $\pm 0.34$& $\pm 0.16$\\
& std. $3.82\%$ & std.  $3.82\%$ & std. $5.76\%$ & std. $4.61\%$  \\
\hline
adaSPOCC & $84.01\%$ & $84.01\%$ & $\mathbf{84.54}\%$ & $\mathbf{84.16}\%$  \\
&$\pm 0.23$ & $\pm 0.23$& $\pm 0.52$& $\pm 0.13$\\
& std. $3.89\%$ & std.  $3.89\%$ & std. $4.12\%$ & std. $4.00\%$  \\
\hline
\hline
Best base Clf.  & $82.65\%$ & $82.65\%$ & $82.65\%$ & $82.65\%$  \\
&$\pm 0.41$ & $\pm 0.41$& $\pm 0.41$& $\pm 0.41$\\
& std. $6.87\%$ & std.  $6.87\%$ & std. $6.87\%$ & std. $6.87\%$  \\
\hline
Optimal Clf. & $87.52\%$ & $87.52\%$ & $87.52\%$ & $87.52\%$  \\
&$\pm \approx 0$ & $\pm 0.52$& $\pm \approx 0$ & $\pm \approx 0$\\
& std. $\approx 0\%$ & std.  $\approx 0\%$& std. $\approx 0\%$ & std. $\approx 0\%$  \\
\hline
\end{tabular}
}
\end{table}

\subsubsection{Other experimental aspects}

The main goal of this experimental section is to illustrate the robustness properties (a) to (c) that adaSPOOC possesses. The results reported in the previous paragraphs match this purpose. 
However, other aspects are also interesting to examine. 
In the following paragraphs, we investigate the behavior of the tested aggregation methods under two different circumstances:
\begin{itemize}
  \item when the base classifiers are heterogeneous,
  \item when the dataset is imbalanced, meaning that $p \left( Y \right) $ is not uniform.
\end{itemize}

\paragraph{Heterogeneous ensemble of classifiers} 
An important advantage of the class label agnostic aggregation setting over others is that no assumption at all are made on the base classifiers and therefore any training algorithm can be used to derive them. 
To illustrate this ability, we reproduce the same experiment as in \ref{subsubsec:redundancy} with different base classifiers. Now, $c_1$ is trained using logistic regression, $c_2$ is a $5$-nearest neighbor classifier, $c_3$ is an SVM with radial basis function as kernel while $c_4$ is a decision tree like before. 
Classifier $c_1$ will achieve smaller accuracy because a linear decision function underfits the data in this case. 
Since this classifier will be progressively duplicated, it will also check the ability of methods to cope with increasingly many lesser accurate classifiers.

\begin{figure}
\begin{center}
  \includegraphics[width=.8\textwidth]{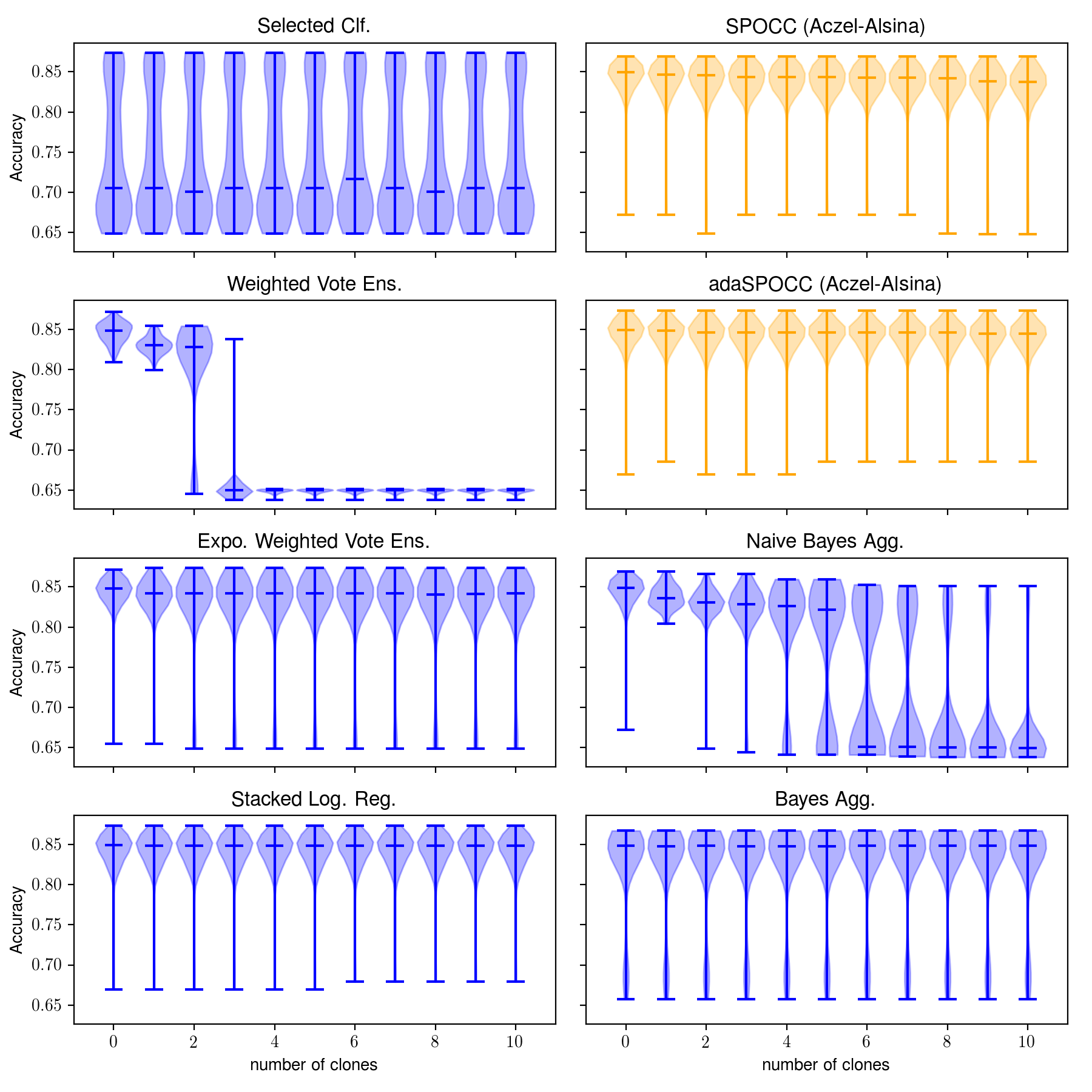}
  \caption{Evolution of accuracy distributions (violin plots) on an heterogeneous ensemble of 4 classifiers for several aggregation methods w.r.t. the number of copies of $c_1$. SPOCC and adaSPOCC are in orange while other methods are in blue.   \label{fig:hetero}} 
\end{center}
\end{figure}

The corresponding results are displayed in Figure \ref{fig:hetero}. Since at least one member of the ensemble performs more poorly than in \ref{subsubsec:redundancy}, all methods have their accuracy distributions eroded. 
However, the conclusions from the previous experiments are confirmed as the same methods achieve robustness to information redundancy and corruption. 
It is also made clear that the ability of the aggregation to overcome these difficulties does not lie with the training algorithms employed to obtain the base classifiers. 

\paragraph{Class label imbalance}
In practice, it is common that the generative process underlying our data is such that the class label probability distribution is not uniform. 
In the previous set of experiments, such an imbalance was in place for the base learner but not for the aggregation methods. 
We now modify the generative process such $p \left( Y=\omega_1 \right) = \beta$ and $p \left( Y=\omega_2 \right)= 1 - \beta $.
The level of imbalance is progressively reduced as $\beta$ increases. 
The set of examined values is: $\beta \in \left\{ 0.05; 0.1; 0.15; ..; 0.5 \right\}$.

\begin{figure}
\begin{center}
  \includegraphics[width=.8\textwidth]{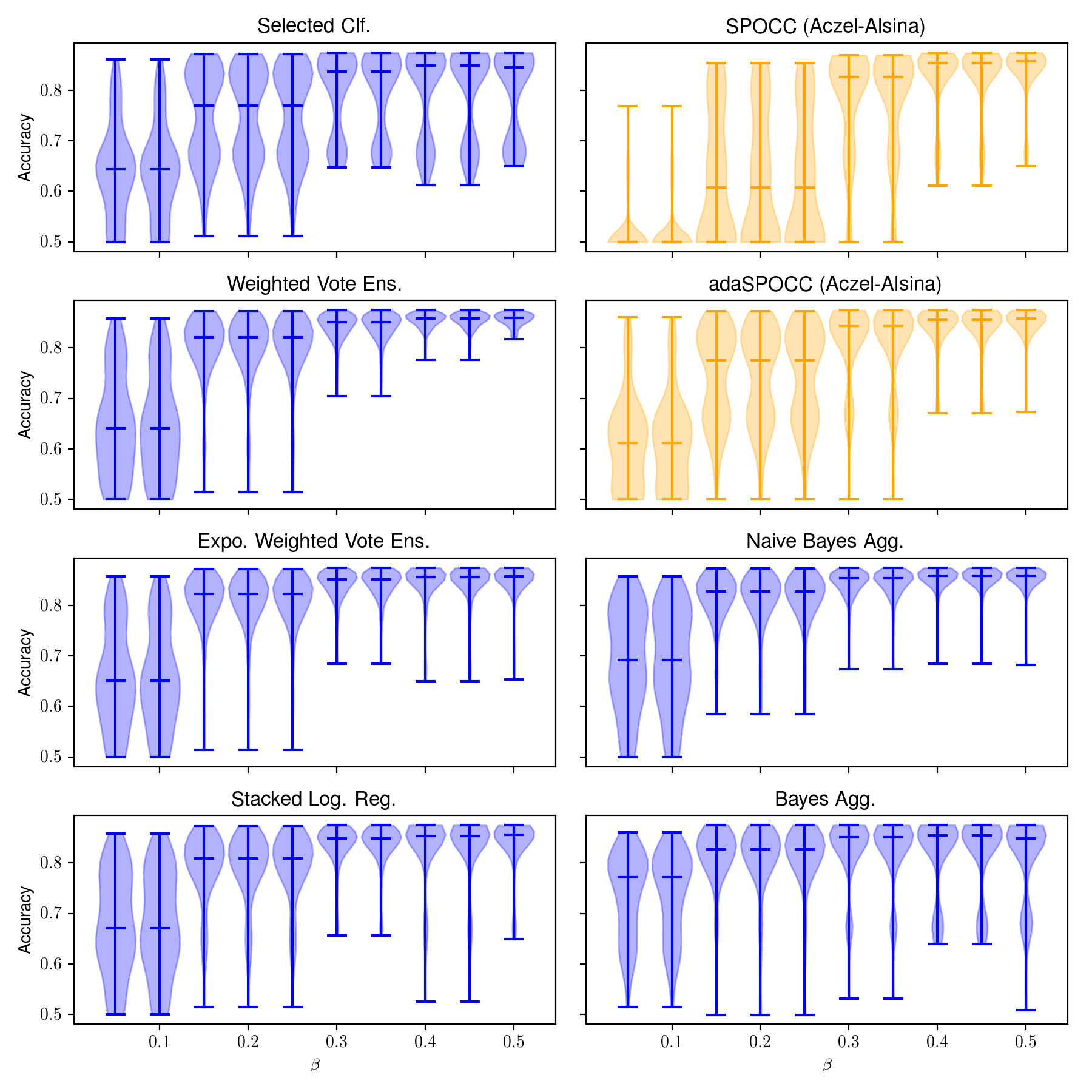}
  \caption{Evolution of accuracy distributions (violin plots) with class label imbalance for several aggregation methods w.r.t. $\beta$ (first class label probability). SPOCC and adaSPOCC are in orange while other methods are in blue.   \label{fig:imbalance}} 
\end{center}
\end{figure}

The corresponding results are displayed in Figure \ref{fig:imbalance}. 
For any value of $\beta$, this setting is very favorable to majority based methods. 
Indeed, if the decision tree training went alright, we should obtain approximately the following predictors
$$ c_1 \left( \mathbf{x} \right) = \begin{cases} \omega_2 & \text{ if } x_1<0 \text{ and } x_2>0 \\ \omega_1 & \text{otherwise} \end{cases} \; c_2 \left( \mathbf{x} \right) = \begin{cases} \omega_1 & \text{ if } x_1>0 \text{ and } x_2>0 \\ \omega_2 & \text{otherwise} \end{cases} $$
$$ c_3 \left( \mathbf{x} \right) = \begin{cases} \omega_2 & \text{ if } x_1>0 \text{ and } x_2<0 \\ \omega_1 & \text{otherwise} \end{cases} \; c_4 \left( \mathbf{x} \right) = \begin{cases} \omega_1 & \text{ if } x_1<0 \text{ and } x_2<0 \\ \omega_2 & \text{otherwise} \end{cases}. $$

We see that, for any $\mathbf{x}$, there are 3 classifiers out of 4 that yield correct predictions therefore majority voting based aggregations are expected to perform very well. 
In addition, we also have $r \left[ c_i \right]\approx r\left[ c_j \right] $ for any $i$ and $j$ so the weighted vote and exponentially weighted vote aggregation are nearly equivalent to majority voting. As we can see, these two methods perform very well when $\beta=0.5$ and can tolerate imbalance up to $\beta=0.15$. Below this value, there are two few points in the class $\omega_1$\footnote{When $\beta=0.1$, classifiers $c_2$ and $c_4$ have access (in expectation) to 8 data points in class $\omega_1$ and only four such points when $\beta=0.05$.} for the decision tree to learn meaningful prediction rules and thus their aggregation (regardless of the method) is not meaningful either. Stacking can also easily learn a combination rule that mimics majority vote. It thus compares favorably to vote based methods.

The naive Bayes aggregation also works very well in this setting because $p \left( c_1 | Y = \omega_1 \right) \approx \mathbb{I}\left\{ c_1 = \omega_1 \right\} $, $p \left( c_2 | Y = \omega_2 \right) \approx \mathbb{I}\left\{ c_2 = \omega_2 \right\} $, $p \left( c_3 | Y = \omega_1 \right) \approx \mathbb{I}\left\{ c_3 = \omega_1 \right\} $ and $p \left( c_4 | Y = \omega_2 \right) \approx \mathbb{I}\left\{ c_4 = \omega_2 \right\} $. Consequently, the naive Bayes aggregation will easily rule out the classifier yielding an incorrect prediction. 
We can see that this method achieves comparable performances as compared to the vote based ones.

The Bayes aggregation is a gold standard because it infers the optimal decision rule as explained in \ref{sub:combining_classifiers}. However it still has $\ell \left( \ell^K - 1 \right) = 30 $ parameters to learn from 40 data points in the validation set and will thus slightly overfit. 
It thus achieves worse accuracies than naive Bayes or vote based aggregation but seems to better handle extreme imbalance. 

Because indicator functions (or Dirac masses) are fixed points of DPT and zero is the absorbing element of t-norms, SPOCC and adaSPOCC can enjoy the same type of information as the naive Bayes aggregation does. They indeed perform well when $\beta>0.35$. 
However, they exhibit higher sensitivity to imbalance than other methods. As often, adaSPOCC appears to be more robust than SPOCC. 
We believe this is due to the fact that they are discriminative aggregation models in the sense that they do not rely on the whole data distribution but solely operate on the conditional distributions $p \left( Y | c_k \right) $. 
As the experiments on real data will show, adaSPOCC already works pretty well on several imbalanced datasets, however possible fixes for this limitation will be investigated in future works and are discussed in section \ref{sec:conclusion}.
 

\subsection{Real Data} 
\label{sub:real_data}
To upraise the ability of the benchmarked methods to be deployed in more realistic situations (such as decentralized learning), we also need to test them on sets of real data. Since this is essentially useful in a big data context, we chose eight from moderate to large public datasets. The specifications of these datasets are reported in Table \ref{tab:real_specs}. 

\begin{table}[!t]
\renewcommand{\arraystretch}{1.2}
\caption{Real dataset specifications}
\label{tab:real_specs}
\centering
\resizebox{.99\textwidth}{!}{
\begin{tabular}{|>{\centering\let\newline\\\arraybackslash}p{.20\textwidth}||>{\centering\let\newline\\\arraybackslash}p{.12\textwidth}|>{\centering\let\newline\\\arraybackslash}p{.20\textwidth}|>{\centering\let\newline\\\arraybackslash}p{.15\textwidth}|>{\centering\let\newline\\\arraybackslash}p{.14\textwidth}|>{\centering\let\newline\\\arraybackslash}p{.15\textwidth}|>{\centering\let\newline\\\arraybackslash}p{.3\textwidth}|}
\hline
 Name   & Size $n$ & Dim. $d$ & Nbr. of classes $\ell$ & Data type & Class imbalance & Source \\
 \hline
 20newsgroup   & $18846$ & $100$ (after red.) & 20 & text & yes & sklearn \\
 \hline
 MNIST & $70000$ & $784$ & $10$ & image & no & sklearn  \\
 \hline
 Satellite & $6435$ & $36$ & $6$ & image features & yes & UCI repo. (Statlog) \\
 \hline
 Wine & $6497$ & $11$ & $2$ (binarized) & chemical features & yes & UCI repo. (Wine Quality) \\
 \hline
 Spam & $4601$ & $57$ & $2$ & text & yes & UCI repo. (Spam) \\
 \hline
 Avila & $10430$ & $10$ & $2$ (binarized) & layout features & yes & UCI repo. (Avila) \\
 \hline
 Drive & $58509$ & $48$ & $11$ & current statistics & no & UCI repo. (Sensorless Drive Diagnosis) \\
 \hline
 Particle & $130064$ & $50$ & $2$ & signal & yes & UCI repo. (MiniBooNE particle identification) \\
 \hline
\end{tabular}
}
\end{table}

Example entries from the 20newsgroup data set are word counts obtained using the term frequency - inverse document frequency statistics. We reduced the dimensionality of inputs using a latent semantic analysis \cite{deerwester1990indexing} which is a standard practice for text data. We kept 100 dimensions. Also, as recommended, we stripped out each text from headers, footers and quotes which lead to overfitting. 
Besides, for the Wine and Avila datasets, the number of class labels is originally 10 and 12 respectively. We binarized these classification tasks because some classes have very small cardinalities which is problematic for our experimental design in which datasets are divided into several distinct subsets. Indeed, some subsets may possess no example at all of some classes which leads to imprecise labeling which is beyond the scope of this paper. 
To circumvent these acute class imbalance issues, classes were merged as follows:
\begin{itemize}
  \item In the Wine data set, class labels are wine quality scores. Two classes are obtained by comparing scores to a threshold of 5. 
  \item In the Avila dataset, class labels are middle age bible copyist identities. The five first copyists are grouped in one class and the remaining ones in the other class.
\end{itemize}

Unlike synthetic data sets, we need to separate the original dataset into a train set and a test set. To avoid a dependency of the reported performances w.r.t train/test splits, we perform 2-fold cross validation (CV). Also, we shuffled at random examples and repeated the training and test phases $100$ times.

To induce diversity in the base classifiers, we separated the training data into 6 distinct pieces using the following procedure: for each data set, for each class, 
\begin{enumerate}
   \item apply principal component analysis to the corresponding data,
   \item project this data on the dimension with highest eigenvalue,
   \item sort the projected values and split them into $6$ subsets of cardinality $n_i/6$ where $n_i$ is the proportion of examples belonging to class $\omega_i$.
 \end{enumerate} 
We argue that this way of splitting data leads to challenging fusion tasks because some base classifiers may see data that are a lot easier to separate than it should and will consequently not generalize very well. Actually, the training data to which classifier $c_k$ has access is a non-i.i.d. sample of the distribution of $\left( X,Y \right) $. 

We used softmax regression with an $L_2$ regularization term to train the base classifiers. The regularization hyperparameter is set to default (i.e. 1.0).

To make sure that robustness observations from the previous subsection are confirmed on real data, we also add two noisy classifiers to the ensemble. Both noisy classifiers are built from $c_1$ with $\theta=0.01$ therefore they are perturbed copies of $c_1$.

\begin{table}[!t]
\renewcommand{\arraystretch}{1.2}
\caption{Classification accuracies (with bootstrap confidence intervals and standard deviations) for several real data sets when $6$ base classifiers were trained separately on disjoint subsets of the datasets. Datasets were split in an non-i.i.d. way using a PCA based protocol. Two slightly noisy copies of $c_1$ were added so there are $K=8$ base classifiers to aggregate.  }
\label{tab:real3}
\centering
\resizebox{\textwidth}{!}{
\begin{tabular}{|c||c|c|c|c|c|c|c|c|}
\hline
 Method & 20newsgroup & MNIST & Satellite & Wine & Spam & Avila & Drive & Particle \\
\hline
Clf. Selection & $40.86\%$ & $70.86\%$ & $75.23\%$ & $64.92\%$ & $87.31\%$ & $60.56\%$  & $51.55\%$  & $82.11\%$ \\
&  $\pm 0.12\%$ &  $\pm 0.08\%$ &  $\pm 0.15\%$ &  $\pm 0.26\%$ &  $\pm 0.18\%$ &  $\pm 0.34\%$  &  $\pm 0.31\%$ &  $\pm 0.15\%$ \\
& std. $0.83\%$ & std.  $0.79\%$ & std. $1.54\%$ & std. $2.60\%$ & std. $1.83\%$ & std. $3.48\%$  & std. $3.18\%$ & std. $1.56\%$ \\
\hline
Weighted Vote & $38.67\%$ & $67.48\%$ & $75.34\%$ & $\mathbf{65.60}\%$ & $85.22\%$ &  $60.77\%$  & $50.72\%$  & $79.35\%$ \\
&  $\pm 0.15\%$ &  $\pm 0.06\%$ &  $\pm 0.08\%$ &  $\pm 0.16\%$ &  $\pm 0.20\%$ &  $\pm 0.35\%$  &  $\pm 0.33\%$ &  $\pm 0.24\%$ \\
& std. $1.59\%$ & std. $0.61\%$ & std. $0.85\%$ & std. $1.62\%$  & std. $2.05\%$ & std. $3.60\%$  & std. $3.38\%$  & std. $2.50\%$ \\
\hline
Exp. Weighted Vote & $\mathbf{42.17}\%$ & $75.10\%$ & $77.88\%$ & $\mathbf{65.63}\%$ & $88.50\%$ &  $62.30\%$  & $57.01\%$  & $82.64\%$ \\
&  $\pm 0.08\%$ &  $\pm 0.06\%$ &  $\pm 0.10\%$ &  $\pm 0.18\%$ &  $\pm 0.14\%$ &  $\pm 0.35\%$  &  $\pm 0.46\%$ &  $\pm 0.12\%$ \\
& std. $0.83\%$ & std. $0.61\%$ & std. $1.01\%$ & std. $1.81\%$  & std. $1.42\%$ & std. $3.60\%$  & std. $4.67\%$  & std. $1.29\%$ \\
\hline
Stacking & $19.01\%$ & $37.82\%$ & $61.72\%$ & $64.65\%$  & $\mathbf{89.67}\%$ &  $\mathbf{65.58}\%$  & $37.10\%$  & $\mathbf{83.75}\%$ \\
&  $\pm 0.15\%$ &  $\pm 0.24\%$ &  $\pm 0.10\%$ &  $\pm 0.26\%$ &  $\pm 0.13\%$ &  $\pm 0.12\%$  &  $\pm 0.28\%$ &  $\pm 0.11\%$ \\
& std. $1.58\%$ & std. $2.43\%$ & std. $1.04\%$ & std. $2.76\%$ & std. $1.28\%$ & std. $1.24\%$  & std. $2.78\%$  & std. $1.12\%$ \\
\hline
Naive Bayes Agg. & $39.32\%$ & $76.11\%$ & $78.06\%$ & $64.21\%$  & $87.83\%$ & $60.76\%$  & $66.39\%$  & $80.10\%$ \\
&  $\pm 0.43\%$ &  $\pm 0.17\%$ &  $\pm 0.09\%$ &  $\pm 0.21\%$ &  $\pm 0.23\%$ &  $\pm 0.33\%$  &  $\pm 0.60\%$ &  $\pm 0.30\%$ \\
 & std. $4.43\%$ & std. $1.69\%$ & std. $0.87\%$ & std. $2.16\%$  & std. $2.34\%$ & std. $3.42\%$  & std. $6.05\%$  & std. $2.90\%$ \\
\hline
Bayes Agg. & Intract. & Intract. & $66.07\%$ & $63.89\%$  & $88.90\%$ & $\mathbf{65.39}\%$  & Intract.  & $83.29\%$ \\
&  &  &  $\pm 0.17\%$ &  $\pm 0.27\%$ &  $\pm 0.14\%$ &  $\pm 0.12\%$  &   &  $\pm 0.11\%$ \\
 &  &  & std. $1.65\%$ & std. $2.77\%$  & std. $1.42\%$ & std. $1.26\%$  &   & std. $1.12\%$ \\
\hline
SPOCC & $35.99\%$ & $78.00\%$ & $77.55\%$ & $63.60\%$  & $86.74\%$ & $61.91\%$  & $66.86\%$  & $73.31\%$ \\
&  $\pm 0.19\%$ &  $\pm 0.22\%$ &  $\pm 0.10\%$ &  $\pm 0.10\%$ &  $\pm 0.34\%$ &  $\pm 0.33\%$  &  $\pm 0.53\%$ &  $\pm 0.23\%$ \\
 & std. $1.93\%$ & std. $2.29\%$ & std. $1.00\%$ &  std. $1.07\%$ & std. $3.32\%$ & $3.27\%$  & std. $5.48\%$  & std. $2.35\%$ \\
\hline
adaSPOCC & $41.19\%$ & $\mathbf{79.13}\%$ & $\mathbf{78.59}\%$ & $64.92\%$  & $89.26\%$ & $63.33\%$  & $\mathbf{67.75}\%$  & $82.13\%$ \\
&  $\pm 0.10\%$ &  $\pm 0.26\%$ &  $\pm 0.08\%$ &  $\pm 0.25\%$ &  $\pm 0.14\%$ &  $\pm 0.32\%$  &  $\pm 0.47\%$ &  $\pm 0.14\%$ \\
 & std. $1.03\%$ & std. $2.66\%$ & std. $0.77\%$ &  std. $2.51\%$ & std. $1.40\%$ & $3.21\%$  & std. $4.79\%$  & std. $1.47\%$ \\
\hline
\hline
Best base Clf.  & $42.11\%$ & $70.89\%$ & $75.82\%$ & $65.66\%$ & $87.73\%$  & $62.59\%$  & $52.06\%$  & $82.33\%$ \\
&  $\pm 0.06\%$ &  $\pm 0.06\%$ &  $\pm 0.08\%$ &  $\pm 0.16\%$ &  $\pm 0.09\%$ &  $\pm 0.11\%$  &  $\pm 0.24\%$ &  $\pm 0.12\%$ \\
& std. $0.66\%$ & std. $0.66\%$ & std. $1.25\%$ & std. $1.42\%$ & std. $1.27\%$ & std. $1.78\%$  & std. $2.63\%$  & std. $1.65\%$ \\
\hline
Centralized Clf. & $57.43\%$ & $91.44\%$ & $82.43\%$ & $73.72\%$ & $92.18\%$ &  $68.23\%$  & $74.72\%$  & $88.56\%$ \\
&  $\pm0.04 \%$ &  $\pm 0.02\%$ &  $\pm 0.04\%$ &  $\pm 0.05\%$ &  $\pm 0.05\%$ &  $\pm 0.04\%$  &  $\pm 0.03\%$ &  $\pm 0.25\%$ \\
& std. $0.39\%$ & std. $0.12\%$ & std. $0.44\%$ &  std. $0.53\%$ & std. $0.50\%$ & std. $0.47\%$  & std. $0.35\%$  & std. $2.58\%$ \\
\hline
\end{tabular}}
\end{table}

\begin{table}[!t]
\renewcommand{\arraystretch}{1.2}
\caption{Maximal accuracy discrepancy w.r.t. the best approach. Max is taken over the 8 datasets.  }
\label{tab:discrepancies}
\centering
\resizebox{\textwidth}{!}{
\begin{tabular}{c|llllllll}

 \rotatebox{45}{Method} & \rotatebox{45}{Clf. Selection} & \rotatebox{45}{Weighted Vote} & \rotatebox{45}{Exp. Weighted Vote} & \rotatebox{45}{Stacking} & \rotatebox{45}{Naive Bayes Agg.} & \rotatebox{45}{Bayes Agg.} & \rotatebox{45}{SPOCC} & \rotatebox{45}{adaSPOCC} \\
max. discr. & 16.2$\%$ & 17.03$\%$& 10.74$\%$& 41.31$\%$&  4.82$\%$& 12.52$\%$& 10.44$\%$&  2.25$\%$ \\

\end{tabular}}
\end{table}

Average accuracies over random shuffles and CV-folds are given in Table \ref{tab:real3} for $K =8$ base classifiers. Train/validation split ratio are identical to the synthetic dataset case. 
A number of observations can be made based on these results:
\begin{itemize}
  \item Classifier selection based on estimated accuracies is always significantly outperformed by some of the aggregation techniques which shows that the experimental protocol meets its purpose (providing a setting allowing to do better than base classifiers). 
  Even "oracle" classifier selection (reported as best base classifier in Table \ref{tab:real3}) is outperformed in 6 datasets out of 8 and achieves comparable performance in the remaining two.
  \item adaSPOOC always obtain better results than SPOCC which indicates that it is safer to analyze classifier dependencies as well as estimated individual performances on real data.
  \item Memory occupation became problematic for Bayes aggregation whenever $\ell > 6$ as $\ell \left( \ell^K - 1 \right) $ parameters need to be estimated and stored. It achieves unsurprisingly poor performances when $\ell=6$ (Satellite dataset) confirming its inability to scale w.r.t. $\ell$ or $K$.
  \item adaSPOCC is one of the most efficient aggregation approach. It achieves the highest average rank (over the 8 datasets). adaSPOCC has average rank of  2.1 followed by the exponentially weighted vote which is in average the top 3 approach.  
  \item adaSPOCC is robust in the sense that it achieves the minimal maximal discrepancy w.r.t. the best concurrent approach. Absolute values of maximal discrepancies (over the 8 datasets) w.r.t. the best approach are reported in Table \ref{tab:discrepancies}.
\end{itemize}

The corresponding normalized confusion matrices are given in \ref{sec:normalized_confusion_matrices_on_real_data}. These matrices exhibit different patterns, thereby showing that the methods converge to significantly different aggregation strategies. Also, interestingly, the methods do not necessarily issue incorrect predictions under the same circumstances and a second stage of aggregation may take advantage of their respective strengths.



\section{Conclusion} 
\label{sec:conclusion}

In this article, a new classifier aggregation technique is introduced. This technique relies on the framework of possibility theory. Conditional probabilities of class labels given a classifier prediction are estimated on a validation set and transformed in possibility distributions. For each input to be classified, the set of possibility distributions issued by the classifier predictions are regarded as a set of propositions that are conjunctively combined using a t-norm. The obtained method, called SPOCC, is scalable w.r.t. to both the number of class labels and the number of base classifiers. It is also incremental as extra-classifiers can be appended later to the ensemble without re-computing previously derived parameters. 

An adaptive version of this method, called adaSPOCC is also introduced. It is proposed to perform hierarchical agglomerative clustering to identify subsets of classifiers which are not statistically independent. Each such cluster can thus be combined sequentially with different t-norms. T-norms are chosen from the Aczel-Alsina parametric family which allows to reduce the impact of redundant predictions when necessary. Moreover, the individual impact of a base classifier can also be regulated by setting discounting coefficients. When one such coefficient is set to one, the corresponding classifier is discarded from the fusion process. These coefficients as well as the t-norm parameters are automatically tuned using heuristic search monitoring the ensemble accuracy on the validation set. 

The adaptive version of this non-probabilistic aggregation method possesses a number of nice statistical properties. These properties are well supported by several numerical experiments and clearly show its ability to tolerate adversaries, faults or information redundancy. In a series of experiments on a synthetic dataset and involving variable numbers of (respectively) noisy, adversarial or redundant classifiers, adaSPOCC achieves the highest average accuracy by a comfortable margin ($\approx 2.5$ times the 95\% confidence interval width). In a second series of experiments on 8 real datasets and involving several (jointly) noisy, adversarial and redundant classifiers, adaSPOCC achieves the highest average accuracy rank. Thanks to its robustness, its worst performance is limited to a $2.25\%$ misclassification error overhead as compared to the best method. This is the minimal such overhead among all other methods, the second smallest such overhead being $4.82\%$ which is more than two times larger.

There are several future research tracks that we plan to investigate to further develop this contribution.
One of them consists in investigating to what extent the approach is modular w.r.t. imprecise classifiers, i.e. classifiers that can only discriminate between subsets of class labels. Since possibility theory is natively compatible with set theory, this modularity seems not too challenging to achieve as opposed to many other concurrent approaches. 

A second one consists in designing a more efficient way to set parameters $\boldsymbol\lambda$ than grid search (and the heuristic search presented in Appendix A). 
To replace grid search, we plan to try to fit $\pi_{\text{ens}}$  to the one-hot encoding of the true label. It is already known that $L_k$ norm based metrics are relevant distance for possibility distributions as illustrated in \cite{Jou12}. A projected gradient descent would allow us to find a relevant estimate of $\boldsymbol\lambda$.

Another interesting question is to adapt the proposed classifier aggregation method to regression tasks. This will require to use possibilities on the real line for instance which may be computationally more demanding.

Another line of work consists in proposing an alternative version of adaSPOCC in which possibility distributions are built from classifier output scores instead of just their class label predictions. Classifier score are much more informative than class label predictions because they provide a ranked list of candidate solutions equipped with levels of confidence. On the downside, they require either not be agnostic on base learners or to perform a calibration step. Mapping scores to possibilities can be regarded as a regression task for which a dedicated portion of the validation set can be used for fitting.

Finally, while adaSPOCC can tolerate a certain level of class imbalance, its sensitivity to this phenomenon could be reduced in several ways. One can think of turning the class probability distribution into a possibility distribution and combine it to the aggregated possibility distribution issued by adaSPOCC. It is also possible to give different weights to the data points in the validation set and optimize the t-norm with respect to a corresponding weighted empirical risk instead of the unweighted version that we used in this article.



\appendix

\section{Heuristic search for dependency parameters} 
\label{sec:heuristic_search_for_dependency_parameters}

In this appendix, we explain how to set hyperparameters $\left( \lambda_a \right)_{a=1}^{K-1}$ which are necessary to execute the computation graph $\mathcal{G}$ as part of adaSPOCC. Each hyperparameter $\lambda_a$ regulates the level of dependency between operands aggregated using the Aczel-Alsina t-norm $\mathcal{T}_{\lambda_a}$. Each $\lambda_a$ lives in $\left[ 1;+\infty \right] $ and we can use a logarithmic grid and the validation set to assess the impact of a given value of $\lambda_a$ in terms of classification accuracy of the ensemble. However, resorting to baseline grid search has exponential complexity in $K$ and we will thus employ a heuristic search to keep computation time at bay.

To cleverly browse possible values for $\left( \lambda_a \right)_{a=1}^{K-1}$, we can remark that HAC agglomerates classifiers from most dependent ones to least dependent ones. This implies that if $V_a$ is a child node of $V_{a'}$ in $\mathcal{G}$ then $\lambda_a \geq \lambda_{a'}$. Consequently, instead of systematically visit all configurations, we will start by performing a grid search on the full grid for the lowest nodes in the hierarchy and freeze the corresponding hyperparameters. Then, we will move to their parent nodes and perform grid search only for smaller values. This sequential grid search has a maximal complexity equal to $K-1$ times the cost of a 1D grid search.

Another trick allowing to improve the procedure consists in jointly setting a subset of hyperparameters. Clusters can be obtained from $\mathcal{G}$. This is actually the original intent behind HAC. For a given number of clusters $N_c$, clusters are obtained by thresholding cophenetic correlation coefficients between pairs of classifiers. 
More precisely, if a pair of classifiers have a cophenetic correlation distance below the threshold, they are considered to belong to different clusters. Starting with a sufficiently high value of the threshold so that all classifiers are in one unique cluster, the threshold is lowered until the constraint on $N_c$ is violated, i.e. further lowering it yields $N_c + 1$ clusters. 
Note that the obtained clusters always correspond to non-overlapping branches of $\mathcal{G}$. 
This strategy is wrapped up by iterating on $N_c$, starting from $N_c=2$ to $K$.

Algorithm \ref{heuris_lambda} summarizes the proposed heuristic search for dependency parameters.
The 1D grid search for one $\lambda_a$ is performed on a predefined grid and the retained value is the one achieving highest accuracy of the classifier ensemble.

    \begin{algorithm}[h!]
    \DontPrintSemicolon
     \caption{Heuristic search for dependency hyperparameters}

     Initialize $\lambda_a \leftarrow 1$, $\forall V_a \in \mathcal{G}$.

     \For{$N_c$ from $2$ to $K$}{
      Obtain $N_c$ clusters denoted by $\mathcal{C}_1 \hdots \mathcal{C}_{N_c}$.

      \For{$i$ from $1$ to $N_c$}{
        Find $V_a$ s.t. its descendants contain $\mathcal{C}_i$ and no other leaf node.

        Perform grid search jointly for $\lambda_a$ and all $\lambda_{a'}$ in correspondence with non-leaf descendants of $V_a$.

        Append node $V_a$ and all its descendants to the set $\mathsf{Treated}$.

      }
      Obtain list $\mathsf{Remain}$ of those nodes in $\mathcal{G}$ which do not belong to $\mathsf{Treated}$.
      
      Sort list $\mathsf{Remain}$ so that $\mathsf{Remain}[i]$ cannot be an antecedent of $\mathsf{Remain}[j]$ if $i<j$.

      \For{i from $1$ to length of $\mathsf{Remain}$}{
        Perform grid search for $\lambda_a$ where $V_a = \mathsf{Remain}[i]$ and subject to $\lambda_a \leq \lambda_{a'}$ for any $V_{a'}$ that is a descendant of $V_a$.

      }

      Save in $r_i$ the empirical error rate achieved with this value of $\boldsymbol\lambda$.

      \If{$N_c \geq 3$ and $r_i\geq r_{i-1}$ }{
        Revert to the value of $\boldsymbol\lambda$ obtained at the previous iteration.

        Stop looping.
      }

     }
     Return $\boldsymbol\lambda$.
    \label{heuris_lambda}
     \end{algorithm}

      
\section{Deeper insights into possibility theory} 
\label{sec:an_overview_of_possibility_theory}
Possibility theory was introduced by \cite{Zad78} and further developed by \cite{dubois1988possibility} with the motivation to offer a well-defined and formal mathematical representation for linguistic statements that permits handling imprecise or vague information. For instance, the word \textit{cheap} can be given a large set of values according to everyone's subjective definition and context of cheapness.
Possibility values can be interpreted as \textit{degrees of feasibility} of event occurrence. 
An important difference with the way probabilities encode uncertainty is that high possibility values are non-informative while high probability values are. Indeed, a very high possibility for event $A$ means that, should $A$ occur or not, we would not be surprised. If $A$ has a very high probability mass, then we would be surprised that $A$ does not occur. Conversely, low possibility and probability values are both informative as they both indicated that an occurrence of $A$ is unlikely.

Generally speaking, possibilities are a special class of imprecise probabilities which is a setting in which probability values can only be bracketed by two bounds. Indeed, the uncertainty of an event in the possibilistic framework is better upraised by a pair of values (possibility and necessity) which can be seen as probability bounds. 
In this regard, possibility theory is not at all an orthogonal theory to probabilities but provide a convenient language to grasp some aspects of uncertainty that lead to highly complex models in probabilistic language. 
  

\subsection{Main concepts} 
\label{sub:formal_concepts}


Possibility theory can be formalized as follows. Suppose events $B\subseteq \Omega$ can be sorted into one of the following families:
\begin{itemize}
  \item $\mathscr{T}$ the family of true propositions,
  \item $\mathscr{F}$ the family of false propositions,
  \item and $\mathscr{U}$ the family of undecided propositions.
\end{itemize}
Define two functions $\text{N}$ and $\Pi$ that respectively represent the certainty of truth and the possibility of truth. More formally,
\begin{align}
\text{N} \left( B \right) &= \mathbb{I} \left\{ B \in \mathscr{T} \right\}    ,\\ 
\Pi \left( B \right) &= \mathbb{I} \left\{ B \in \mathscr{T}\cup \mathscr{U}  \right\},
\end{align}
where $\mathbb{I}$ is the indicator function. 
Take two events $B_1$ and $B_2$. If they are both certain then so is $B_1\cap B_2$ and if at least of one of them is not certain then $B_1\cap B_2$ is not certain either. We thus have
\begin{equation}
  \text{N} \left( B_1\cap B_2 \right) = \min \left\{   \text{N} \left( B_1 \right);\text{N} \left( B_2 \right) \right\}.
\end{equation}
Conversely, if at least either $B_1$ or $B_2$ is possible then so is $B_1\cup B_2$, hence
\begin{equation}
  \Pi \left( B_1 \cup B_2 \right) =  \max \left\{ \Pi \left( B_1 \right);\Pi \left( B_2 \right) \right\}. \label{eq:poss_measure}
\end{equation}
Moreover, if $B$ is impossible then $B^c$ is surely true, hence
\begin{equation}
  \Pi \left( B \right)  = 1 - \text{N} \left( B^c \right) . \label{eq:poss_necess}
\end{equation}
Now, by preserving the above logical rules but allowing $\text{N}$ and $\Pi$ to take values in $\left[ 0;1 \right] $ and to represent graded membership of families $\mathscr{T}$ and $\mathscr{T}\cup \mathscr{U}$, then we obtain the framework of possibility theory. 

In this framework, the function $\text{N}$ is called necessity measure and the function $\Pi$ is called possibility measure and it is immediate from \eqref{eq:poss_necess} that they encode the same information. From \eqref{eq:poss_measure}, it is also obvious that the possibility measure of any subset $B$ can be obtained from the possibility measures of singletons. The restriction of $\Pi$ to singletons is denoted $\pi$ and called possibility distribution. A possibility distribution has the same memory complexity as a probability distribution.

\begin{ex}\label{ex:1}
Suppose $\Omega=\left\{ \omega_1; \omega_2; \omega_3\right\}$. Suppose classifier $c_1$ is a 4-nearest-neighbor and class labels $\omega_1$ and $\omega_2$ receive two votes when $c_1$ is asked to predict the class label $y$ of some input $\mathbf{x}$. Consequently, aggregation methods receive the following message from $c_1$: $B_1=\left\{ \omega_1 ; \omega_2 \right\} \in \mathscr{T} $. This piece of information is encoded in the following possibility distribution:

\begin{center}
\begin{tabular}{cccc}
  $y$ & $\omega_1$ & $\omega_1$ & $\omega_3$ \\
  $\pi$ & 1 & 1 & 0 
\end{tabular}
\end{center}
The corresponding possibility and necessity measures are given by
\begin{center}
\begin{tabular}{ccccccccc}
  $y$ & $\left\{ \omega_1; \right\}$ & $\left\{\omega_2\right\}$ & $\left\{ \omega_1; \omega_2\right\}$ & $\left\{  \omega_3\right\}$ & $\left\{ \omega_1; \omega_3\right\}$ & $\left\{  \omega_2; \omega_3\right\}$ & $\Omega$ \\
  $\text{N}$ & 0 & 0 & 1 & 0 & 0 & 0 & 1  \\
  $\Pi$ & 1 & 1 & 1 & 0 & 1 & 1 & 1
\end{tabular}
\end{center}

\end{ex}

The above example illustrates the ability of possibility distributions to easily encode set theoretic information. The same information can be encoded by an infinite set of probability distributions, i.e. the set of those distributions such that $p \left( \left\{ \omega_1; \omega_2\right\} \right)=1 $. 
Observe that the probability distribution $p_0 \left( \omega_1 \right) = p \left( \omega_2 \right) = \frac{1}{2}  $ belongs to those distributions in question. 
The distribution $p_0$ does not only inform us that $y$ is either $\omega_1$ or $\omega_2$ but also that there are evidence that these outcomes are equally probable. It is thus more informative than the statement $B_1\in \mathcal{T}$

Note that SPOCC does not exploit this aspect of possibility theory but this is mentioned as one of the perspectives for future work. In this paper, SPOCC leverages the flexibility of possibility theory when it comes to aggregate propositions in the form of possibility distributions.

\subsection{Probability / Possibility transforms} 
\label{sub:probability_possibility_transforms}

In this article, the information issued by classifier $c_k$ that we would like to encode as a possibility distribution corresponds to one column of matrix $\mathbf{M}^{(k)}$. 
After dividing by $n_{\text{val}}$ these entries of $\mathbf{M}^{(k)}$, we obtain a maximum likelihood estimate of the conditional distribution $p \left( y | c_k \left( \mathbf{x} \right)  \right) $. It is therefore necessary to turn the information encoded by this probability distribution into a possibility distribution in order to exploit possibilistic aggregation mechanisms.

To turn a probability distribution $p$ into a possibility one $\pi$, three desirable properties stated by \cite{dubois2004probaposs} are

  \begin{itemize}
  \item \textit{consistency}: $\forall A\subseteq \Omega$, $\Pi(A) \geq p(A)$ where $\Pi$ is the possibility measure spanned by $\pi$. So $\Pi$ is a probability upper bound.
  \item \textit{preference preservation}: $\forall (\omega,\omega') \in \Omega^2$, $p \left( \omega \right)  > p \left( \omega' \right)  \Leftrightarrow \pi \left( \omega \right)  > \pi \left( \omega' \right) $, so there is a form of compatibility between the preferences encoded by $\pi$ and those encoded by $p$. 
  \item \textit{maximal specificity}: $\pi$ is the most informative possibility distribution among those possibility distributions consistent and preserving preferences with $p$. Considering two possibility distribution $\pi^{(1)}$ and $\pi^{(2)}$, the possibility distribution $\pi^{(1)}$ is said to be more informative than $\pi^{(2)}$, denoted $\pi^{(1)} \preceq \pi^{(2)}$, if $\forall \omega$ , $\pi^{(1)}(\omega) \leq \pi^{(2)}(\omega)$. 
  \end{itemize} 

The only transformation verifying the above properties is DPT which is presented in \ref{sub:from_classifier_confusion_matrices_to_possibility_distributions}.

\begin{ex}\label{ex:2}
Suppose $\Omega=\left\{ \omega_1; \omega_2; \omega_3\right\}$. Below are two examples of application of DPT to two conditional probability distributions:

\begin{center}
\begin{tabular}{cccc}
  $y$ & $\omega_1$ & $\omega_1$ & $\omega_3$ \\
  $p \left( y | c_1 \left( \mathbf{x} \right)=y_3  \right)$ & 0.1& 0.15 & 0.75  \\
  $\pi_{1|3}$ & 0.1 & 0.25 & 1 
\end{tabular}
\hspace{0.4cm}
\begin{tabular}{cccc}
  $y$ & $\omega_1$ & $\omega_1$ & $\omega_3$ \\
  $p \left( y | c_2 \left( \mathbf{x} \right)=y_2  \right)$ & 0.2& 0.6 & 0.2  \\
  $\pi_{2|2}$ & 0.4 & 1 & 0.4 
\end{tabular}

\end{center}
\end{ex}


\subsection{Conjunctive aggregation of possibility distributions} 
\label{sub:conj_aggregation_of_possibility_distributions}
In classical logic, the logical conjunction operator (logical AND) is true when all operands are true. 
When the operands are subsets, it thus amounts to taking the intersection of these subsets. 
Logical conjunction allows us to combine possibility distributions that are basically indicator functions of a set (as in example \ref{ex:1}). 
When possibility values are in $]0;1[$ (as in example \ref{ex:2}), we need to define an aggregation operator that is a generalization of logical conjunction.

The parametric operator $\mathcal{T}_{\lambda}$ presented in \ref{sub:aggregation_of_possibility_distributions} and \ref{sub:adaptive_aggregation_w_r_t_dependency} that relies on Aczel-Alsina t-norms is one such generalization in the sense that if $\pi_1 \left( \omega \right)  = \mathbb{I} \left\{ \omega \in B \right\}$ and $\pi_2 \left( \omega \right)  = \mathbb{I} \left\{ \omega \in C \right\}$ then, $\mathcal{T}_{\lambda} \left( \pi_1,\pi_2 \right) \left( \omega \right)  = \mathbb{I} \left\{ \omega \in B\cap C \right\} $. 

Actually, the operator $\mathcal{T}_{\lambda}$ achieves a more general form of conjunctivity which reads

\begin{equation}
  \mathcal{T}_{\lambda} \left( \pi_1,\pi_2 \right) \preceq \pi_1 \text{ and } \pi_2, \forall \pi_1,\pi_2.\label{eq:conj}
\end{equation}

This means that the result of the aggregation of possibility distributions using $\mathcal{T}_{\lambda}$ is always at least as informative than the most informative input distribution. This property comes from the fact that for any distributions $\pi_1$ and $\pi_2$ and any $\lambda < \lambda'$, we have 

$$\mathcal{T}_{\lambda} \left( \pi_1,\pi_2 \right) \preceq \mathcal{T}_{\lambda'} \left( \pi_1,\pi_2 \right). $$

Remembering that $\mathcal{T}_{\infty}\left( \pi_1,\pi_2 \right)$ is the entrywise minimum of $\pi_1$ and $\pi_2$, it is obvious that $\mathcal{T}_{\infty}\left( \pi_1,\pi_2 \right) \preceq \pi_1 \text{ and } \pi_2$ and by transitivity \eqref{eq:conj} holds. 

\begin{ex}\label{ex:3}
Suppose $\Omega=\left\{ \omega_1; \omega_2; \omega_3\right\}$. Below is a figure displaying the aggregated possibility distribution $\pi_{\text{ens}}$ using $\mathcal{T}_{\lambda}$ when one operand is distribution $\pi_{1|3}$ from example \ref{ex:2}, a second operand is distribution $\pi_{2|2}$ from example \ref{ex:2} and the final and third operand is $\pi_{3|2}=\pi_{2|2}$.

\begin{center}
\begin{figure}[h!]
  \centering
  \includegraphics[width=.7\textwidth]{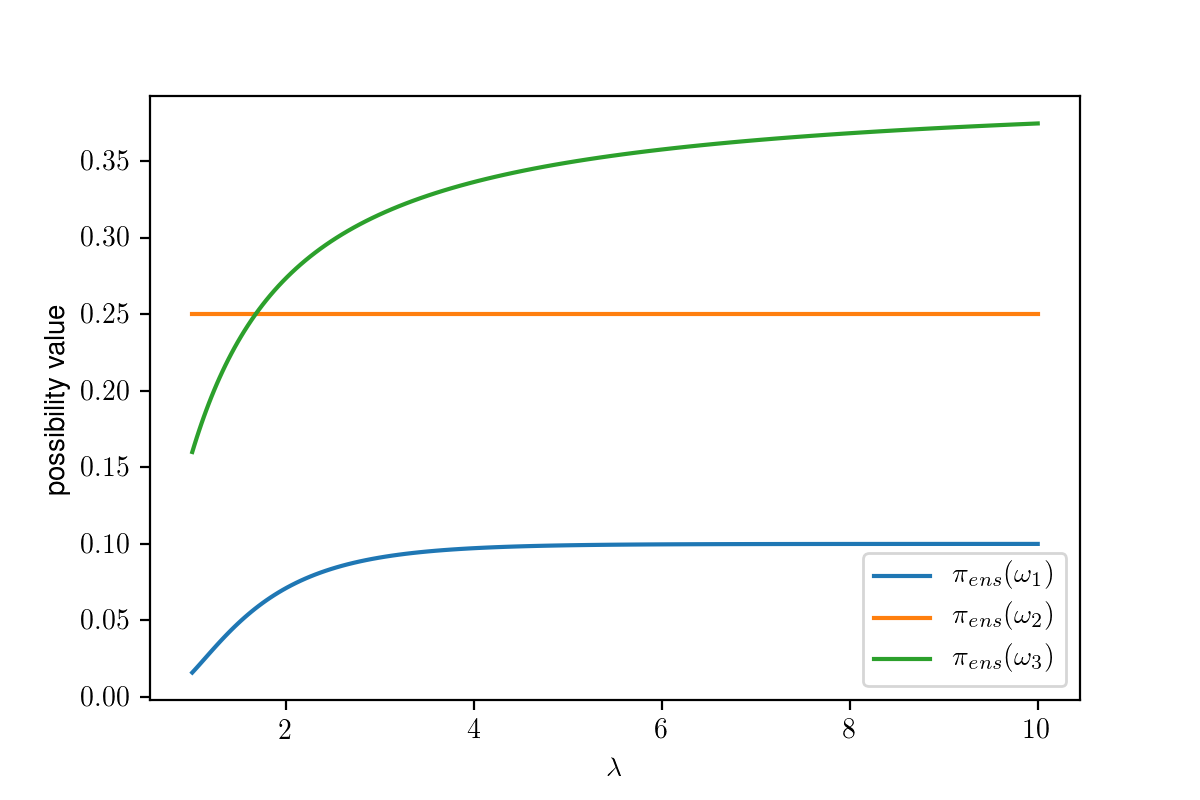}
  \caption{Evolution of possibility values of $\pi_{\text{ens}}$ w.r.t. to $\lambda$}
\end{figure}


\end{center}
\end{ex}

The above example illustrates that depending on the value of $\lambda$, the class label achieving maximal possibility is either a class label strongly supported by one classifier or moderately supported by a majority of classifiers. Since $\lambda$ is tuned by a grid search based heuristic in adaSPOCC, the chosen aggregation policy is data driven and will maximize accuracy.

The key aspect of possibility theory that we exploit in the derivation of adaSPOCC is that the above described aggregation operator is flexible and has a number of parameters to learn that is linear in $K$ the number of classifiers in the ensemble. In probability theory, aggregating $p \left( y | c_1 \right) $ and $p \left( y | c_2 \right) $ consists in inferring the distribution $p \left( y | c_1,c_2 \right)$. As explained earlier and illustrated by the numerical experiments, direct inference leads to overfitting or is simply intractable. When simplifying conditional independence assumptions are made, the computations become tractable at the expense of accuracy. 

In possibility theory, the philosophy behind aggregation operators such as $\mathcal{T}_{\lambda}$ is fundamentally different. Aggregation operators are not derived through calculus rules but by logical rules. Desirable properties are stated and an operator satisfying them is derived (normative approach). This is a knowledge based system view, in which possibility distributions encode partial knowledge that must be combined in a principled way. 


\subsection{Decision making from possibilities} 
\label{sub:decision_making_from_possibilities}
It is clear that based on the aggregated possibility distribution $\pi_{\text{ens}}$ obtained from the ensemble of classifiers using Algorithm \ref{spocc-test} or \ref{adaspocc-test}, the intuition calls for taking the argmax of this possibility distribution in order to produce a reliable prediction. 

There are actually strong theoretical justifications behind this decision rule coming from decision theory. Indeed, a theorem by \cite{GILBOA198765} explains that decisions based on non-additive measures (such as possibility measures) are in line with several decision theoretic axioms. These axioms are closely related to those of the famous theorem from \cite{savage1954foundations} which holds for (finitely) additive probability distributions.

\section{Normalized confusion matrices on real data} 
\label{sec:normalized_confusion_matrices_on_real_data}

This section provides normalized confusion matrices corresponding to the experiments on real datasets exposed in \ref{sub:real_data} and summarized in Table \ref{tab:real3}. 
The matrices are grouped by dataset and for each of the 8 datasets there is one such matrix for each aggregation or reference method. 
The confusion matrix of the Bayes aggregation is missing for 20newsgroup, Drive and MNIST datasets because Bayes aggregation is intractable as soon as $\ell>6$ with our equipment.

\newpage
\paragraph{20newsgroup}
\begin{center}
  \includegraphics[width=.99\textwidth]{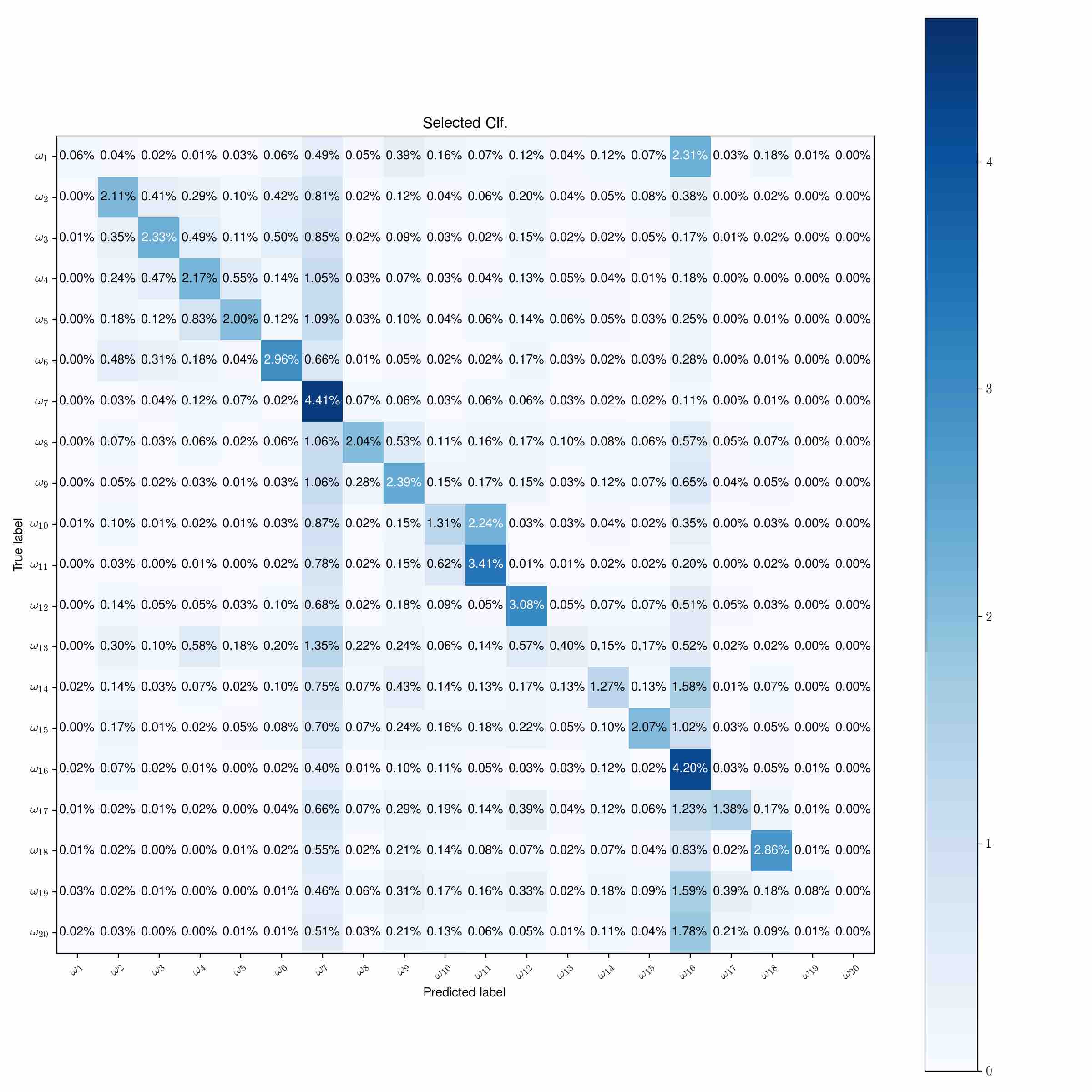} 

  \includegraphics[width=.99\textwidth]{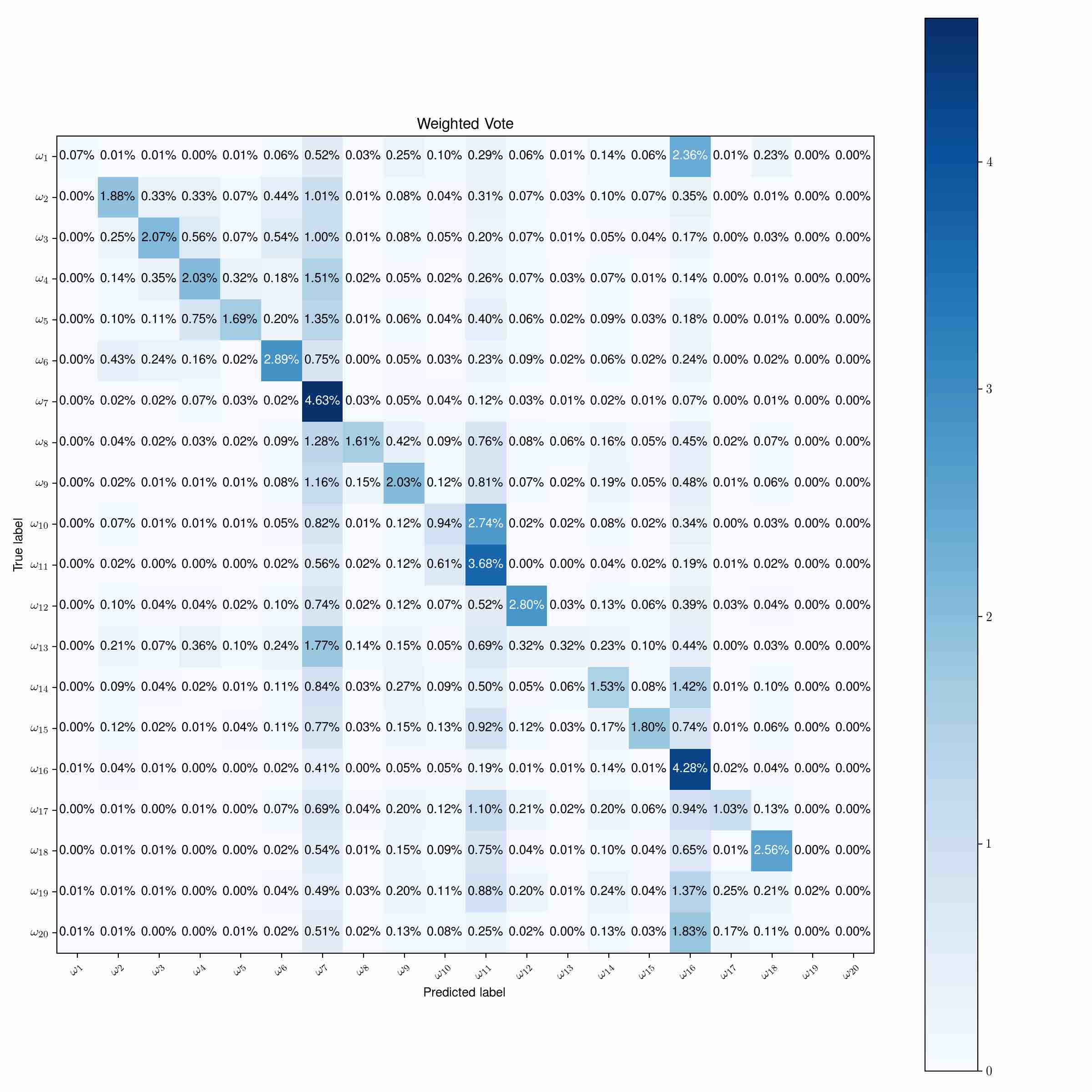}

  \includegraphics[width=.99\textwidth]{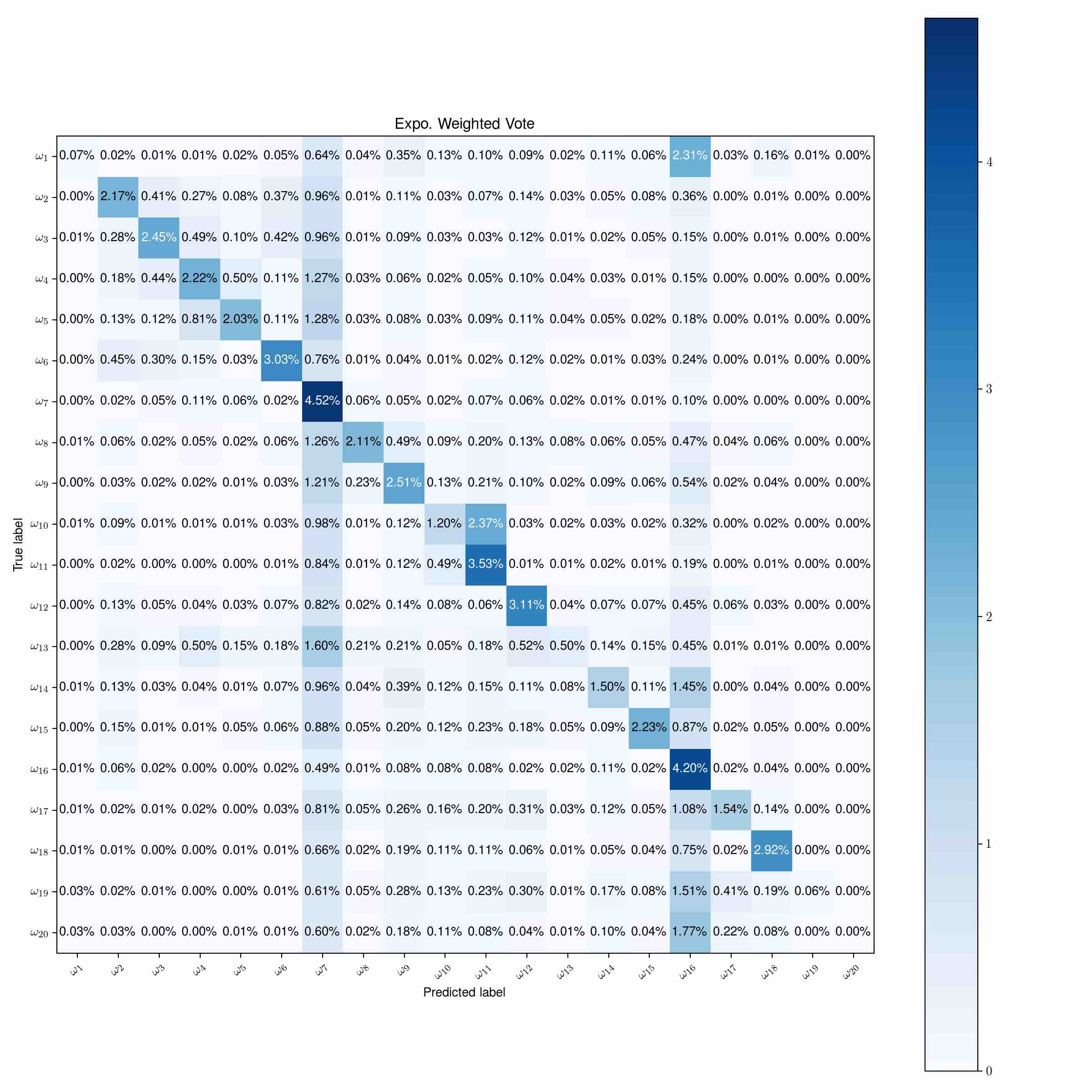} 

  \includegraphics[width=.99\textwidth]{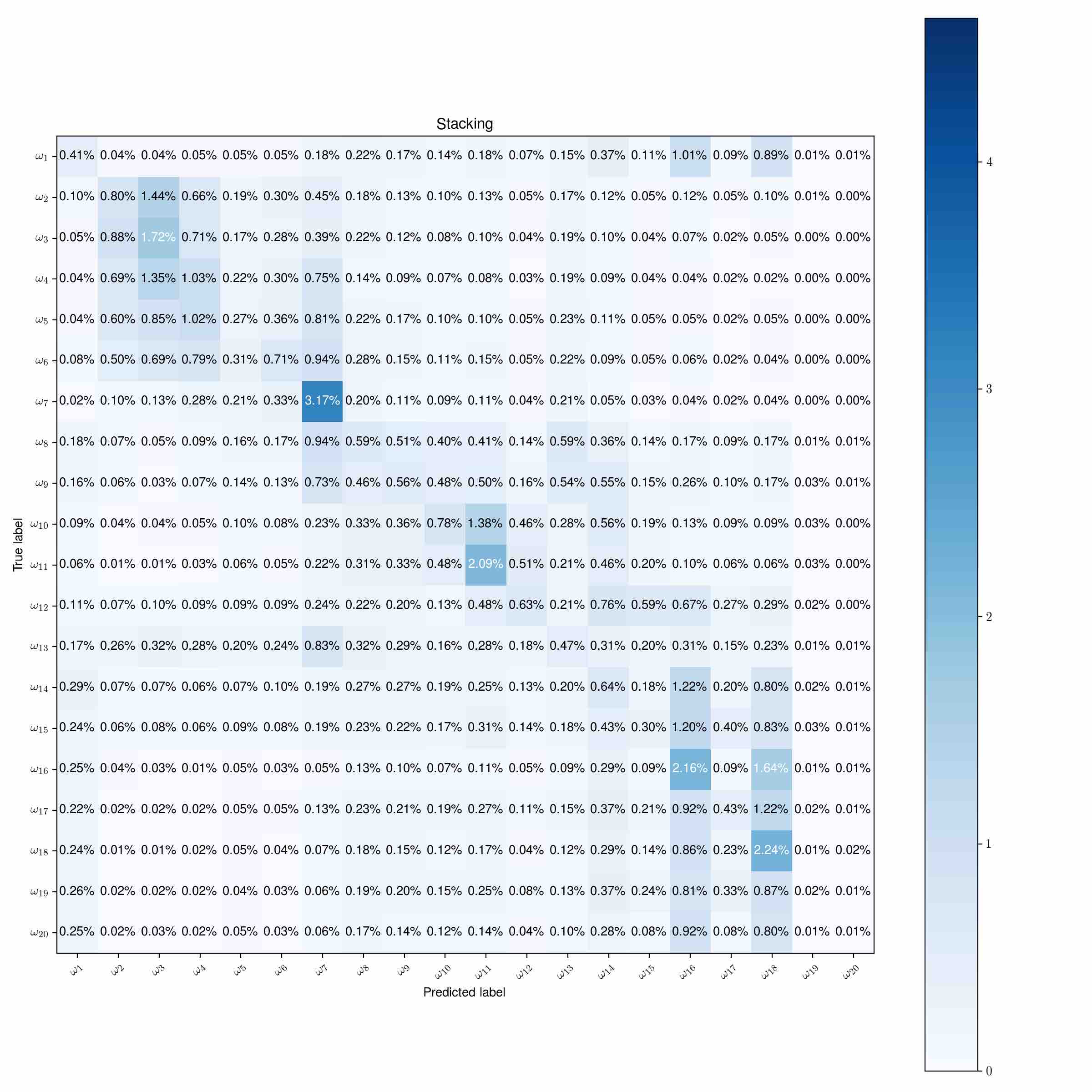} 

  \includegraphics[width=.99\textwidth]{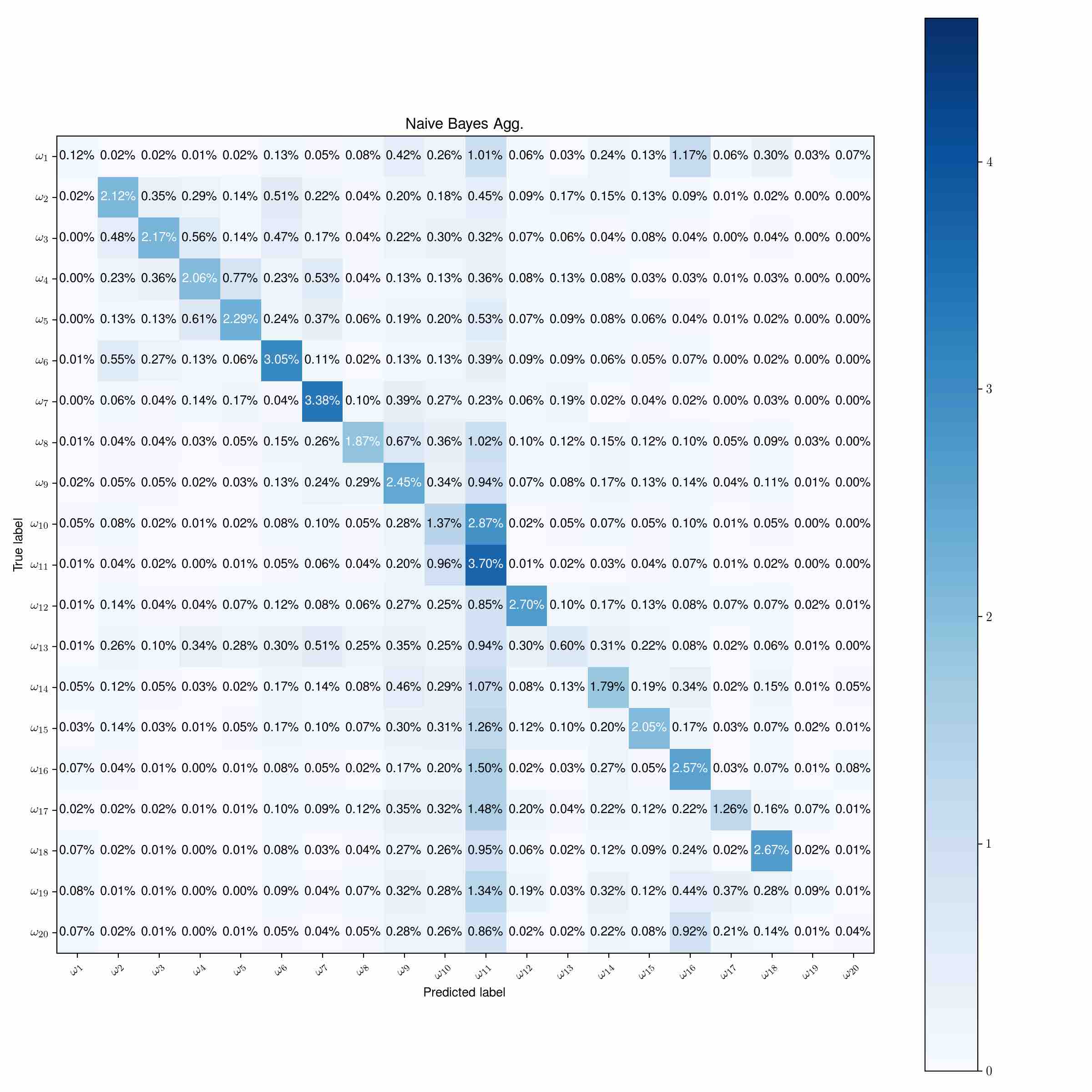} 

  \includegraphics[width=.99\textwidth]{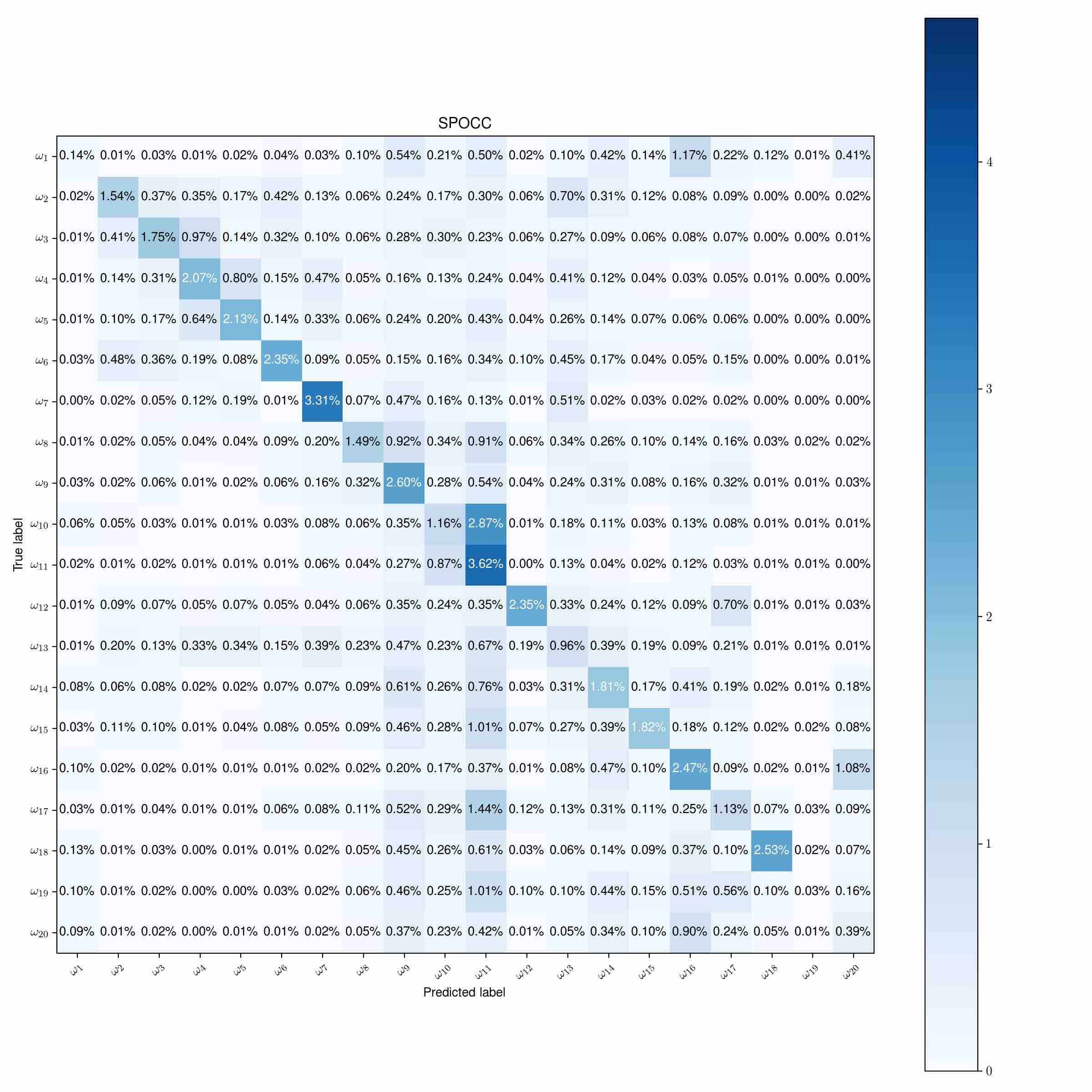}

  \includegraphics[width=.99\textwidth]{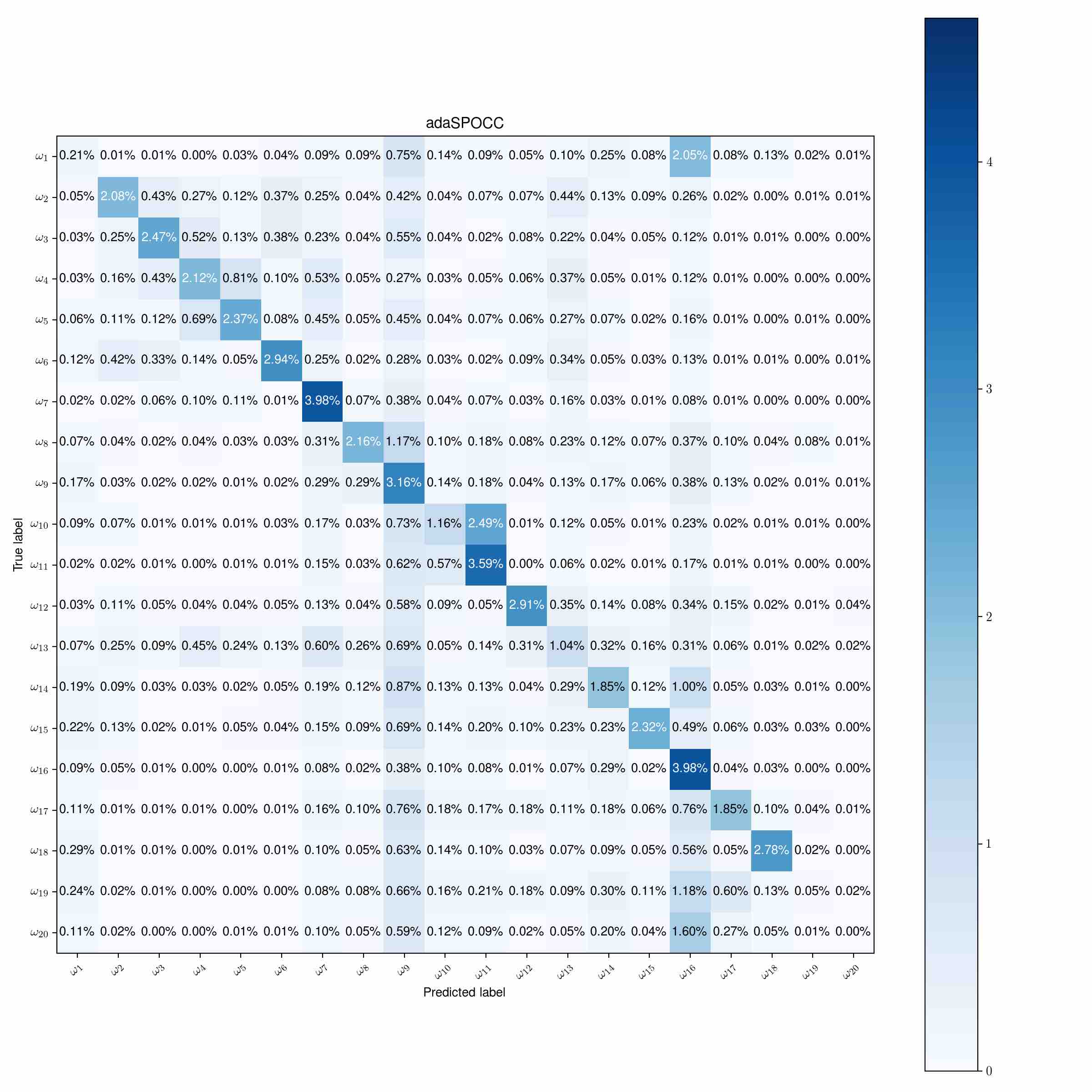} 

  \includegraphics[width=.99\textwidth]{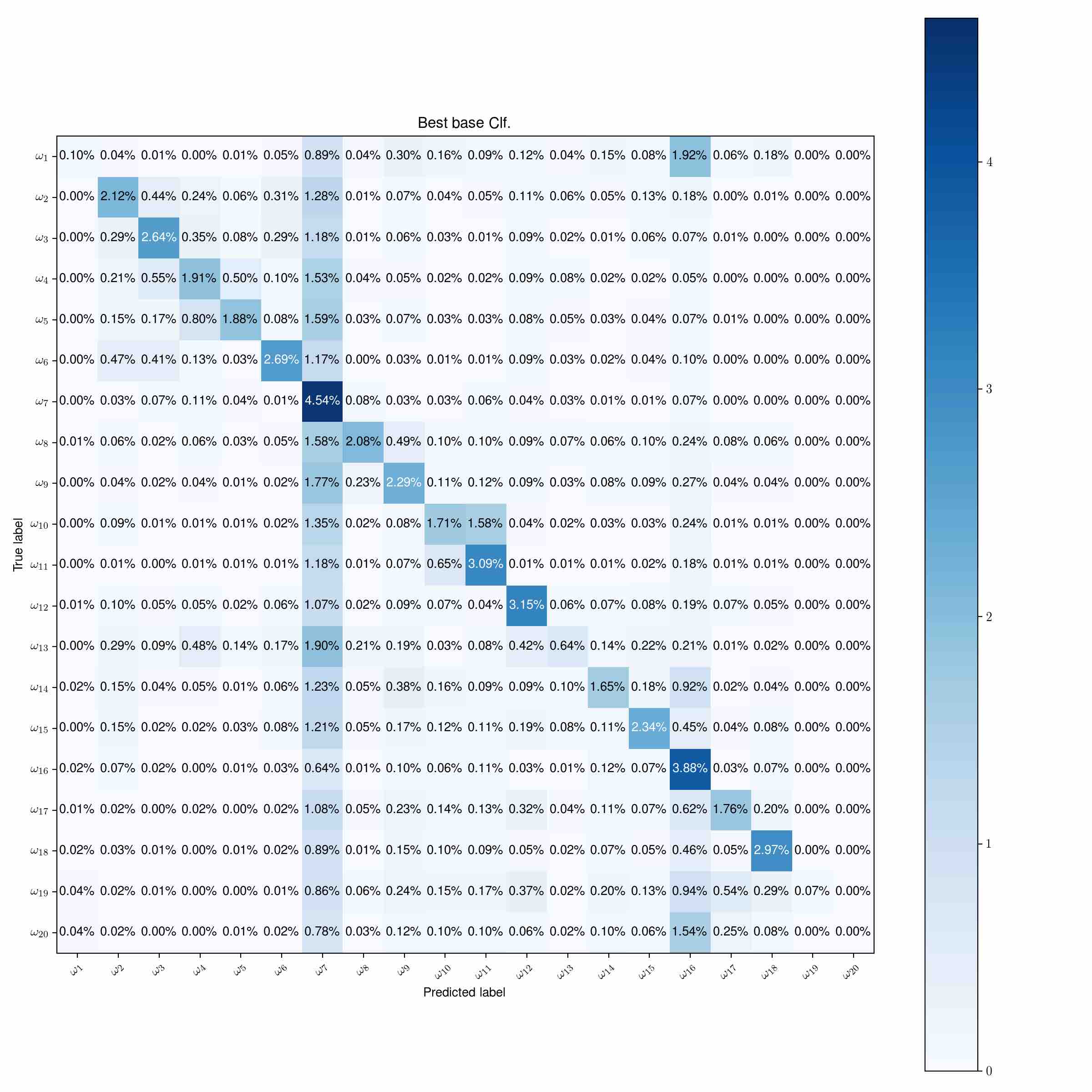} 

  \includegraphics[width=.99\textwidth]{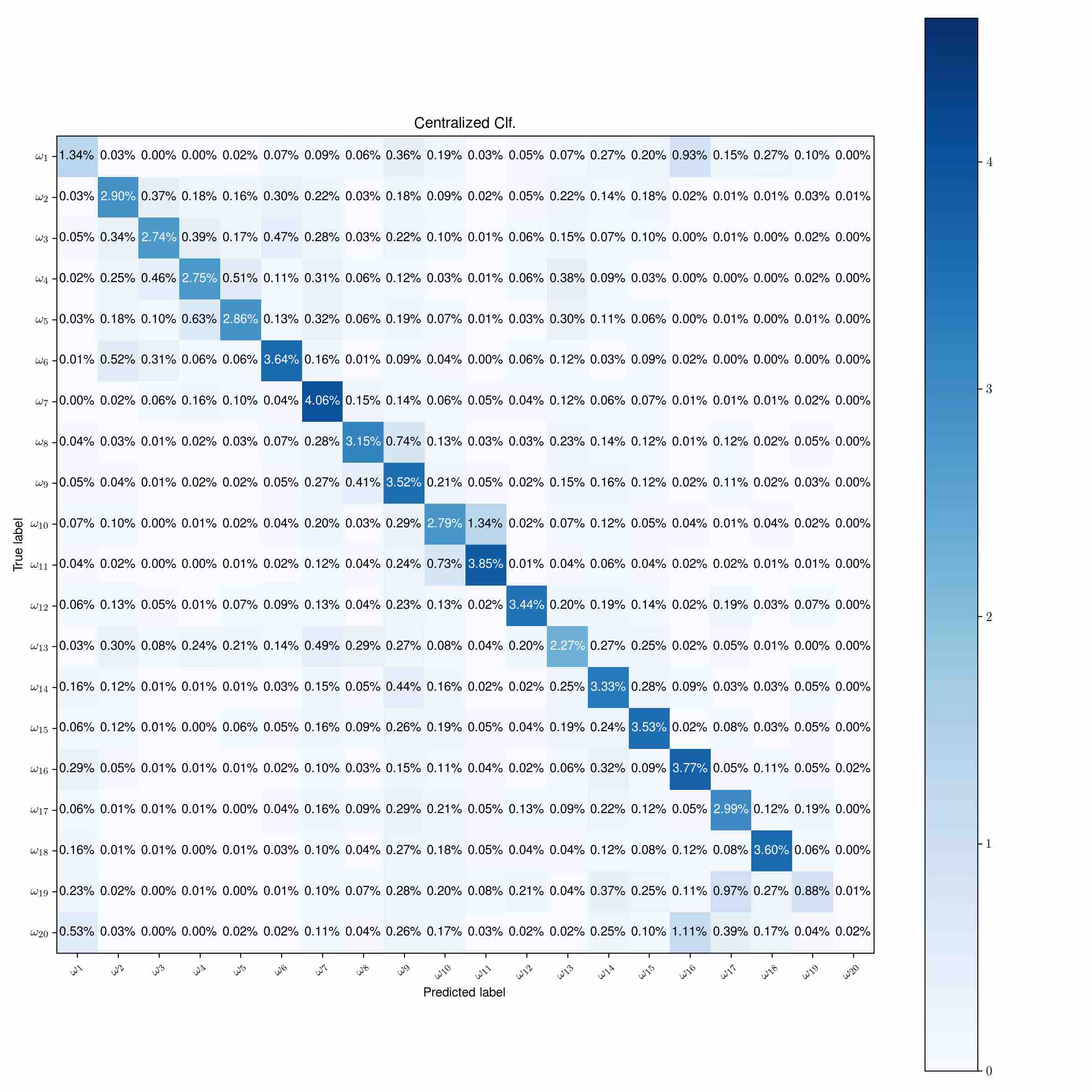} 
\end{center}

\paragraph{MNIST}
\begin{center}
  \includegraphics[width=.60\textwidth]{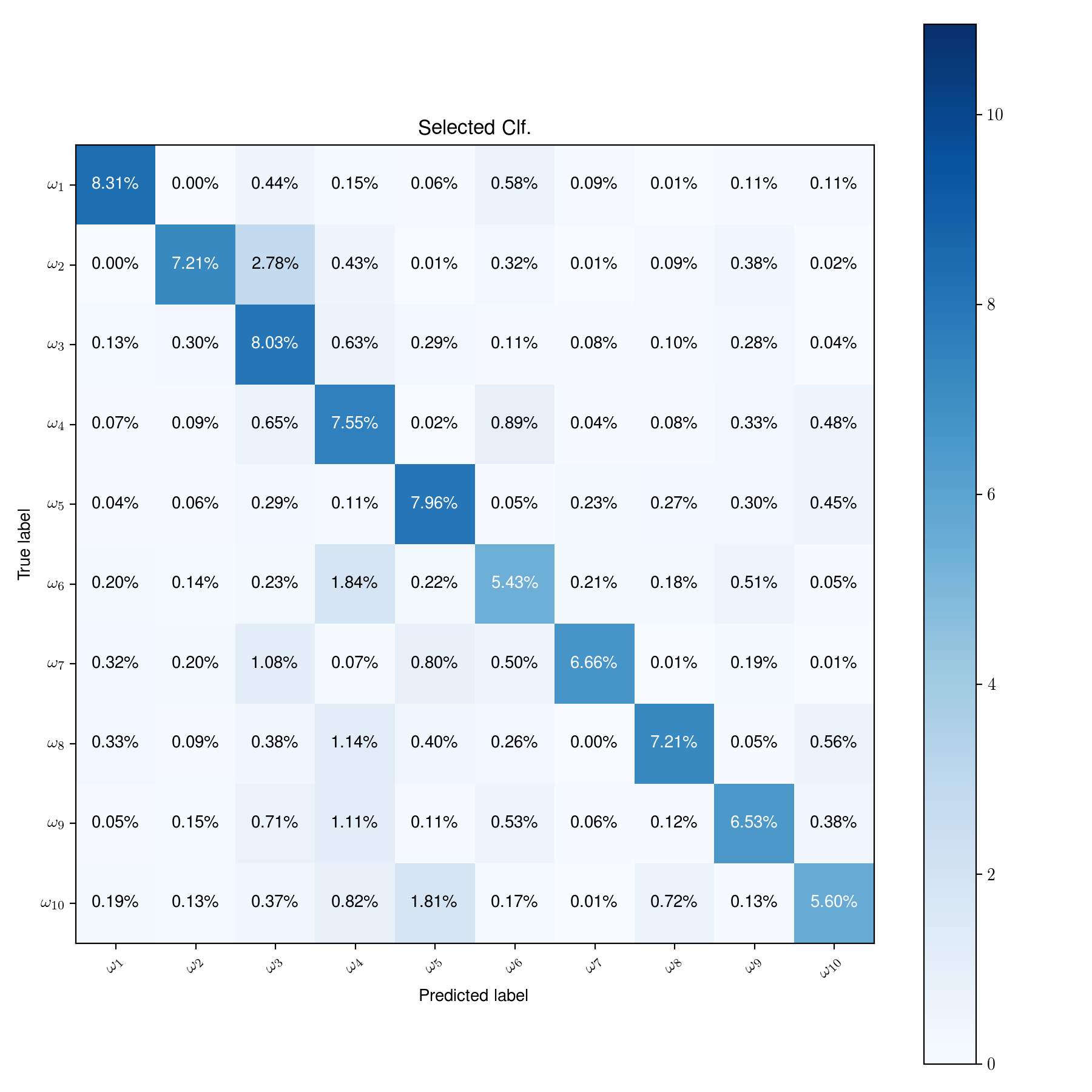} 

  \includegraphics[width=.60\textwidth]{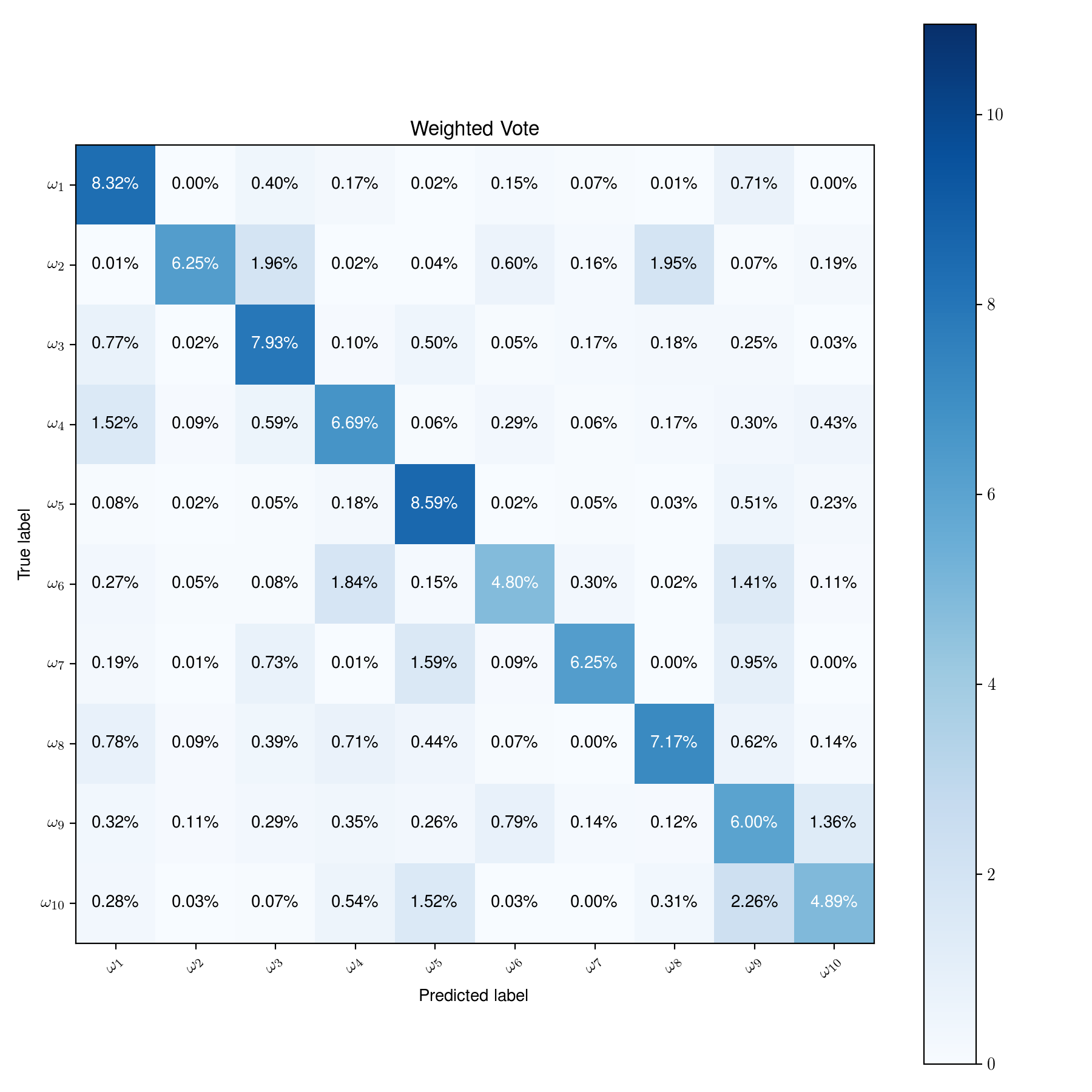}

  \includegraphics[width=.60\textwidth]{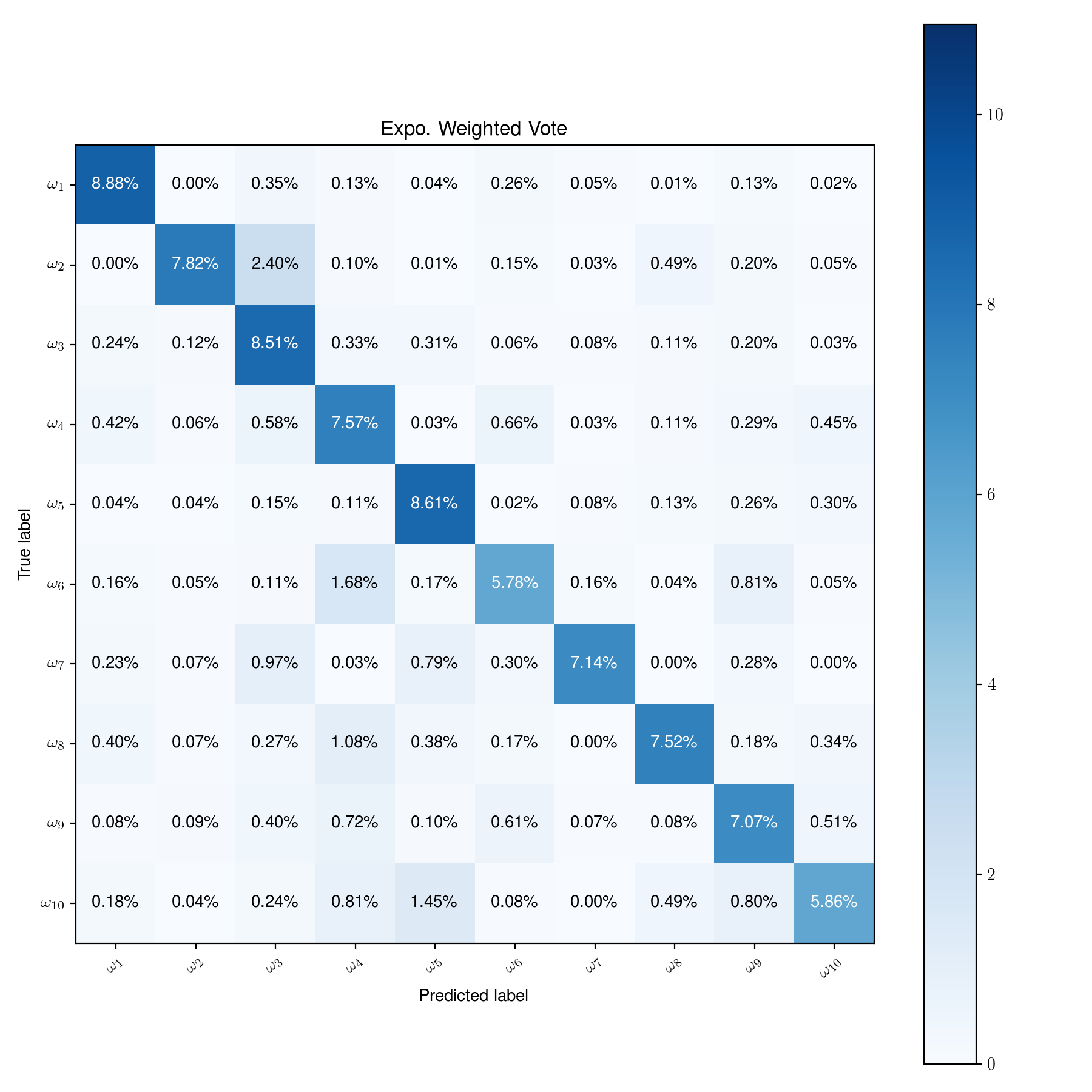} 

  \includegraphics[width=.60\textwidth]{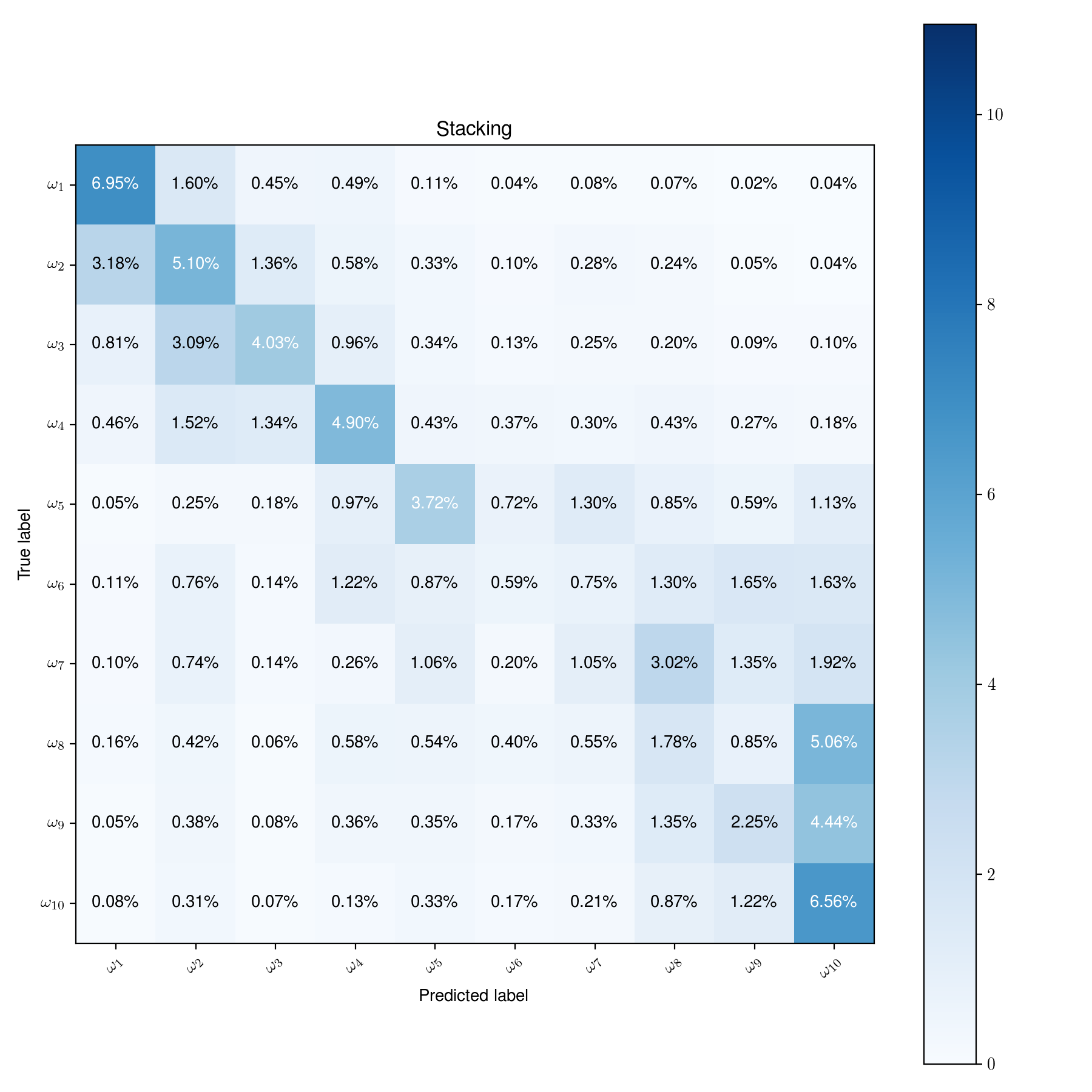} 

  \includegraphics[width=.60\textwidth]{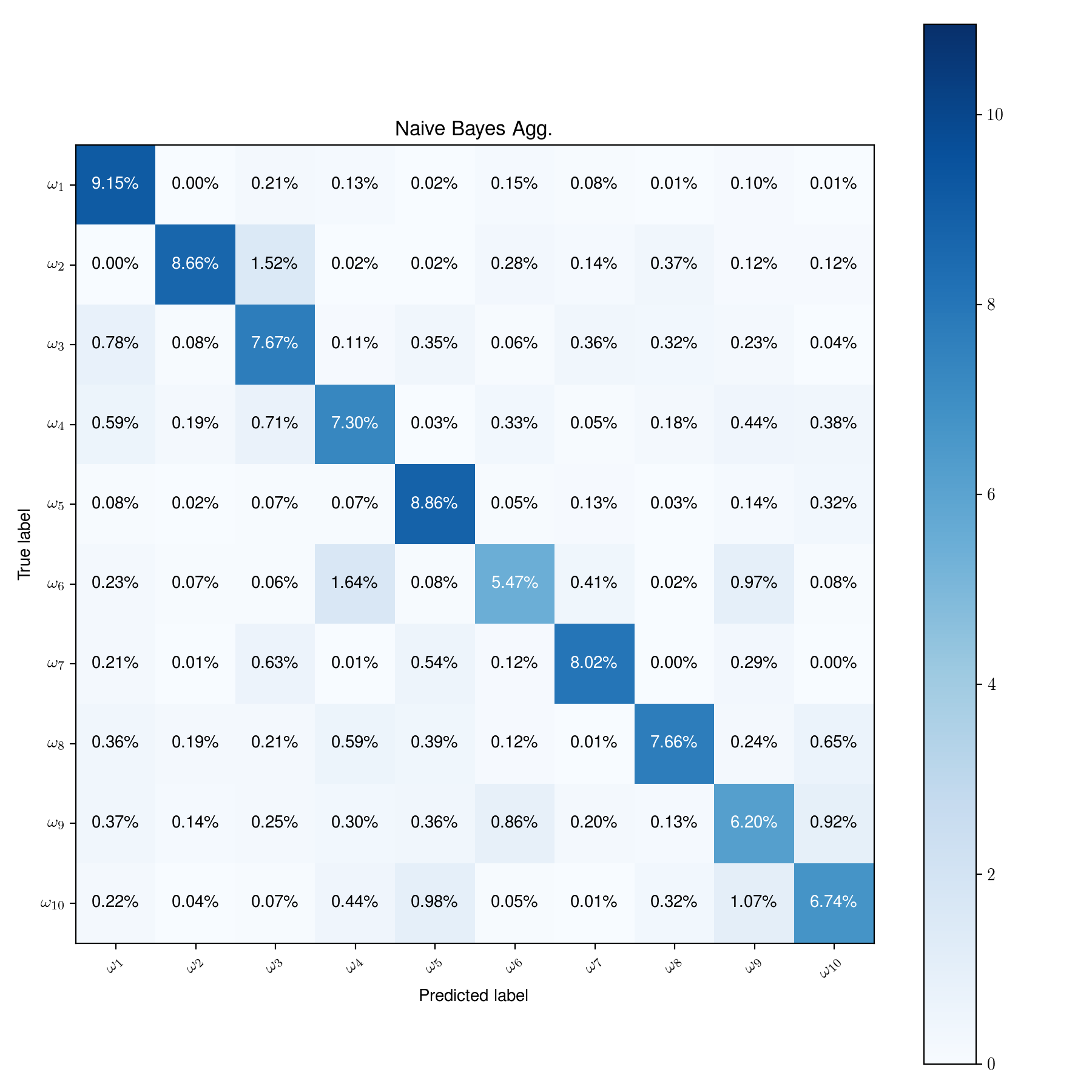} 

  \includegraphics[width=.60\textwidth]{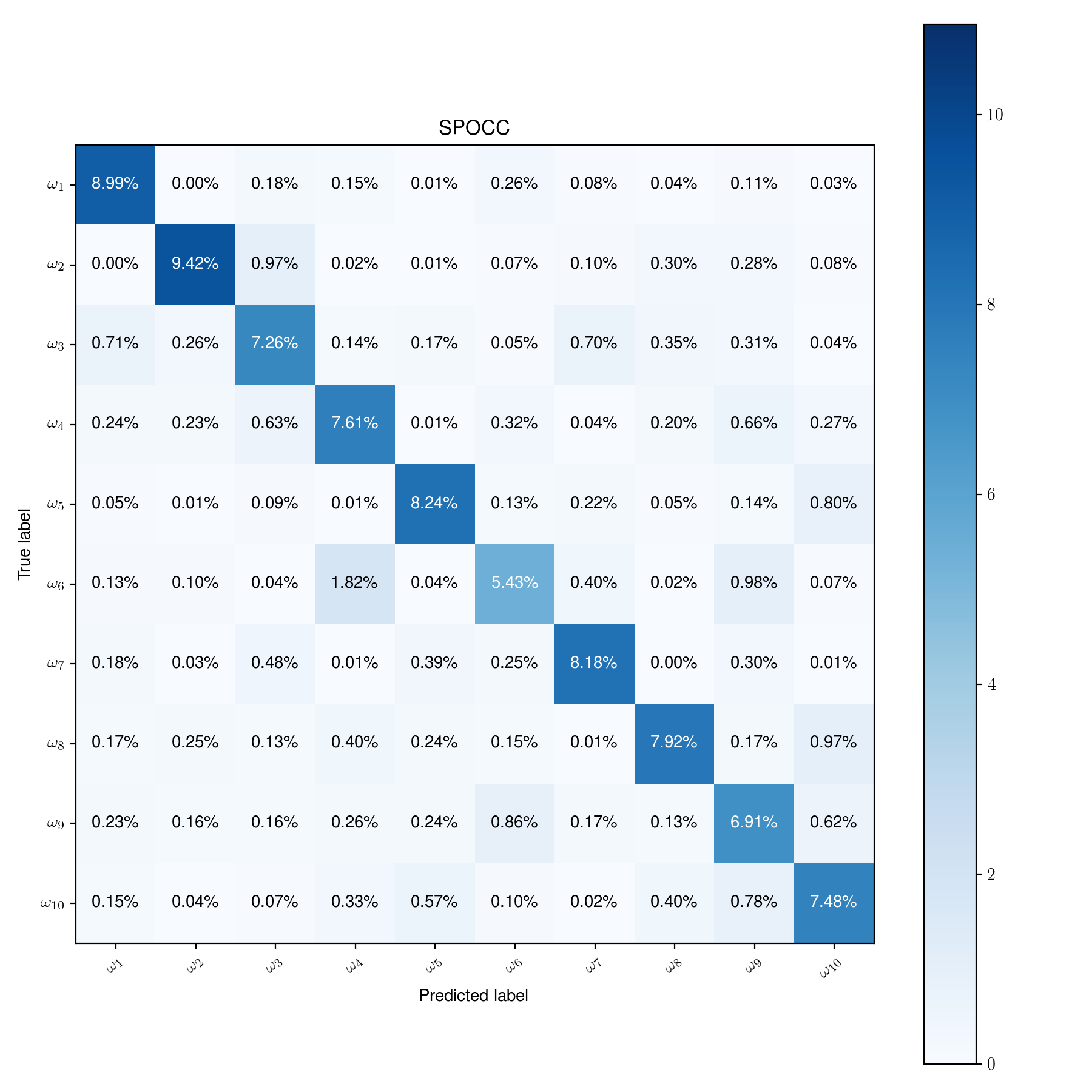}

  \includegraphics[width=.60\textwidth]{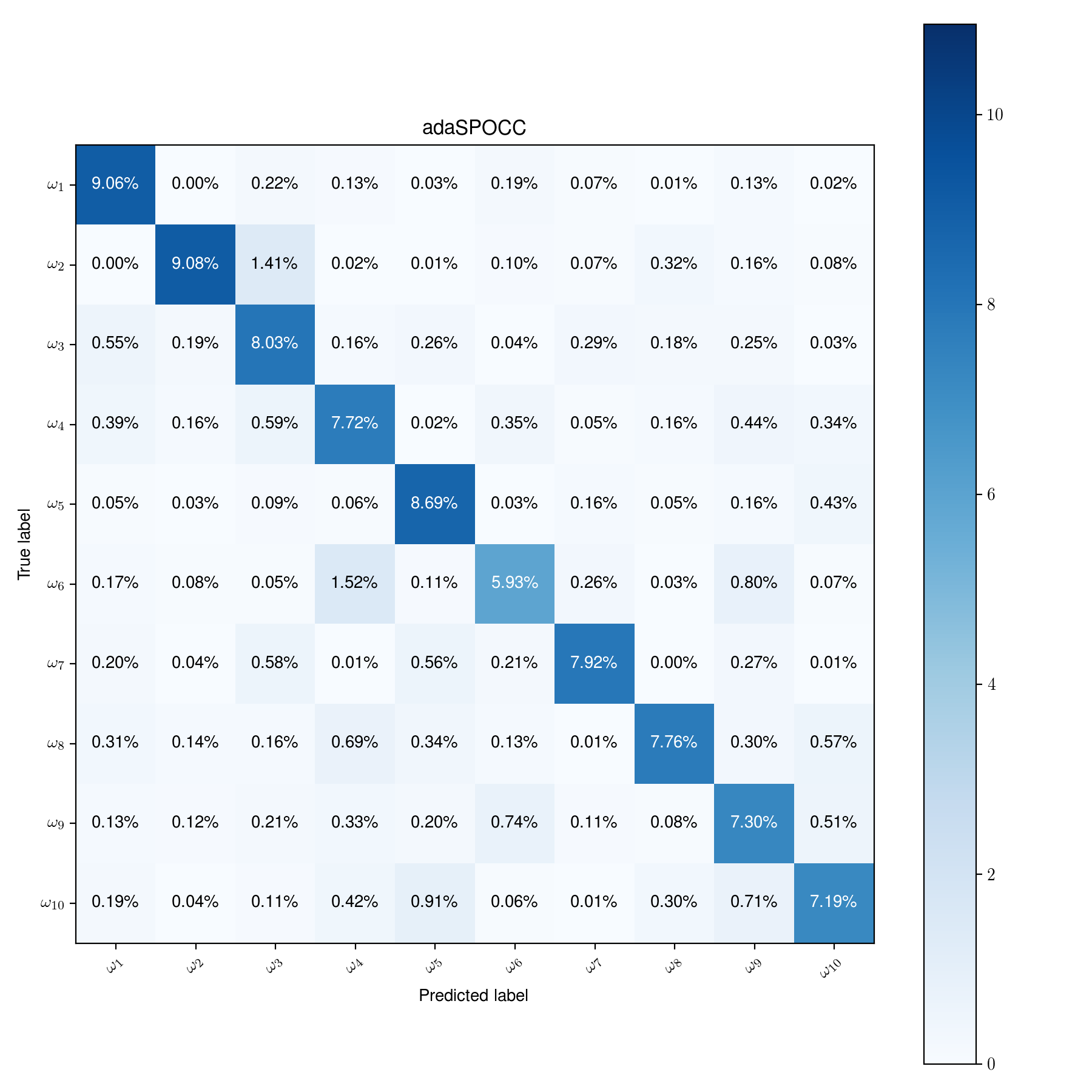} 

  \includegraphics[width=.60\textwidth]{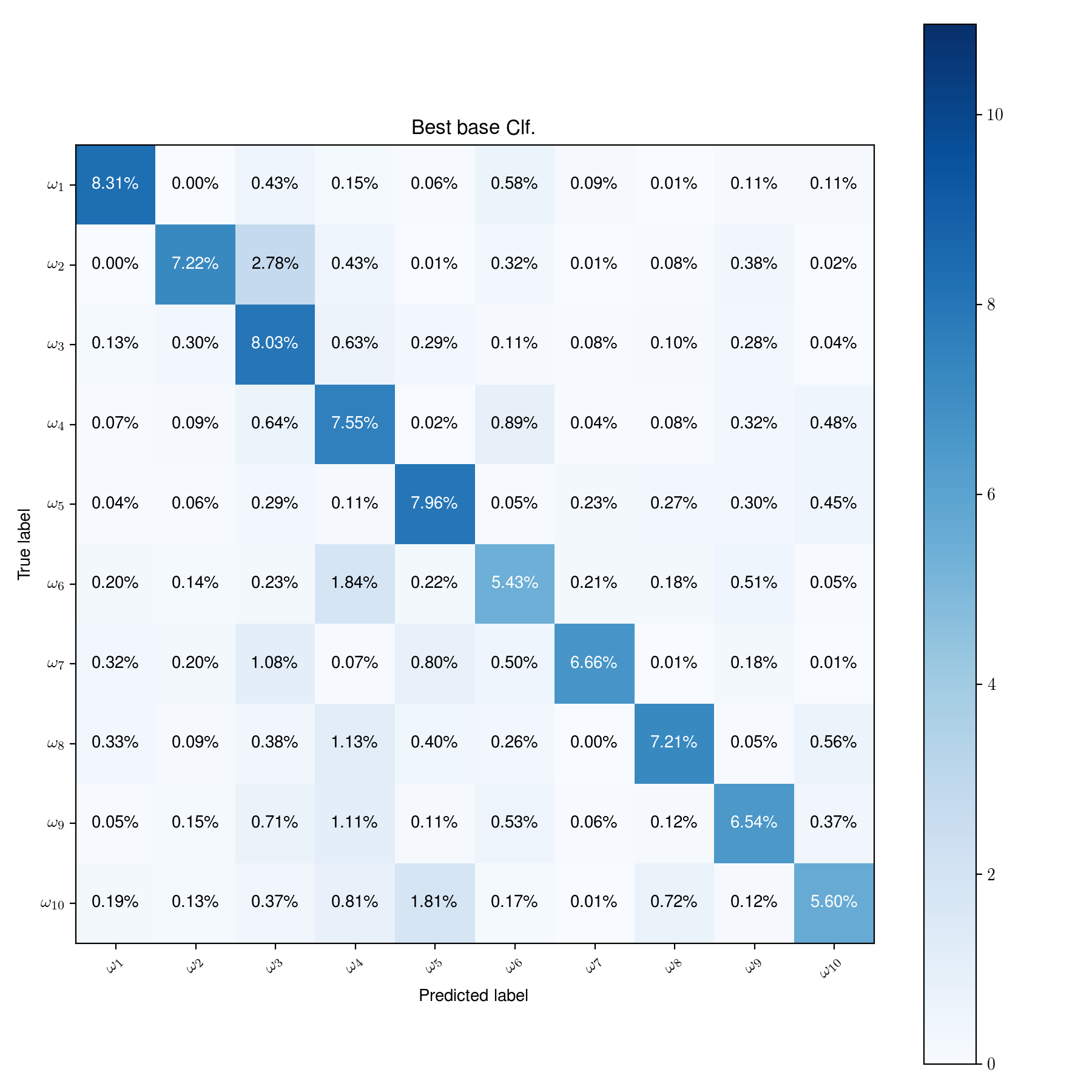} 

  \includegraphics[width=.60\textwidth]{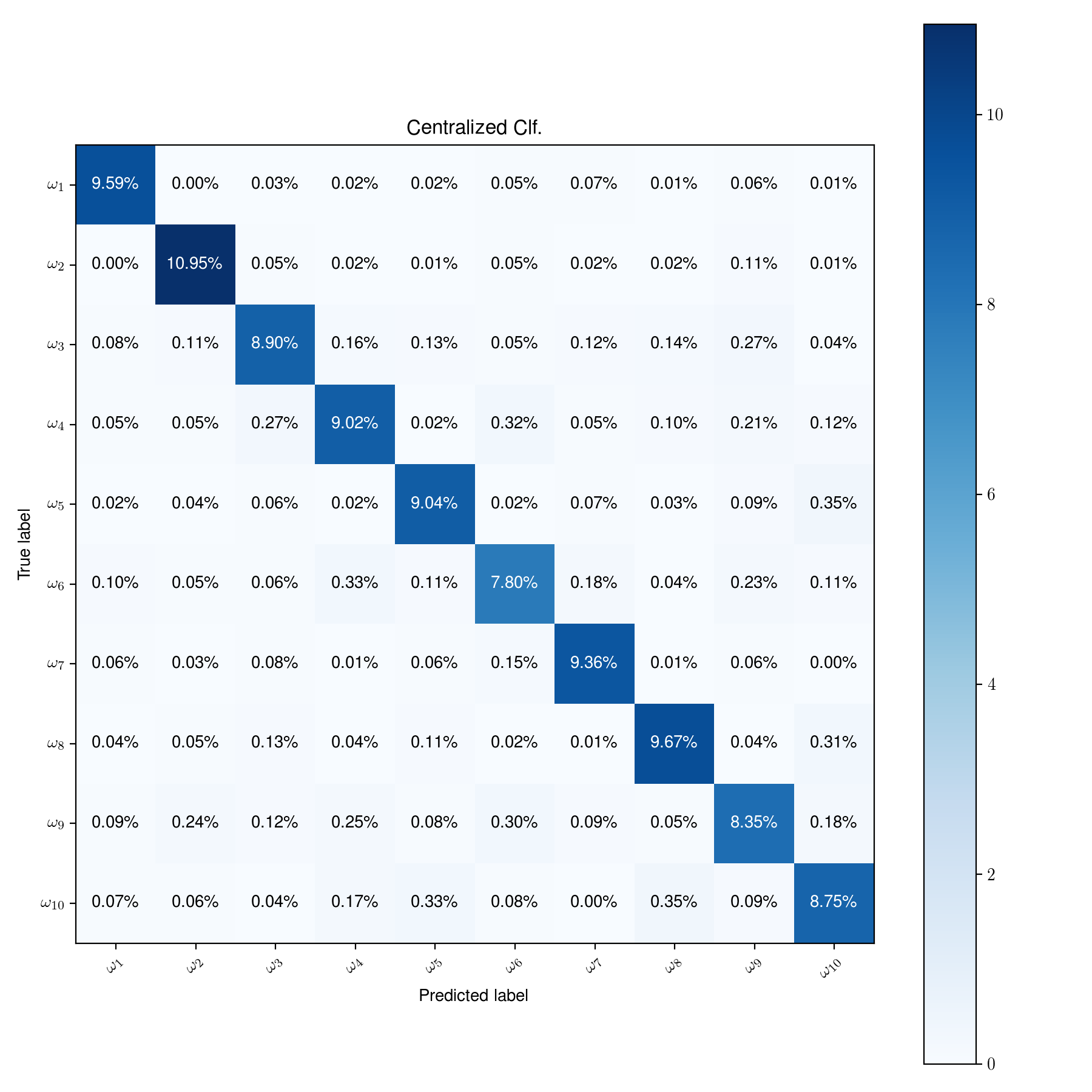} 
\end{center}

\newpage
\paragraph{Satellite}
\begin{center}
  \includegraphics[width=.49\textwidth]{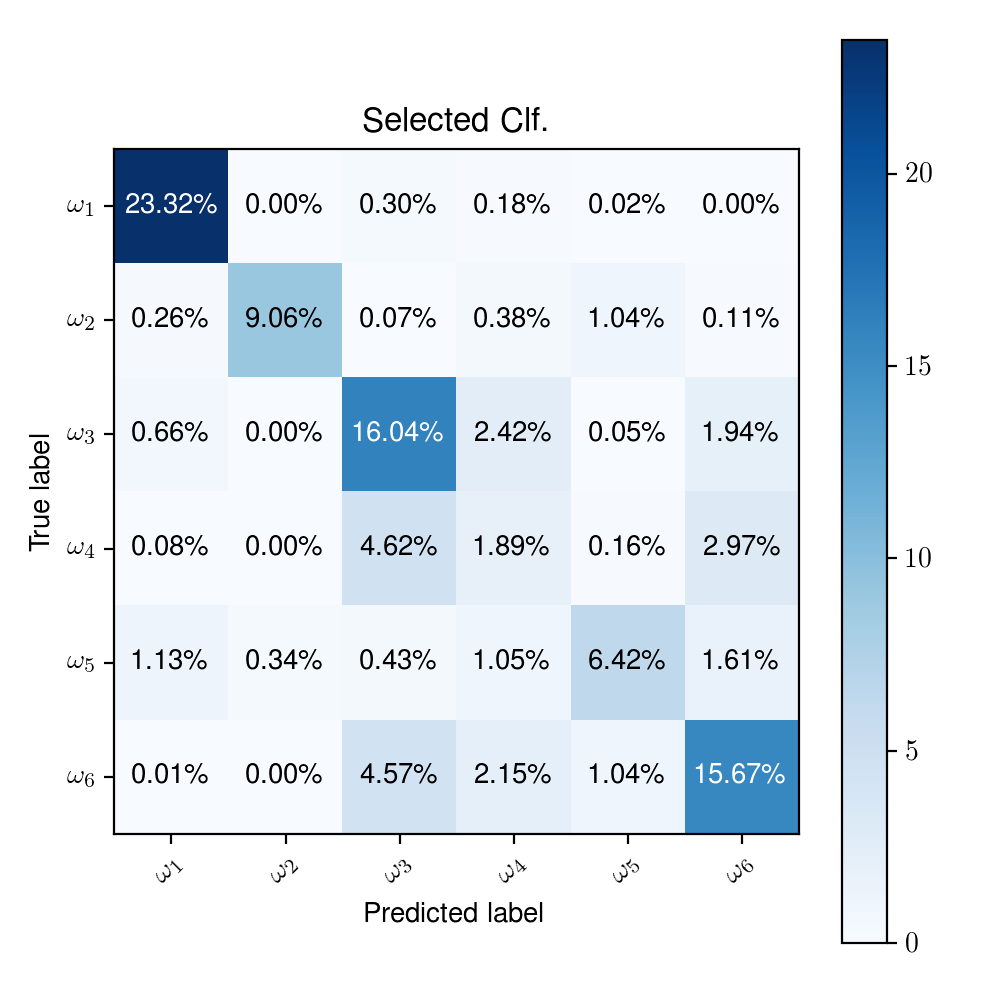} 
  \includegraphics[width=.49\textwidth]{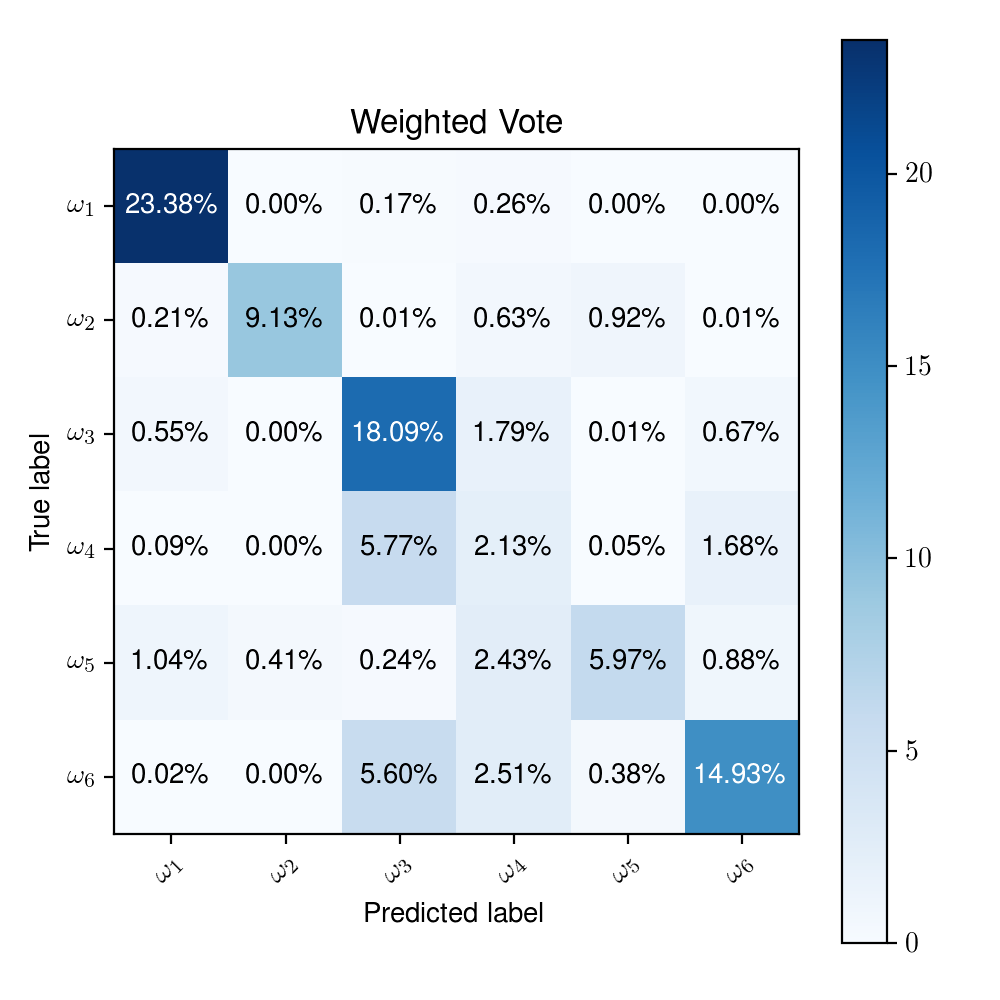}

  \includegraphics[width=.49\textwidth]{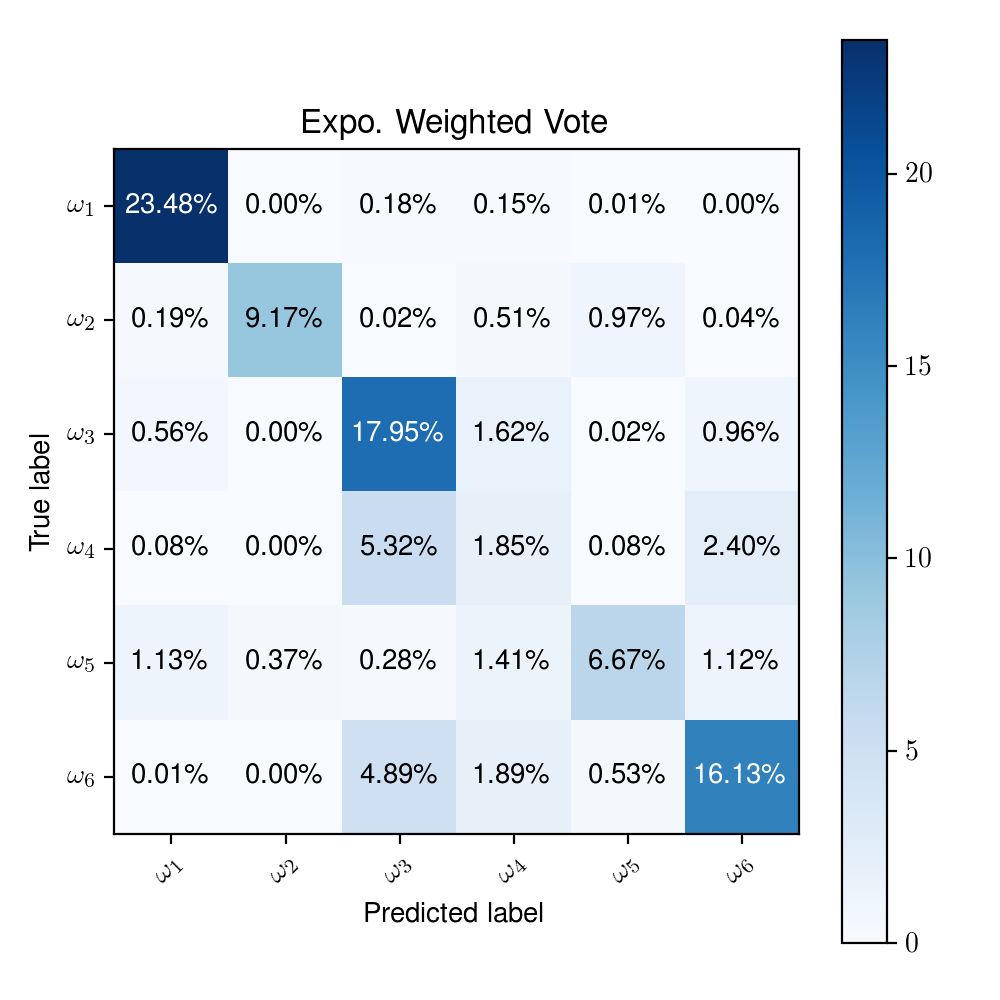} 
  \includegraphics[width=.49\textwidth]{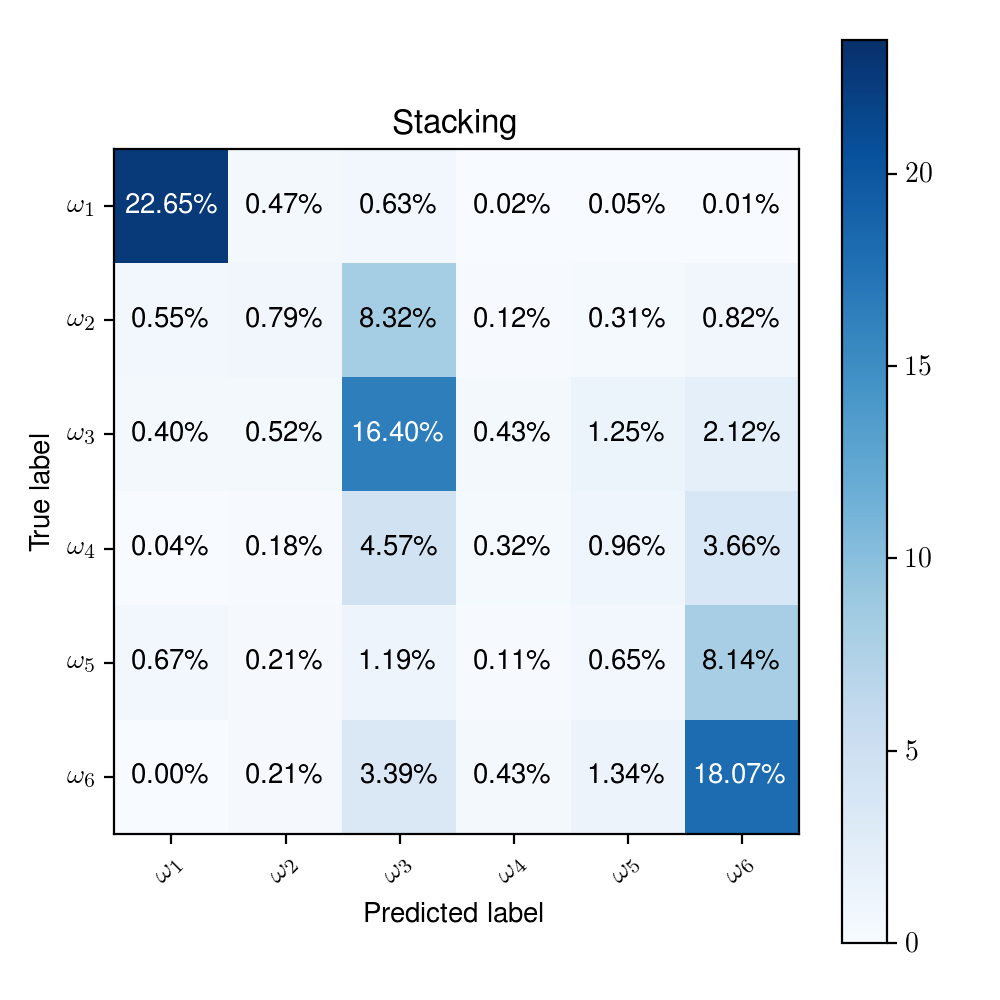} 

  \includegraphics[width=.49\textwidth]{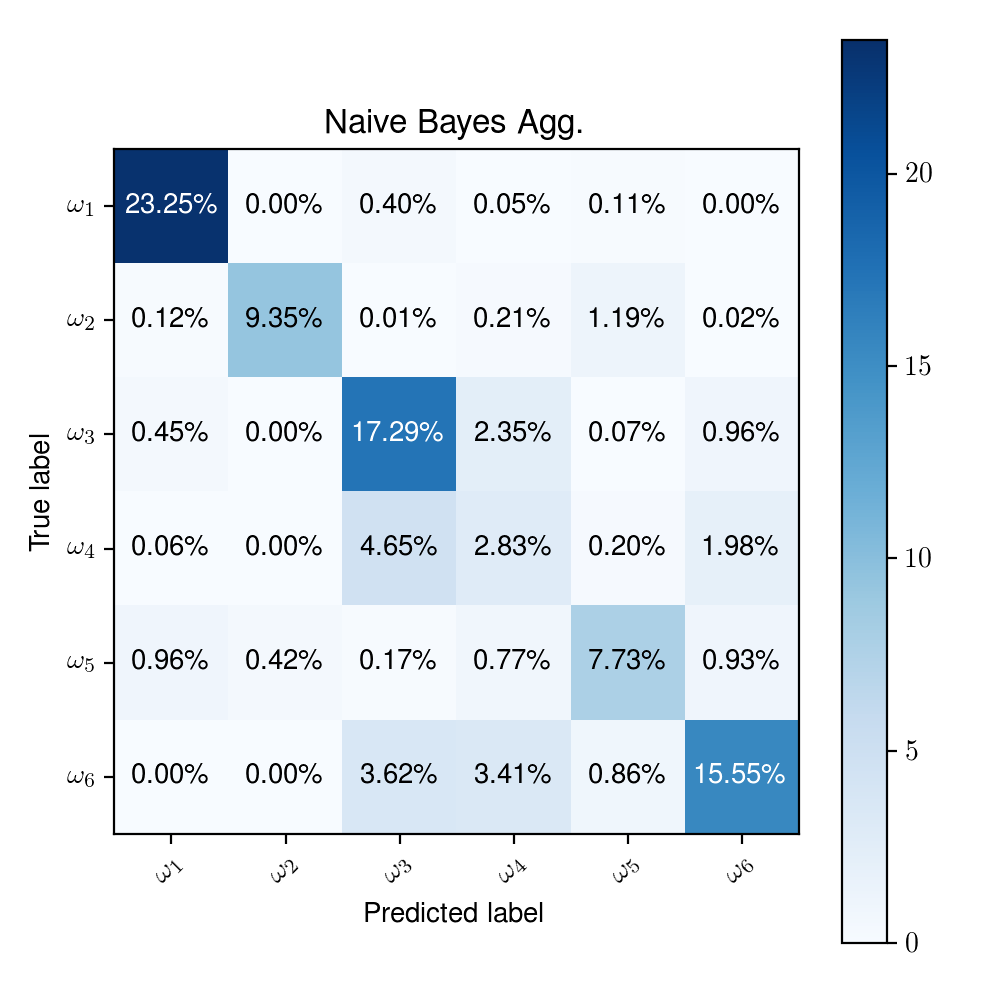} 
  \includegraphics[width=.49\textwidth]{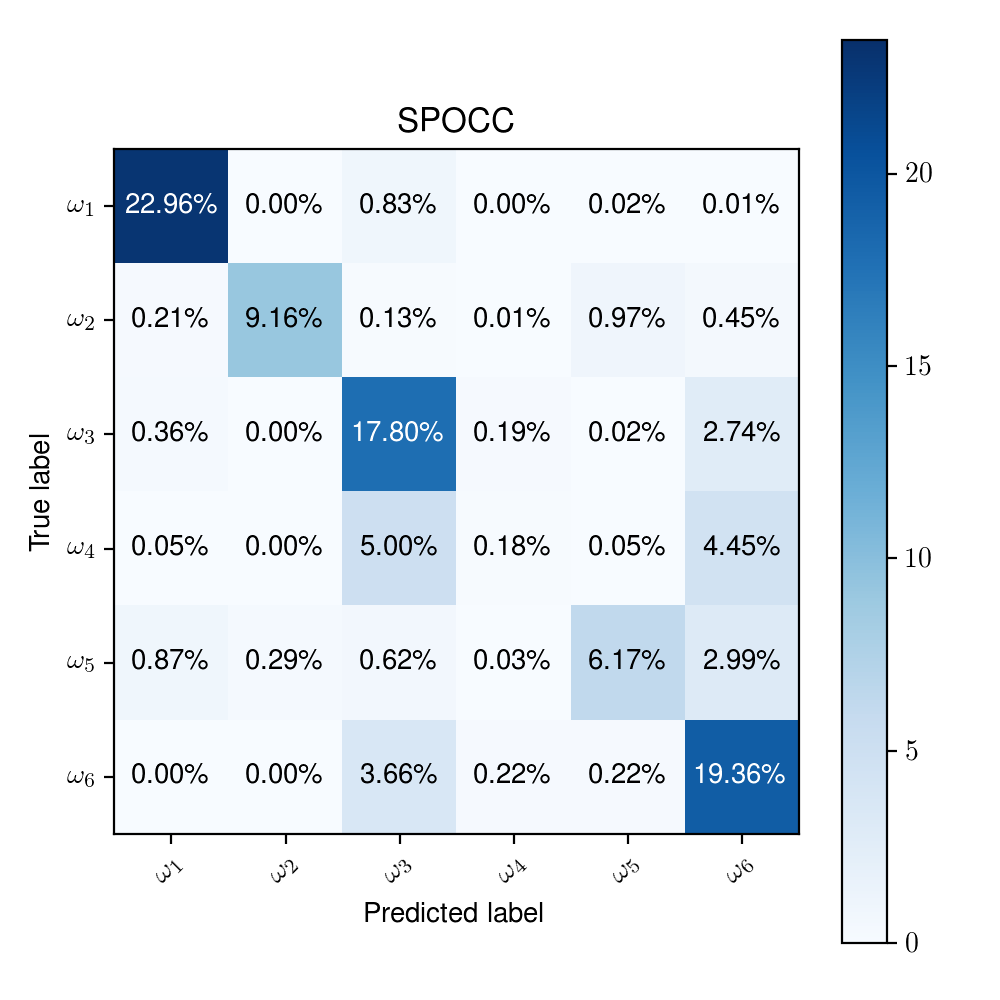}

  \includegraphics[width=.49\textwidth]{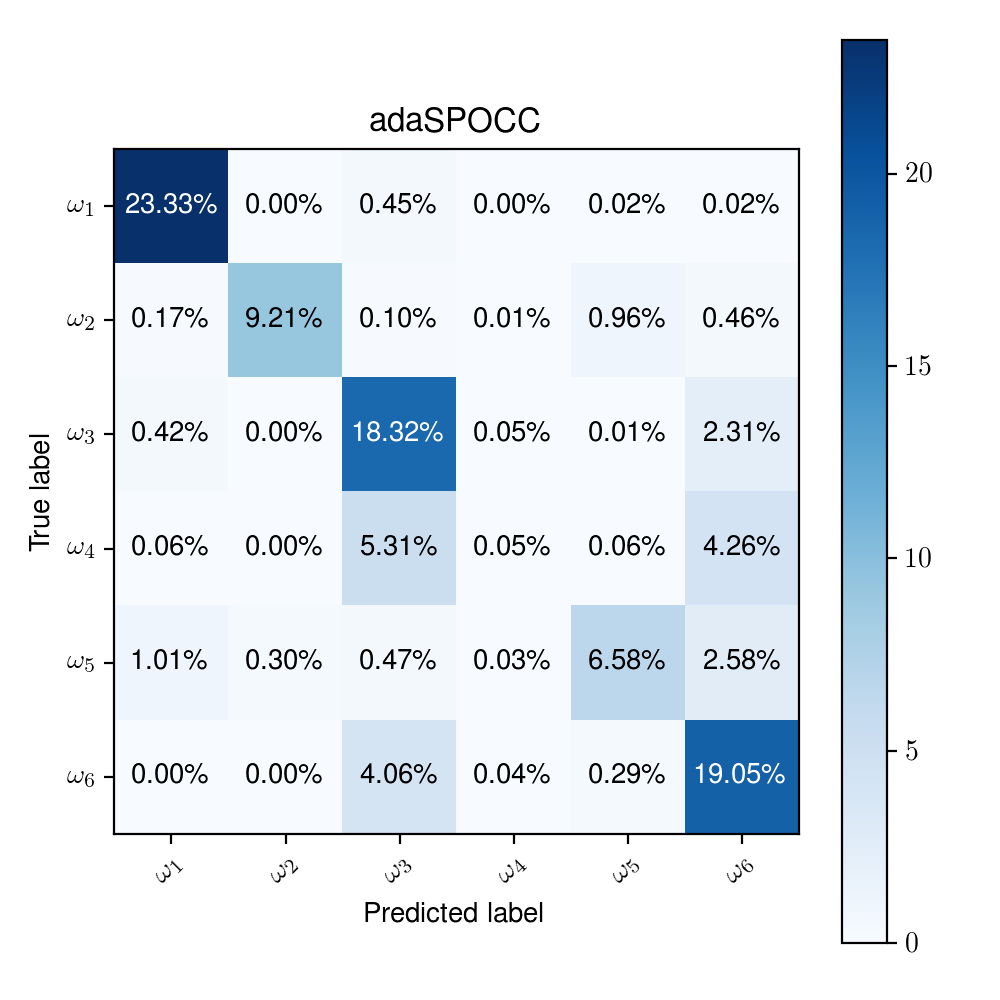} 
  \includegraphics[width=.49\textwidth]{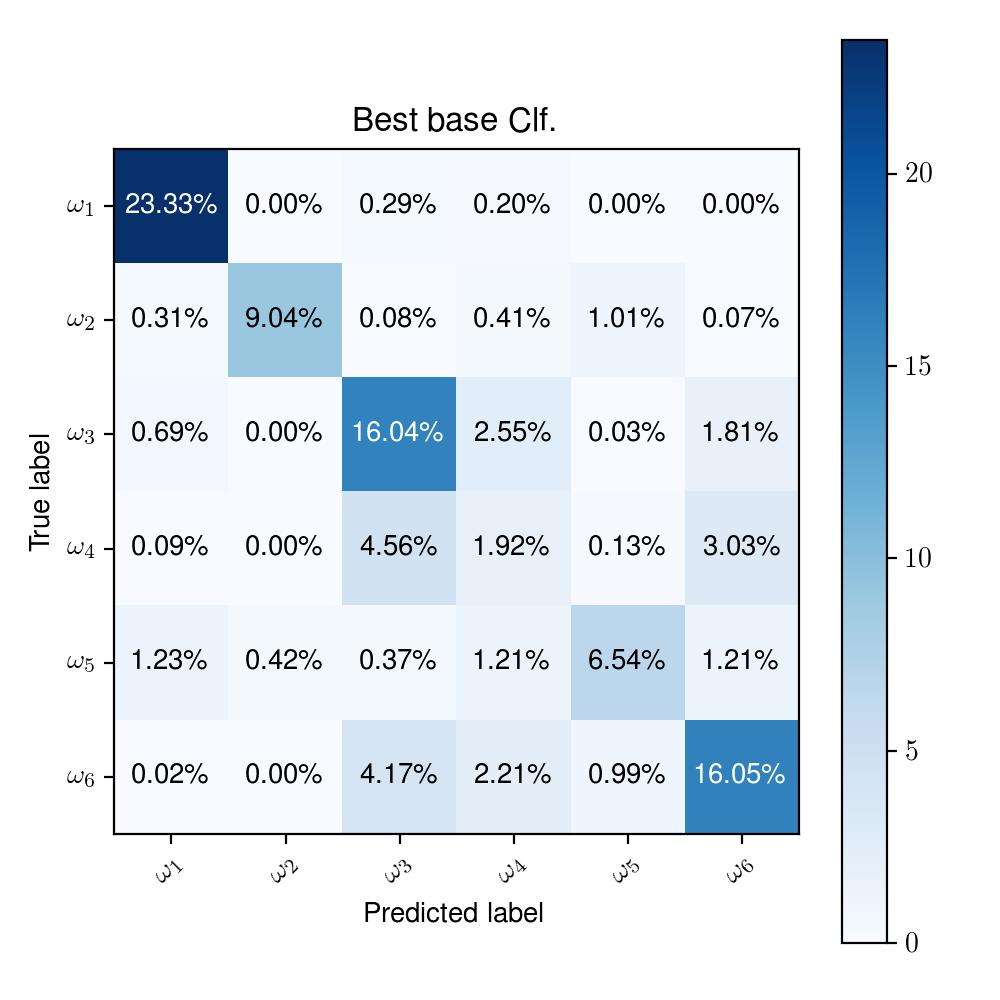} 

  \includegraphics[width=.49\textwidth]{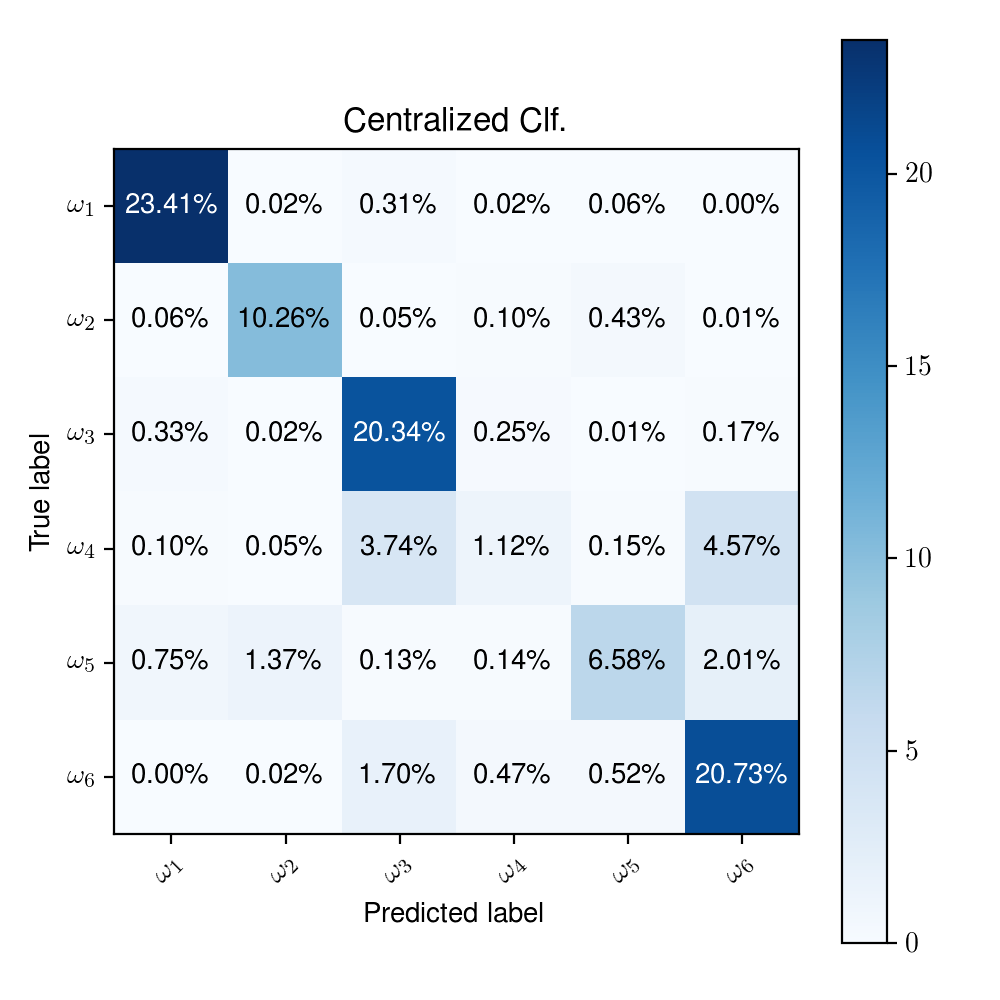} 
\end{center}

\paragraph{Wine}
\begin{center}
  \includegraphics[width=.24\textwidth]{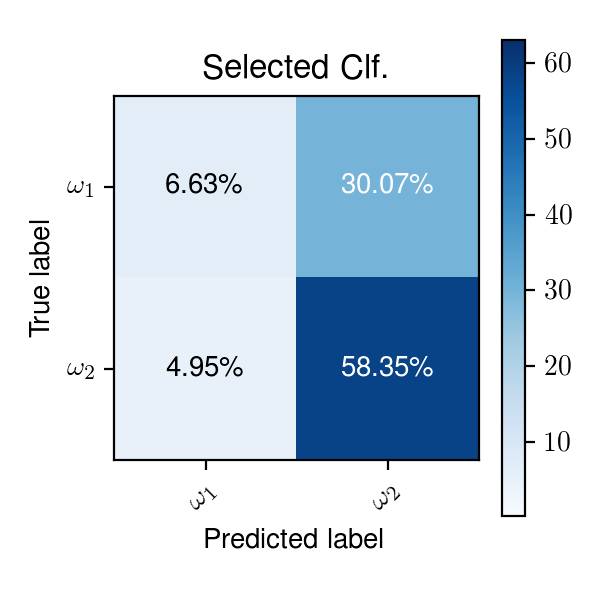} 
  \includegraphics[width=.24\textwidth]{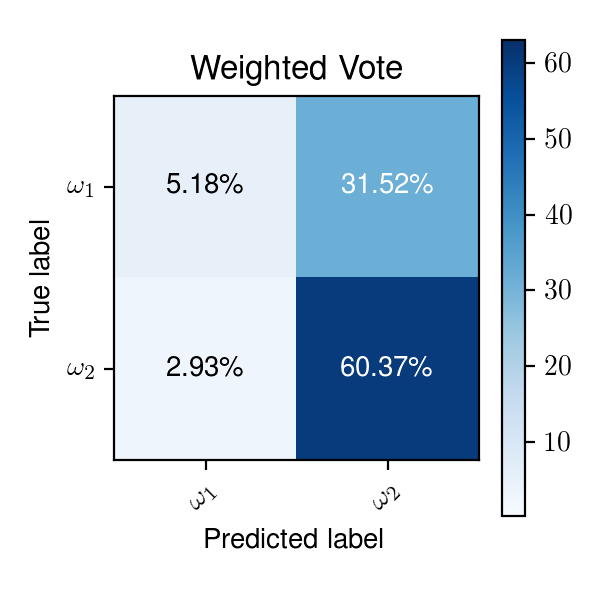}
  \includegraphics[width=.24\textwidth]{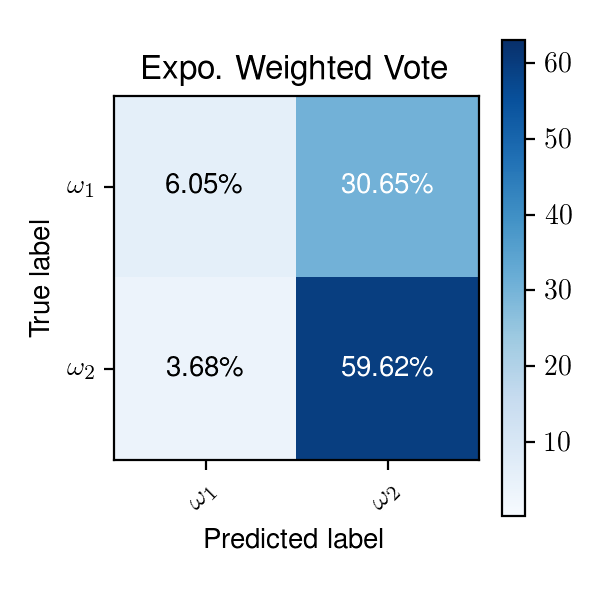} 
  \includegraphics[width=.24\textwidth]{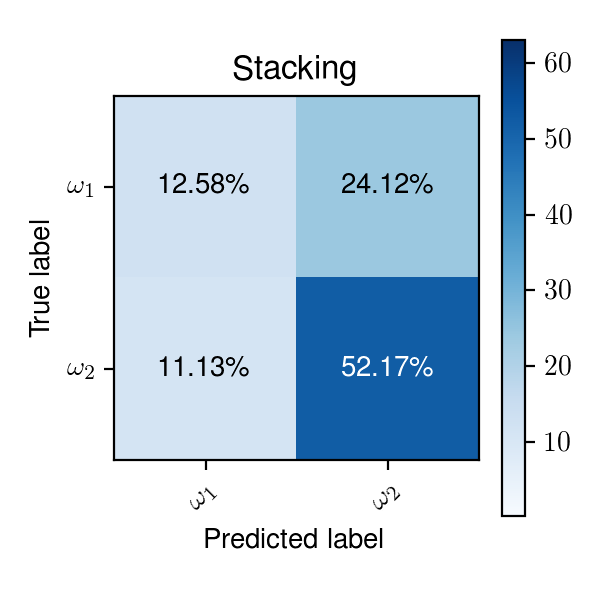} 

  \includegraphics[width=.24\textwidth]{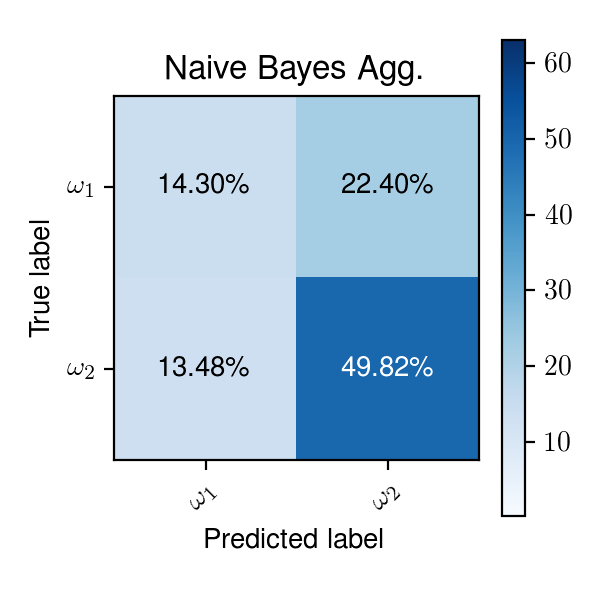} 
  \includegraphics[width=.24\textwidth]{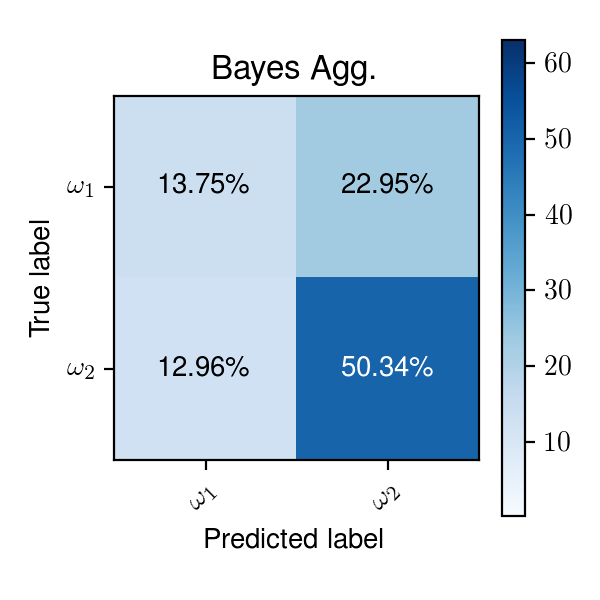} 
  \includegraphics[width=.24\textwidth]{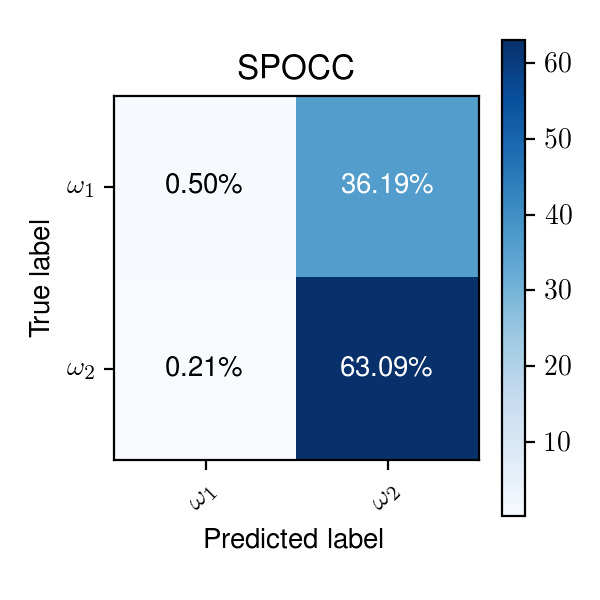}
  \includegraphics[width=.24\textwidth]{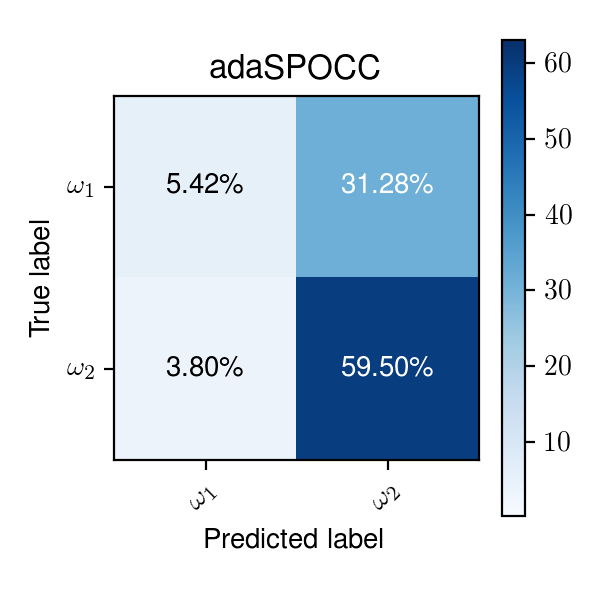} 

  \includegraphics[width=.24\textwidth]{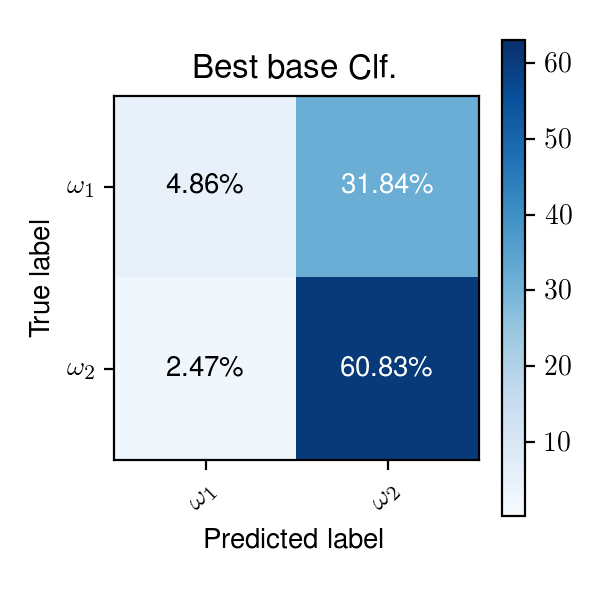} 
  \includegraphics[width=.24\textwidth]{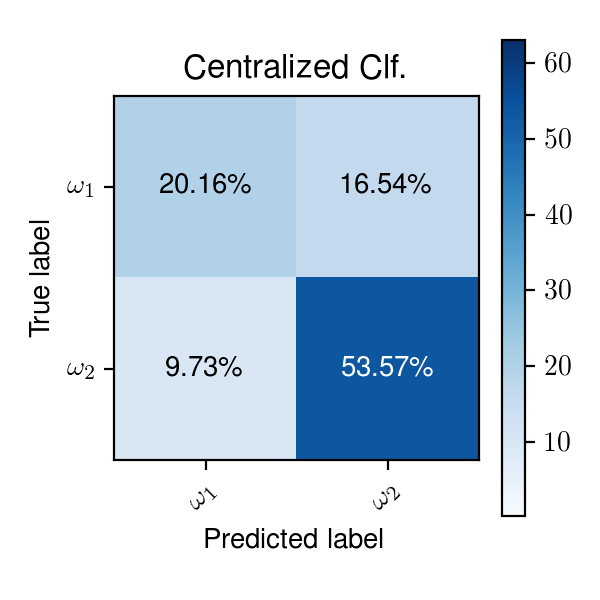} 
\end{center}

\paragraph{Spam}
\begin{center}
  \includegraphics[width=.24\textwidth]{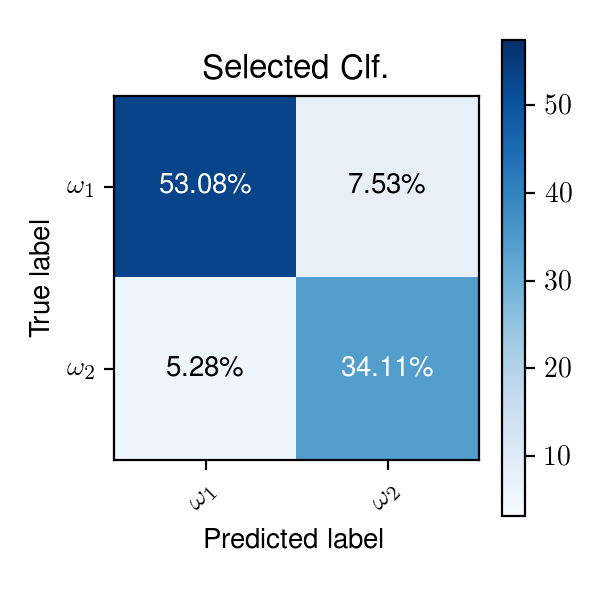} 
  \includegraphics[width=.24\textwidth]{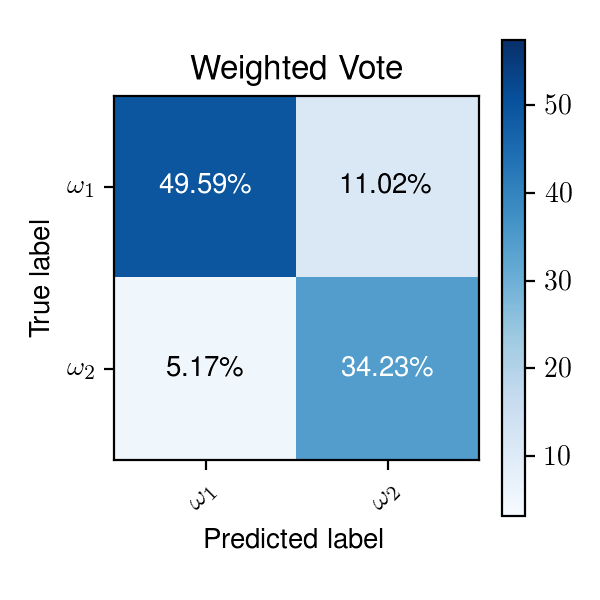}
  \includegraphics[width=.24\textwidth]{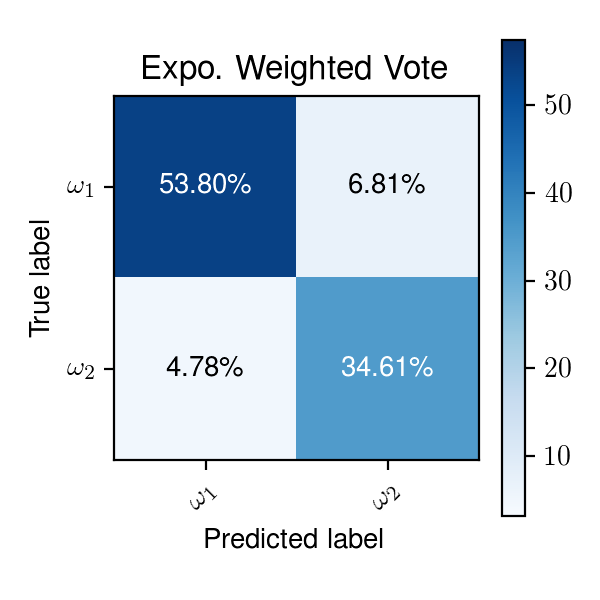} 
  \includegraphics[width=.24\textwidth]{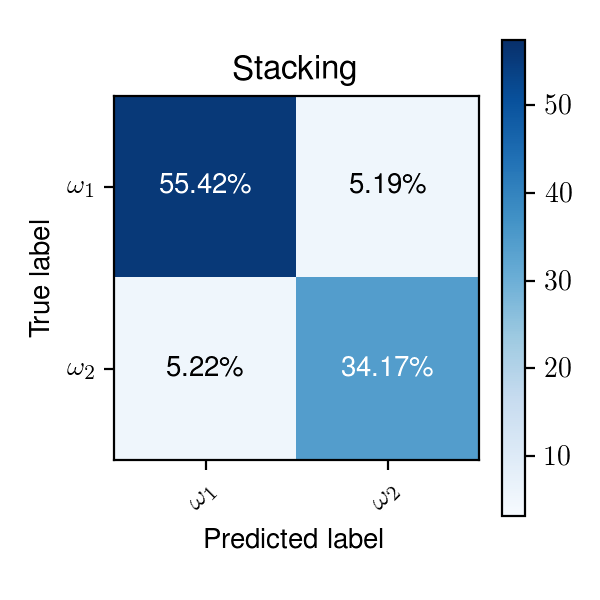} 

  \includegraphics[width=.24\textwidth]{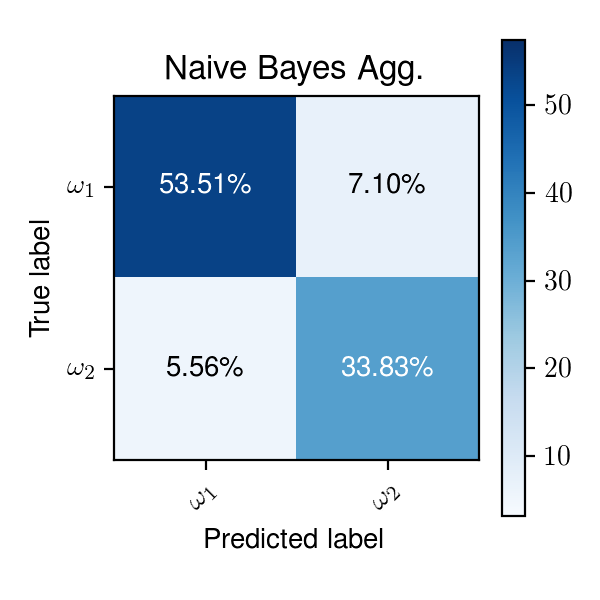} 
  \includegraphics[width=.24\textwidth]{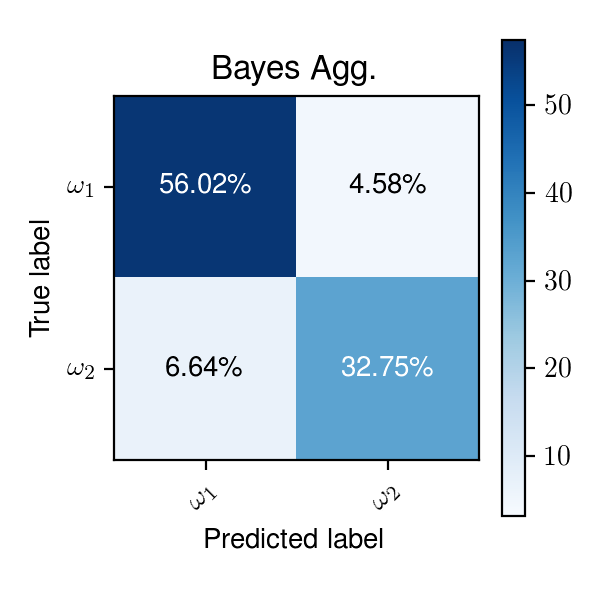} 
  \includegraphics[width=.24\textwidth]{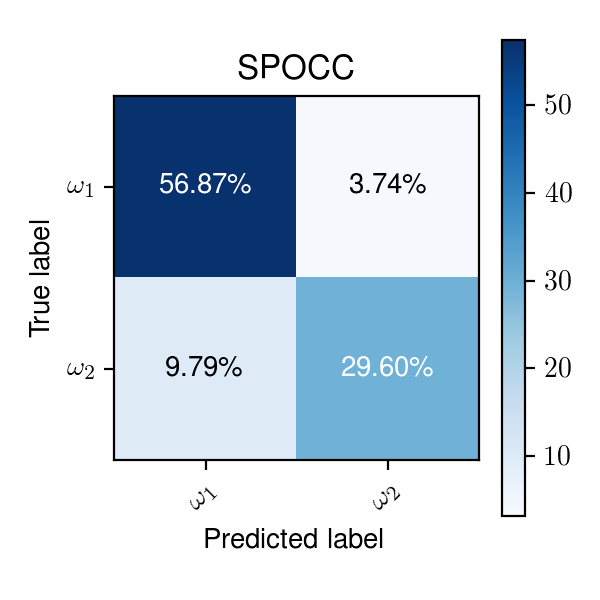}
  \includegraphics[width=.24\textwidth]{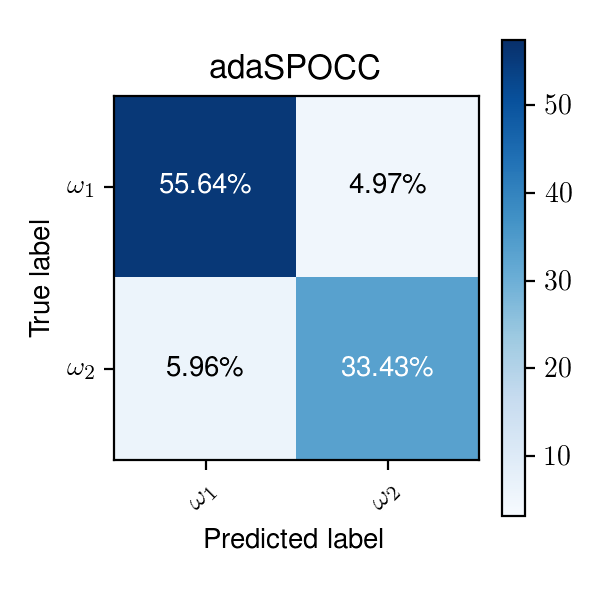} 

  \includegraphics[width=.24\textwidth]{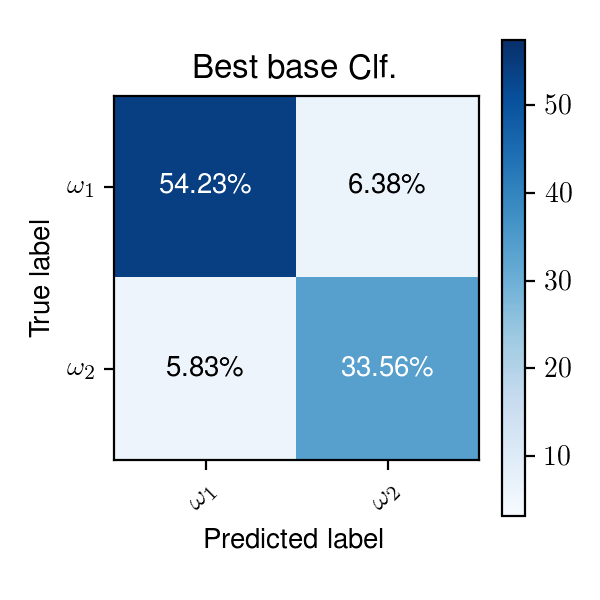} 
  \includegraphics[width=.24\textwidth]{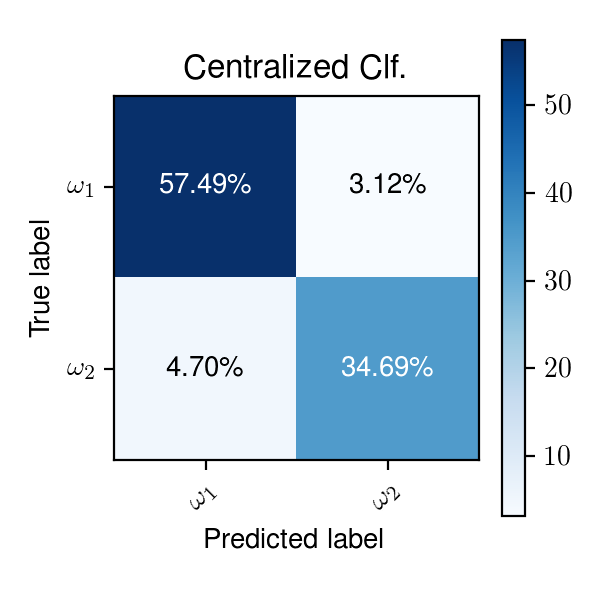} 
\end{center}

\paragraph{Avila}
\begin{center}
  \includegraphics[width=.24\textwidth]{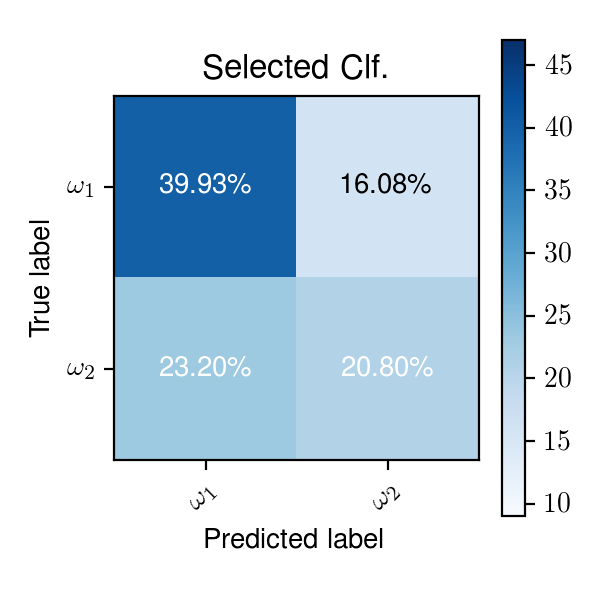} 
  \includegraphics[width=.24\textwidth]{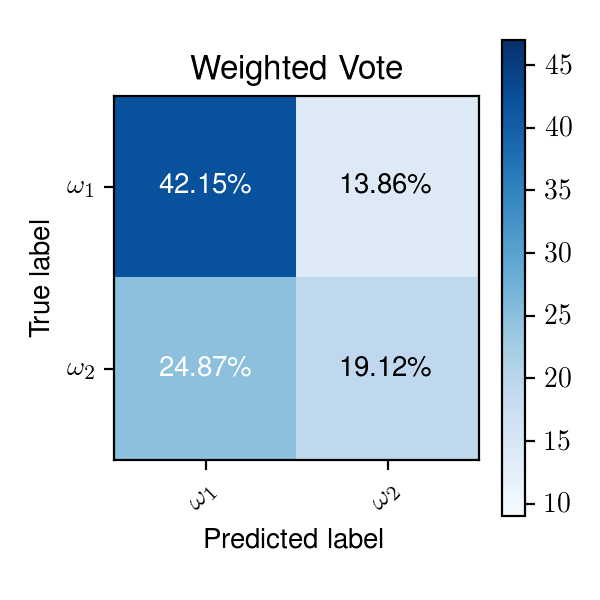}
  \includegraphics[width=.24\textwidth]{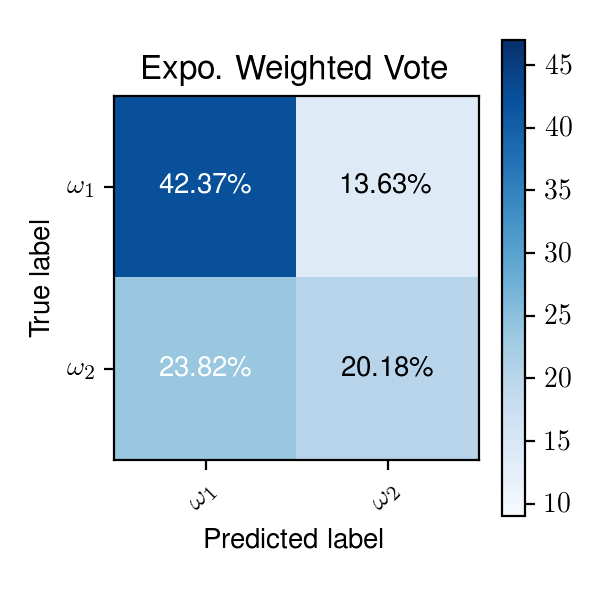} 
  \includegraphics[width=.24\textwidth]{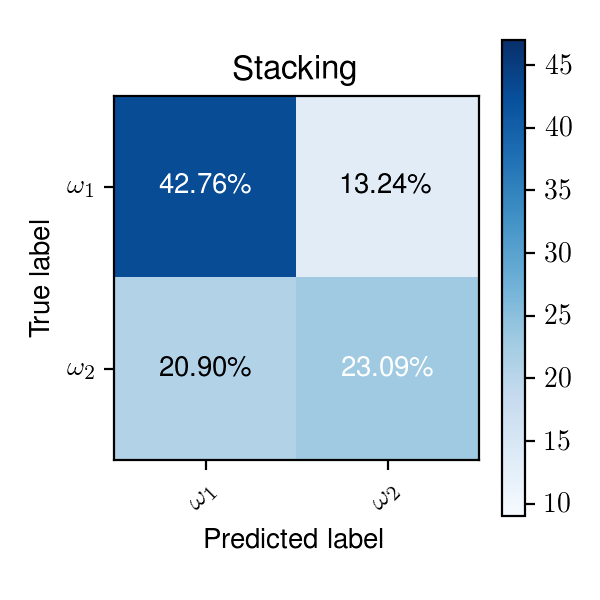} 

  \includegraphics[width=.24\textwidth]{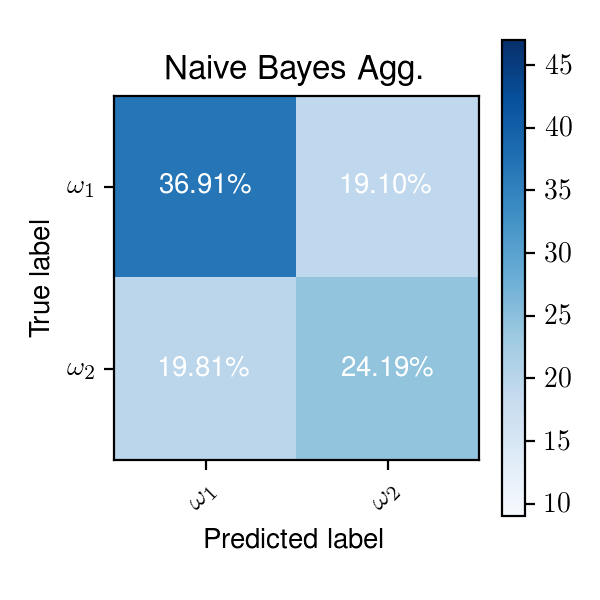} 
  \includegraphics[width=.24\textwidth]{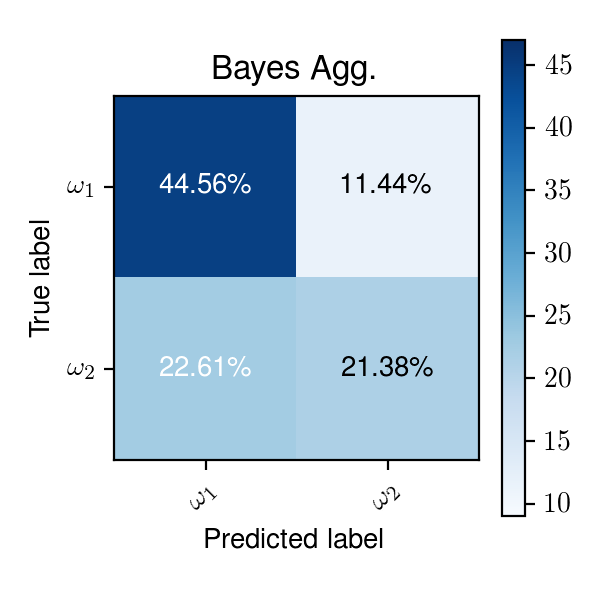} 
  \includegraphics[width=.24\textwidth]{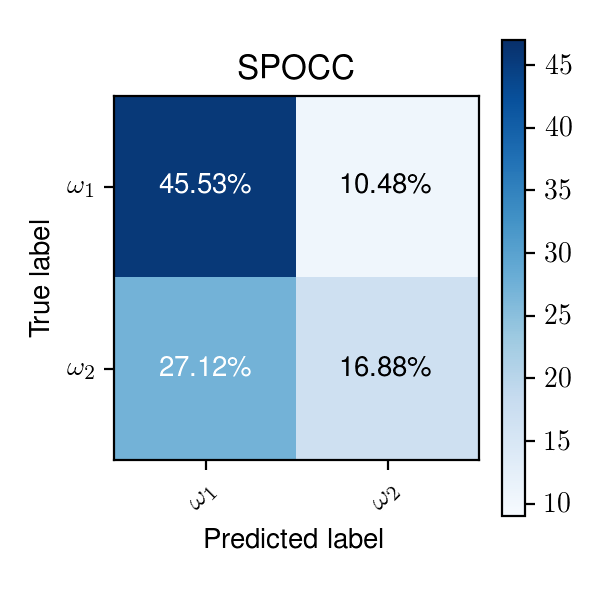}
  \includegraphics[width=.24\textwidth]{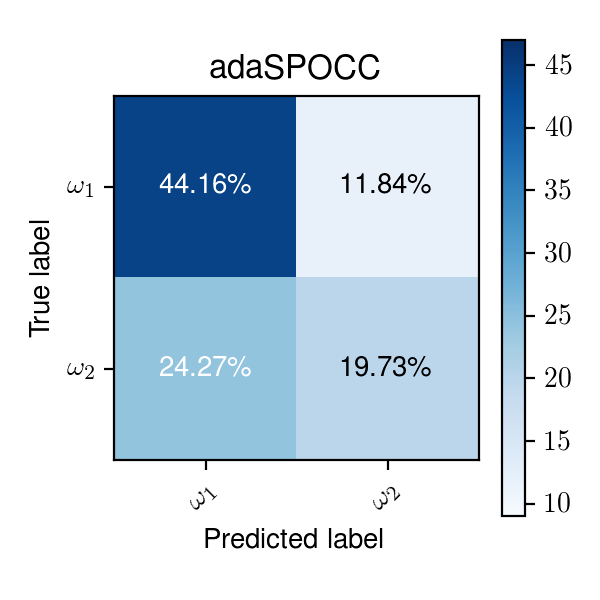} 

  \includegraphics[width=.24\textwidth]{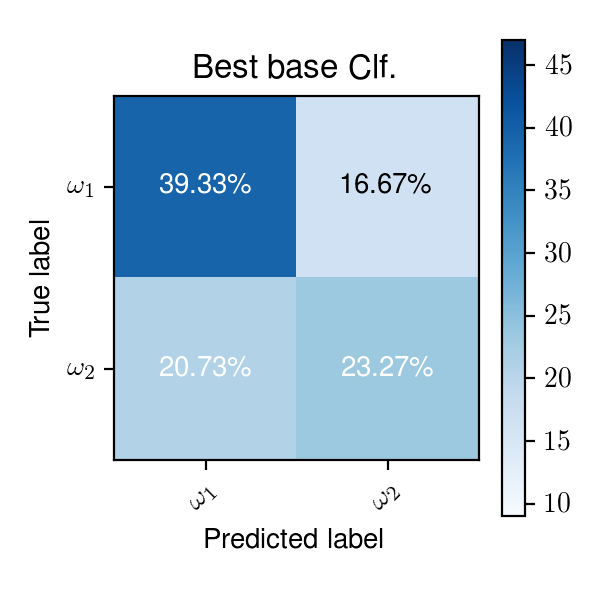} 
  \includegraphics[width=.24\textwidth]{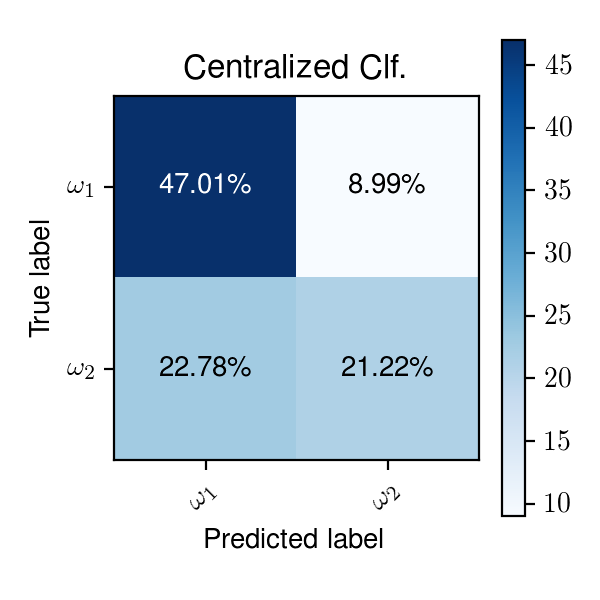} 
\end{center}

\paragraph{Drive}
\begin{center}
  \includegraphics[width=.65\textwidth]{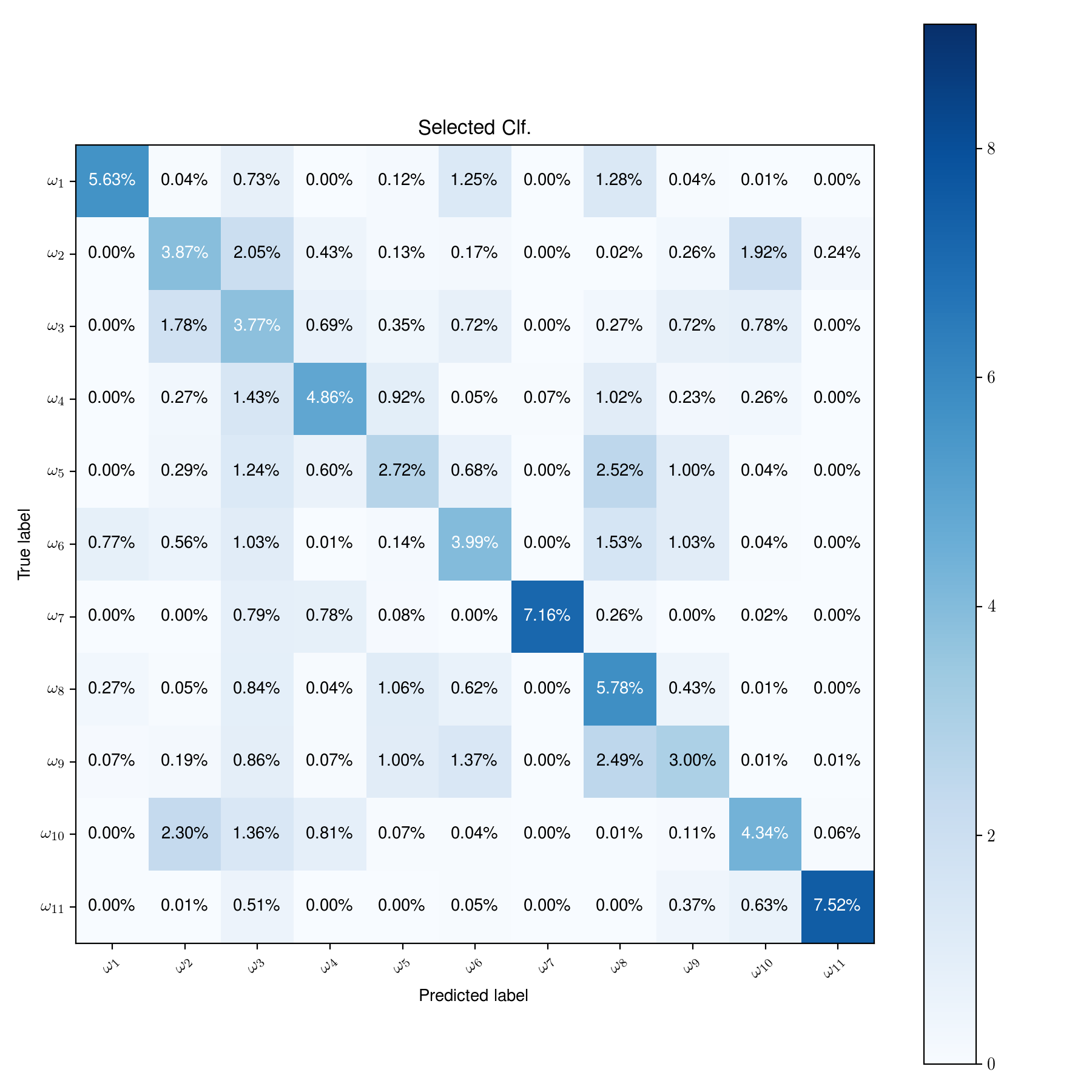} 

  \includegraphics[width=.65\textwidth]{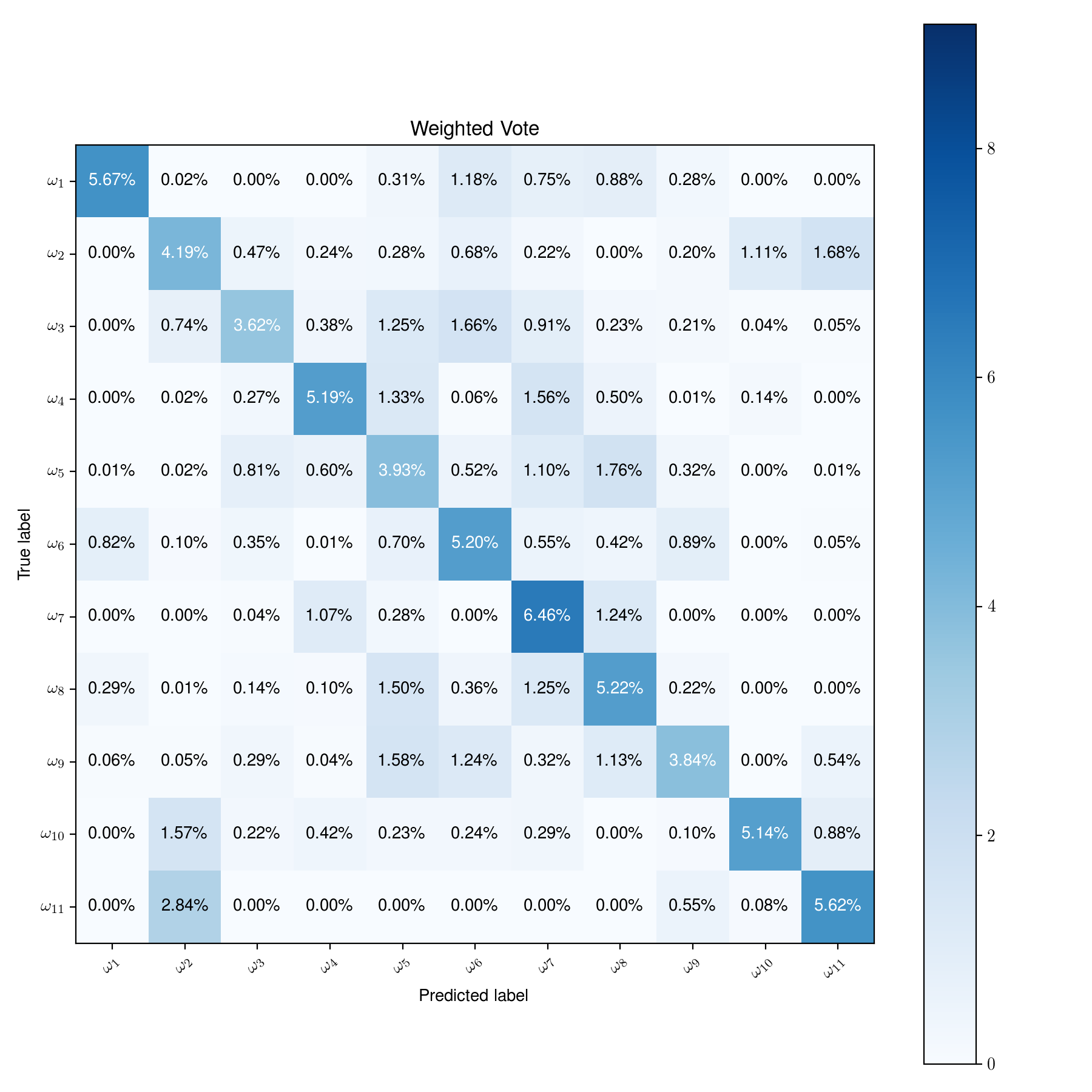}

  \includegraphics[width=.65\textwidth]{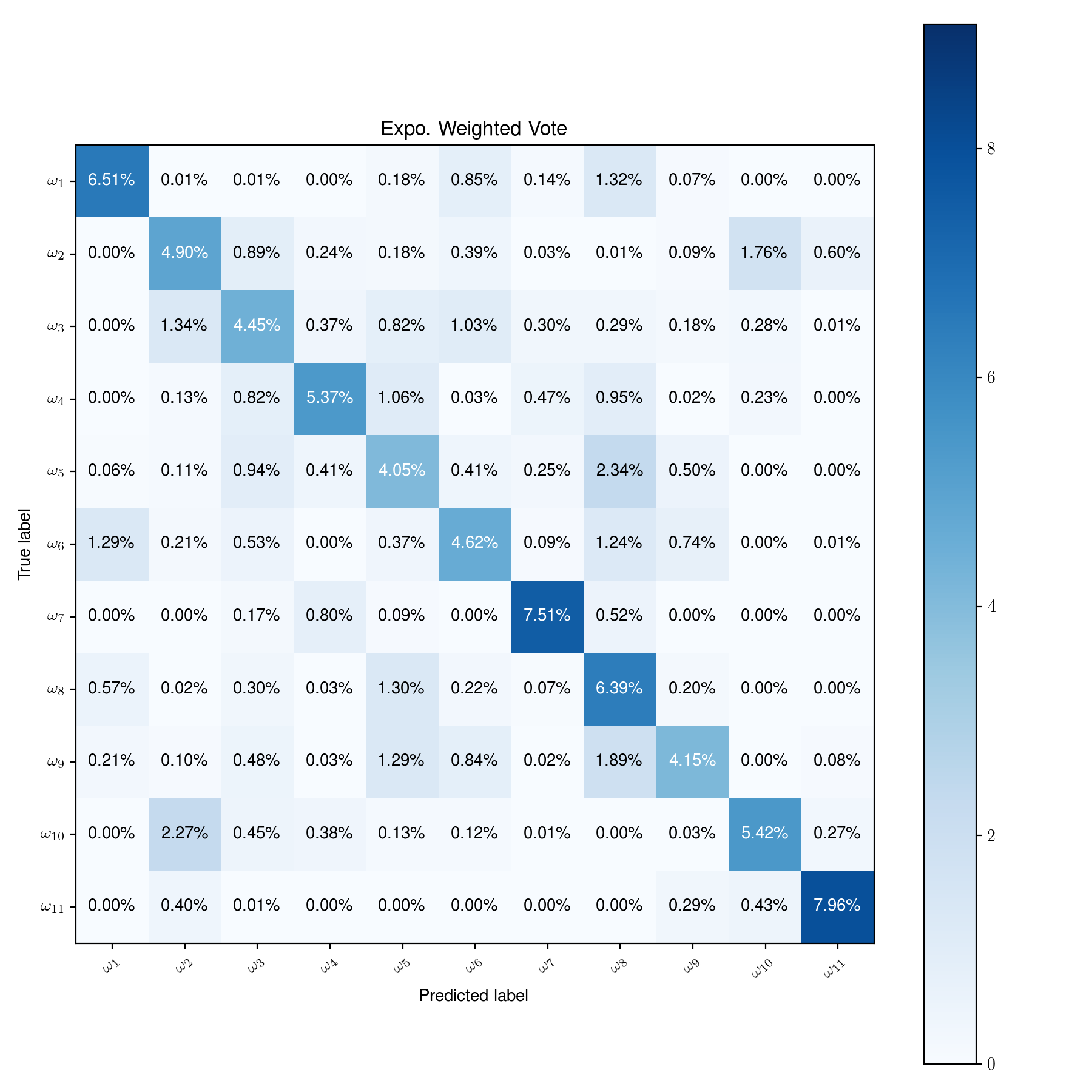} 

  \includegraphics[width=.65\textwidth]{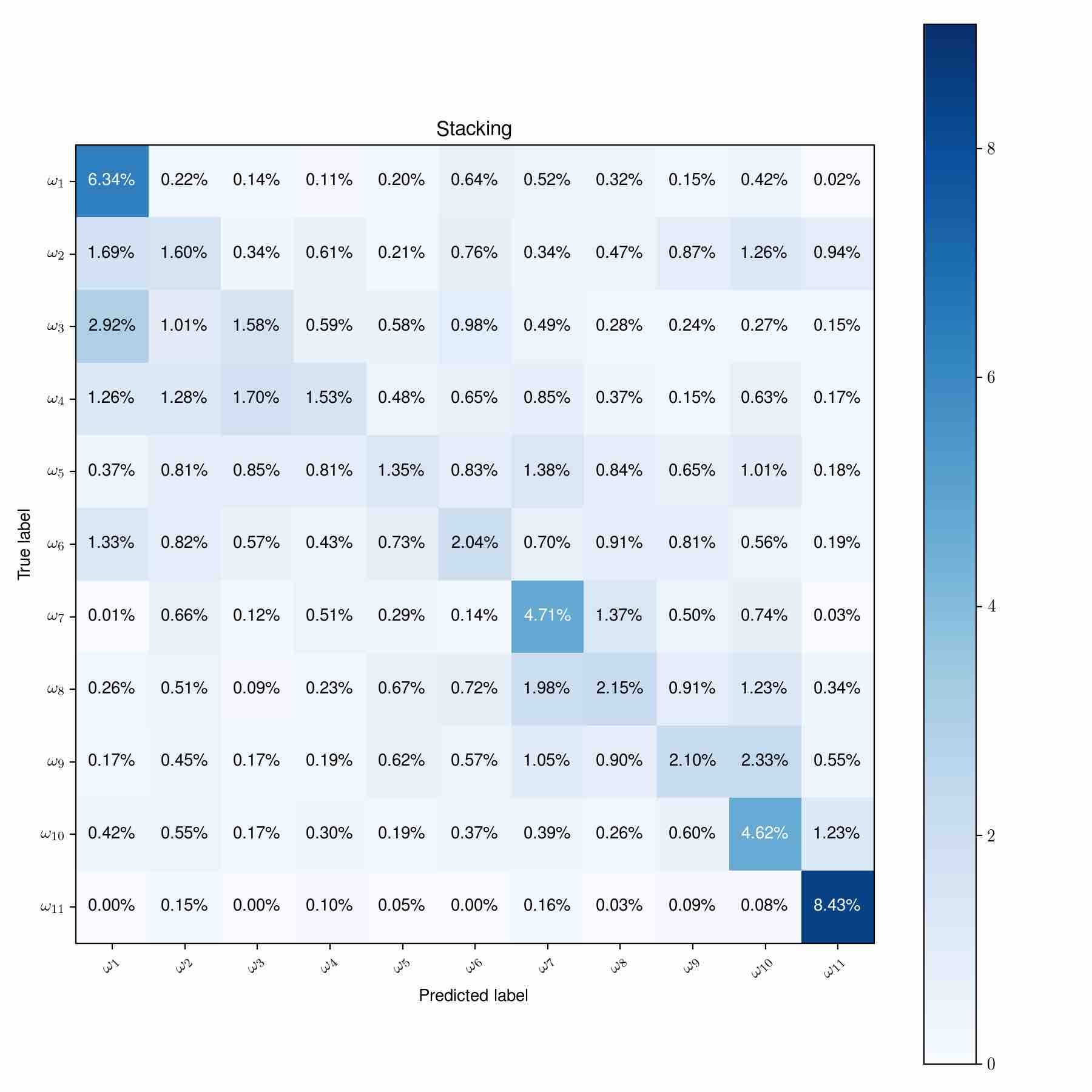} 

  \includegraphics[width=.65\textwidth]{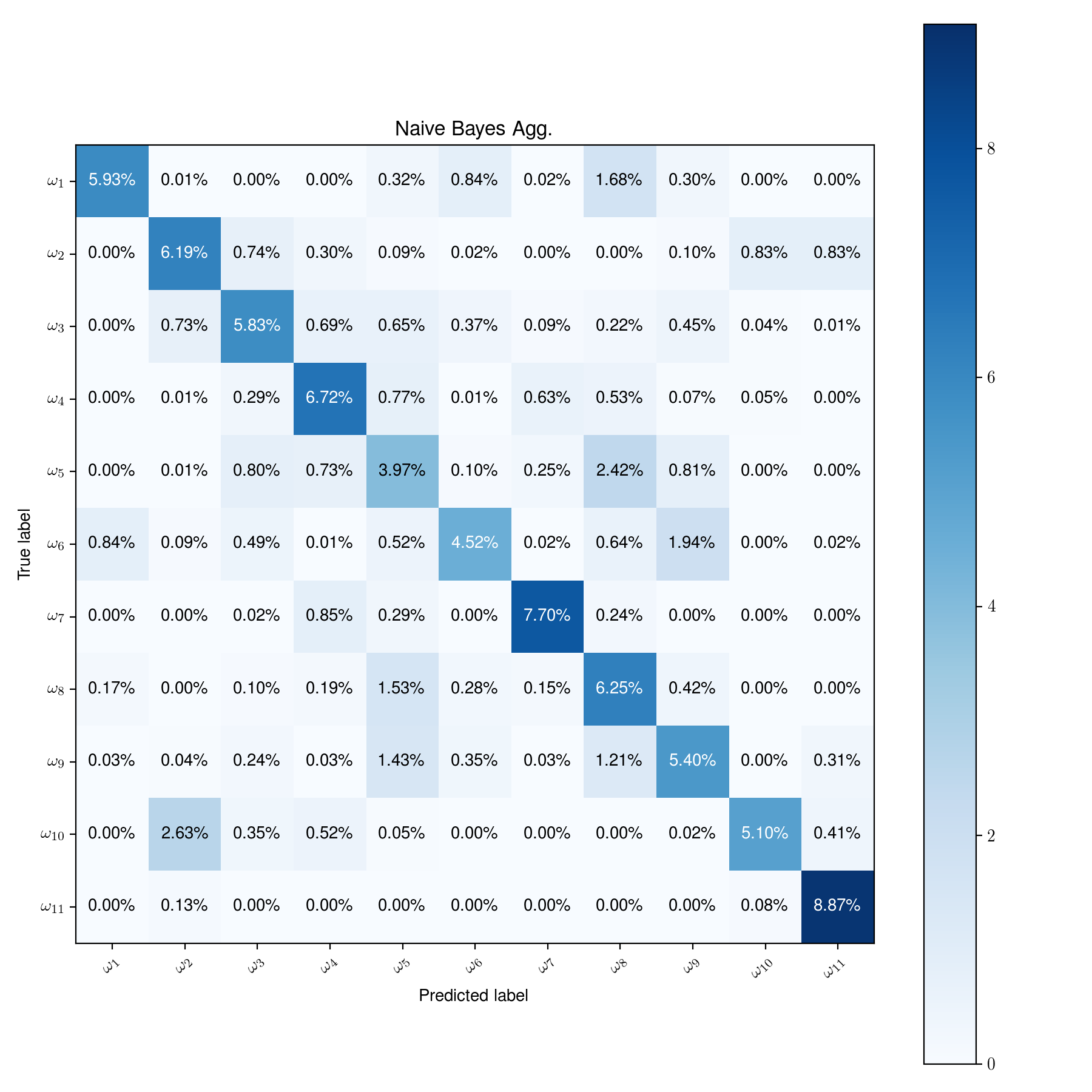} 
  \includegraphics[width=.65\textwidth]{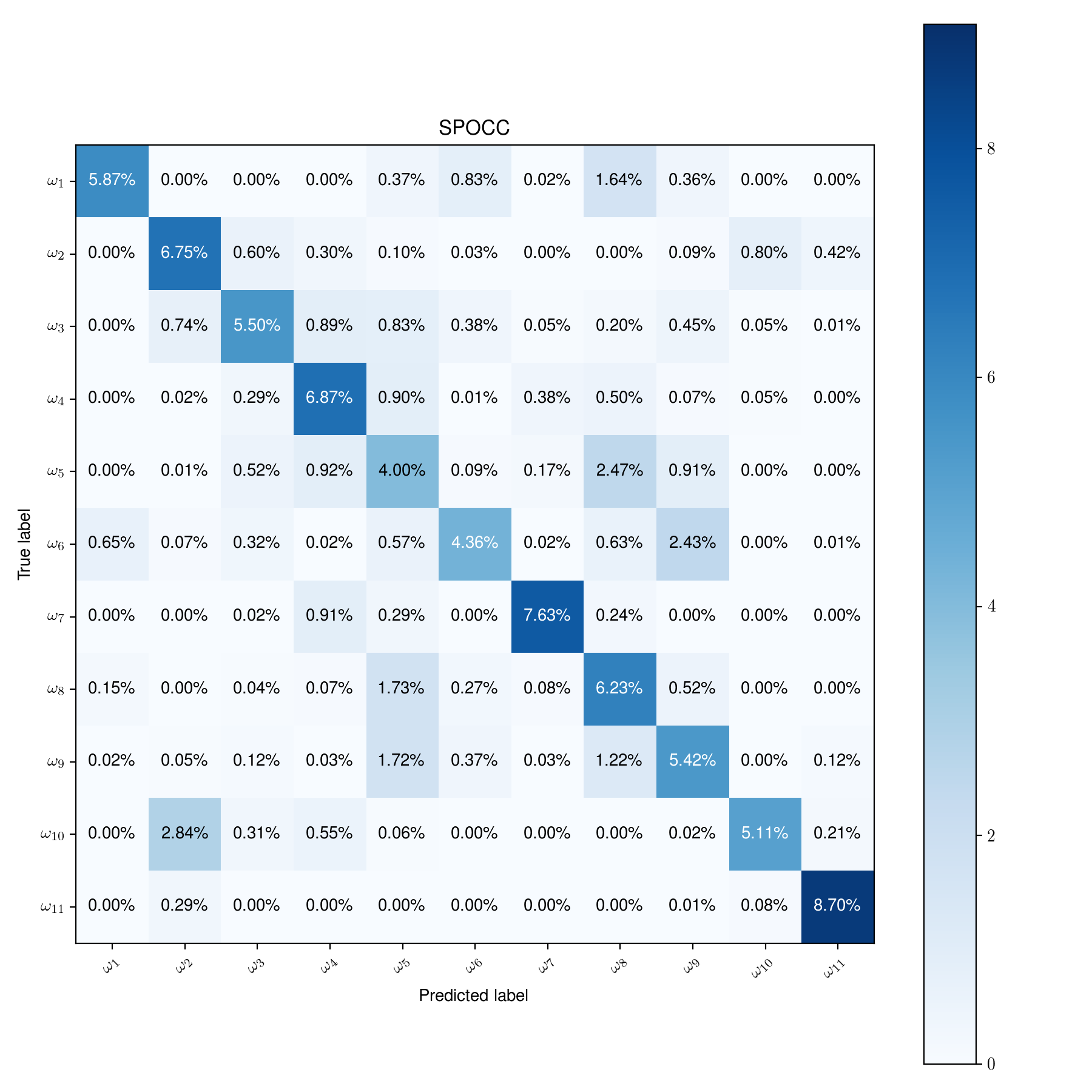}

  \includegraphics[width=.65\textwidth]{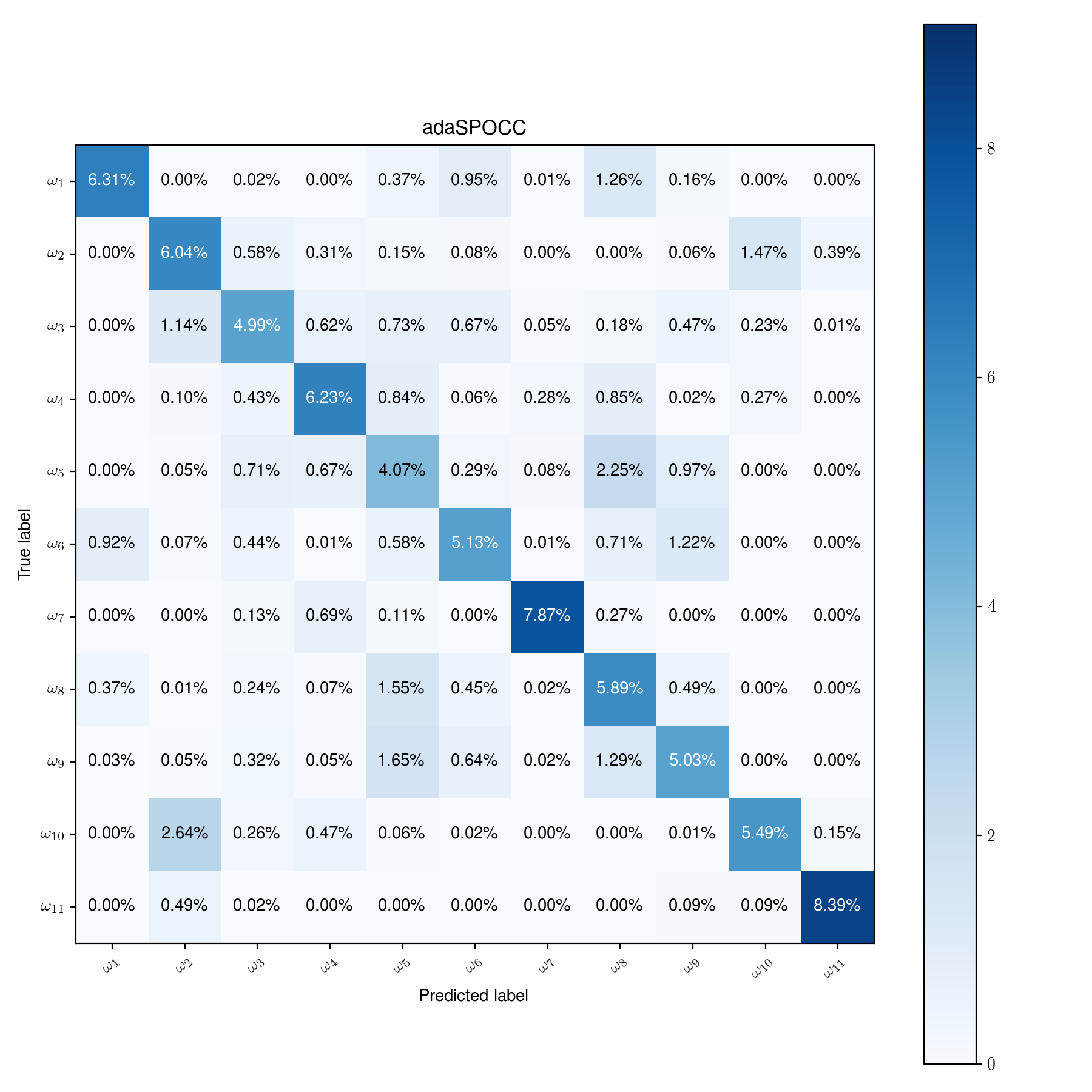} 

  \includegraphics[width=.65\textwidth]{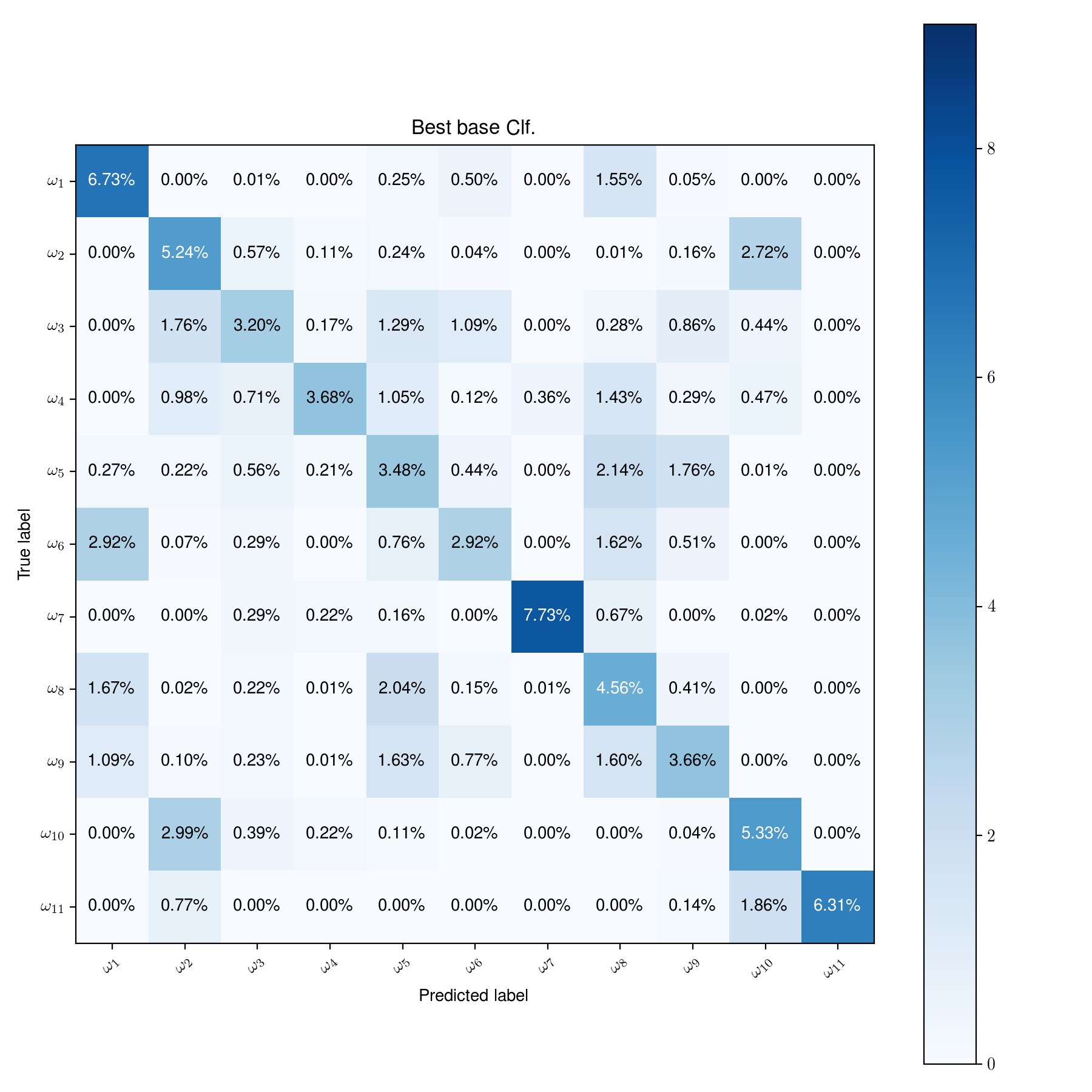} 

  \includegraphics[width=.65\textwidth]{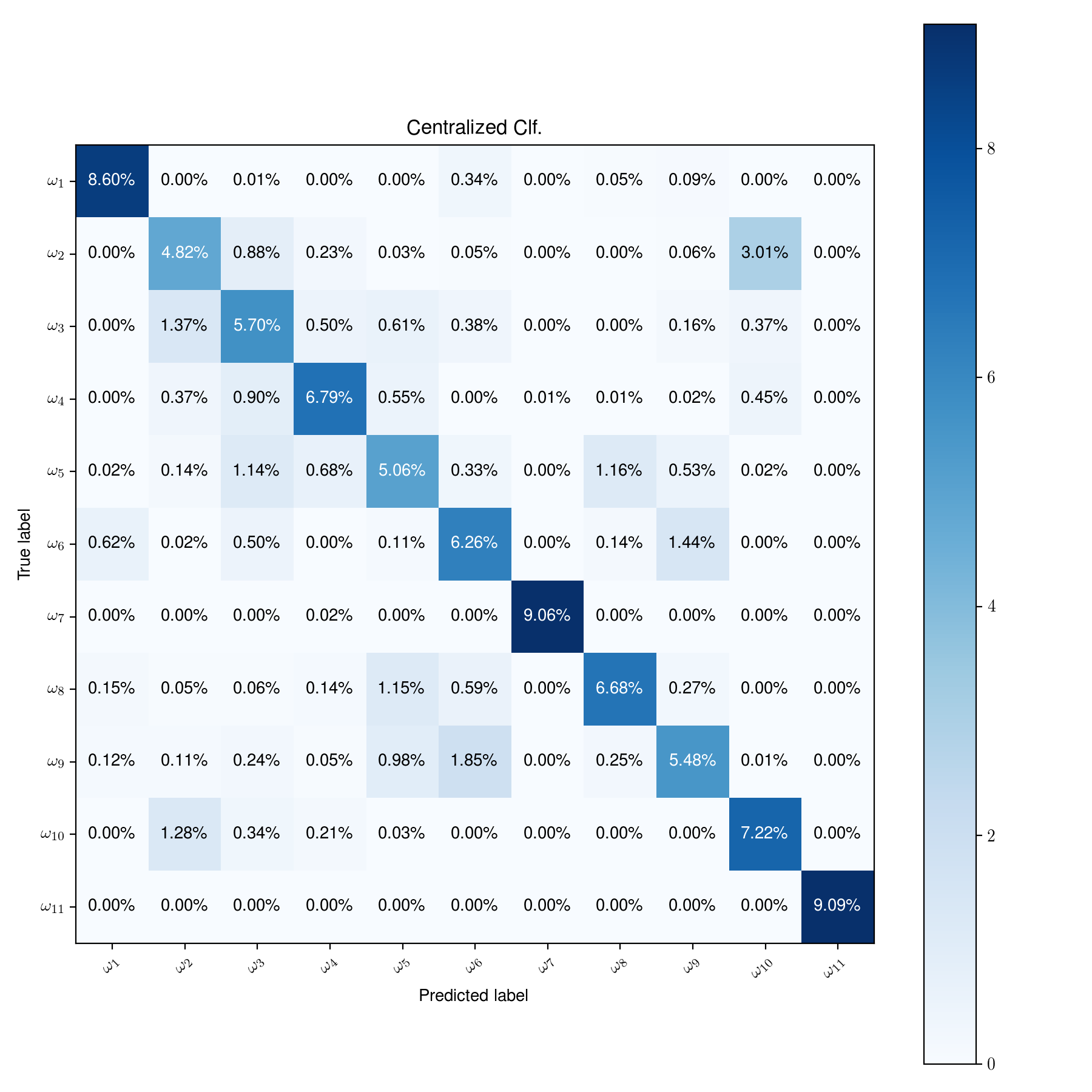} 
\end{center}

\paragraph{Particle}
\begin{center}
  \includegraphics[width=.24\textwidth]{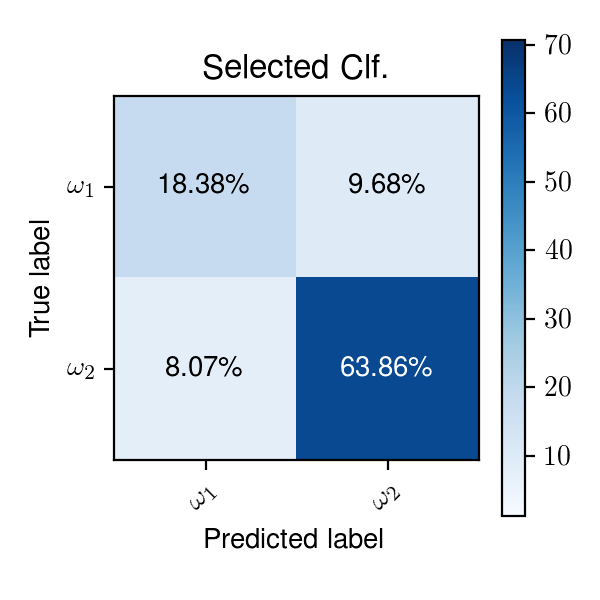} 
  \includegraphics[width=.24\textwidth]{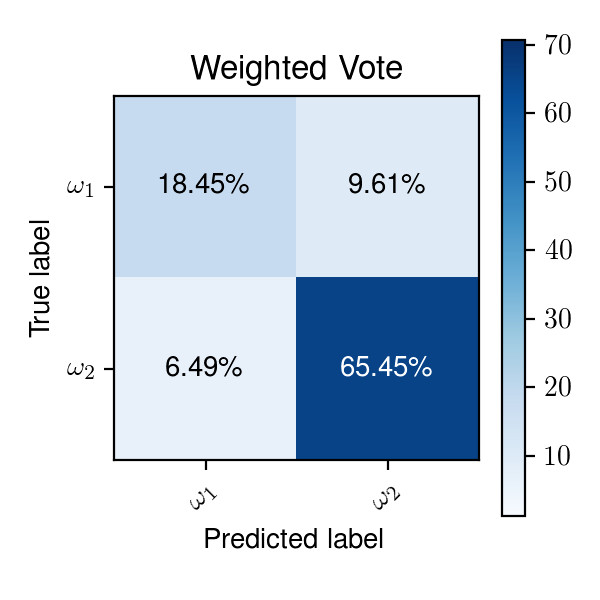}
  \includegraphics[width=.24\textwidth]{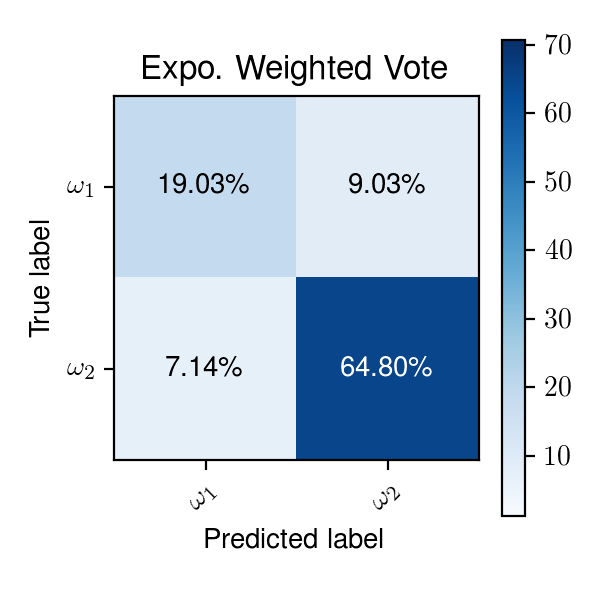} 
  \includegraphics[width=.24\textwidth]{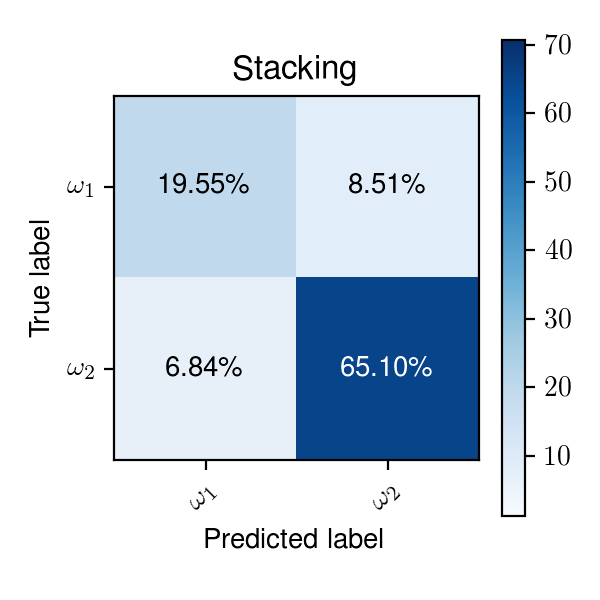} 

  \includegraphics[width=.24\textwidth]{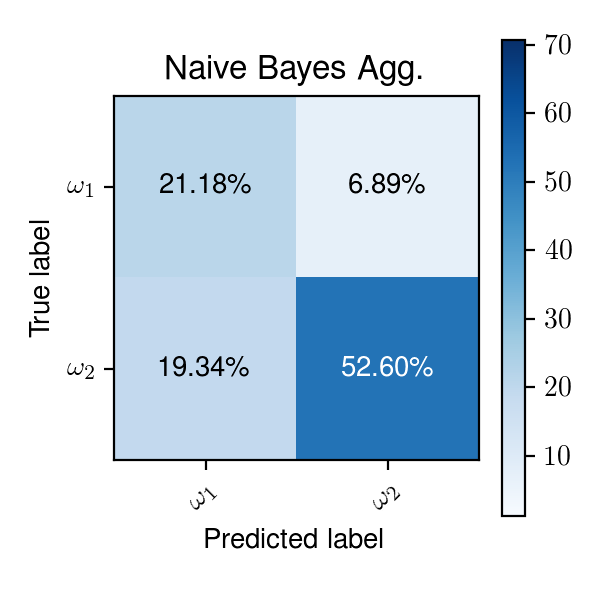} 
  \includegraphics[width=.24\textwidth]{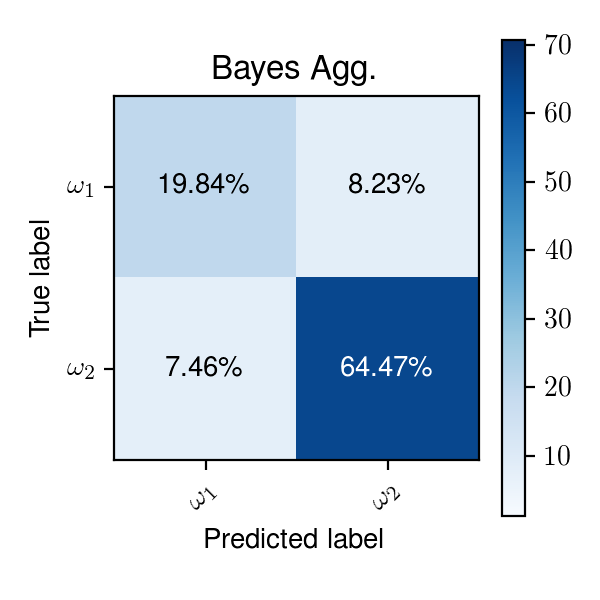} 
  \includegraphics[width=.24\textwidth]{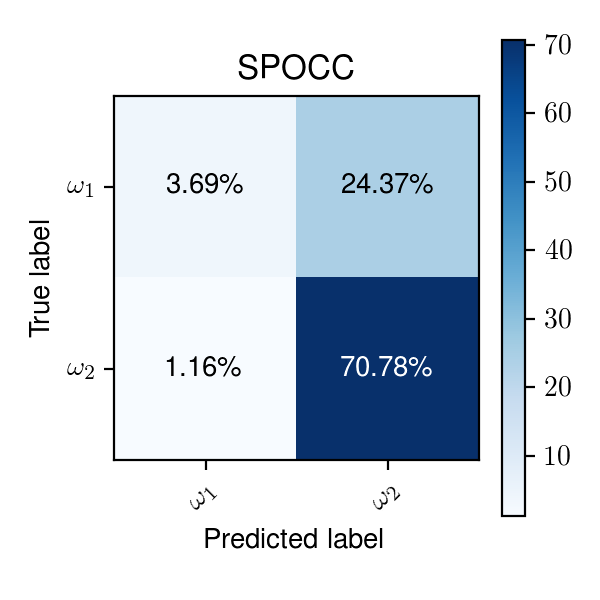}
  \includegraphics[width=.24\textwidth]{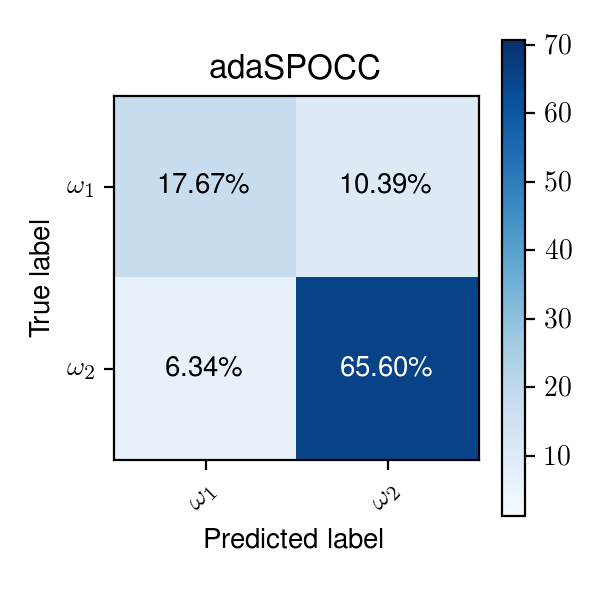}

  \includegraphics[width=.24\textwidth]{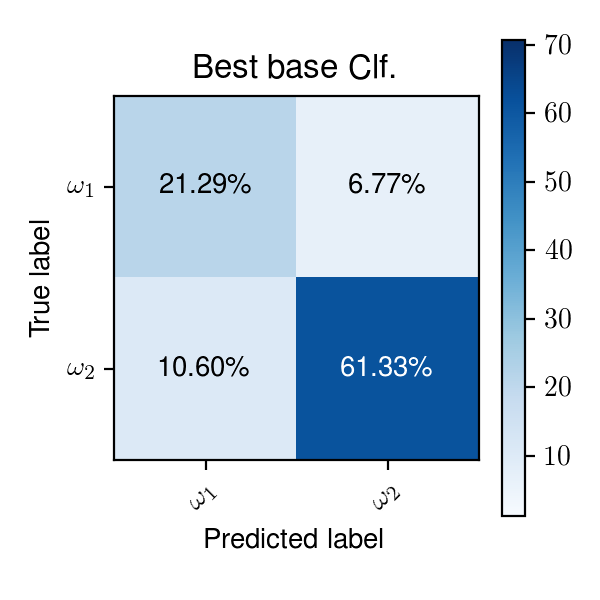} 
  \includegraphics[width=.24\textwidth]{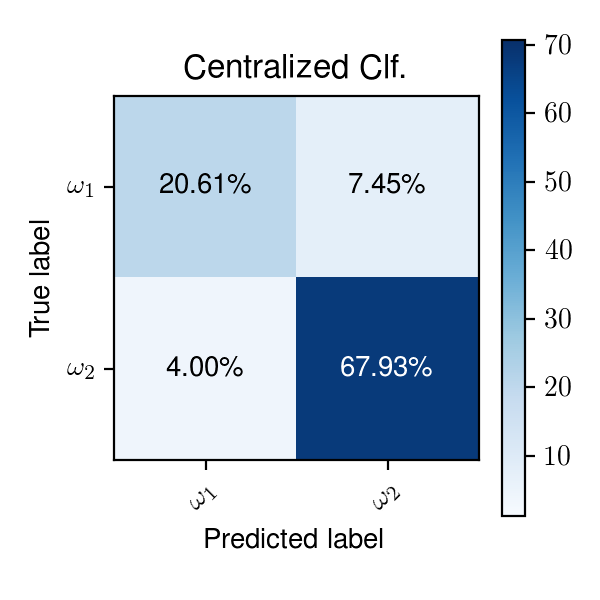} 
\end{center}
 

\bibliographystyle{abbrv}
\bibliography{biblio_JK}

\begin{thebibliography}{10}

\bibitem{alsinet2008logic}
T.~Alsinet, C.~I. Ches{\~n}evar, L.~Godo, and G.~R. Simari.
\newblock A logic programming framework for possibilistic argumentation:
  Formalization and logical properties.
\newblock {\em Fuzzy Sets and Systems}, 159(10):1208--1228, 2008.

\bibitem{amor2018possibilistic}
N.~B. Amor, D.~Dubois, H.~Gouider, and H.~Prade.
\newblock Possibilistic preference networks.
\newblock {\em Information Sciences}, 460:401--415, 2018.

\bibitem{balakrishnan2015simple}
N.~Balakrishnan and M.~Mojirsheibani.
\newblock A simple method for combining estimates to improve the overall error
  rates in classification.
\newblock {\em Computational Statistics}, 30(4):1033--1049, 2015.

\bibitem{bella2013effect}
A.~Bella, C.~Ferri, J.~Hern{\'a}ndez-Orallo, and M.~J. Ram{\'\i}rez-Quintana.
\newblock On the effect of calibration in classifier combination.
\newblock {\em Applied intelligence}, 38(4):566--585, 2013.

\bibitem{benferhat1999possibilistic}
S.~Benferhat, D.~Dubois, L.~Garcia, and H.~Prade.
\newblock Possibilistic logic bases and possibilistic graphs.
\newblock In {\em Proceedings of the Fifteenth conference on Uncertainty in
  artificial intelligence}, pages 57--64. Morgan Kaufmann Publishers Inc.,
  1999.

\bibitem{biau2016cobra}
G.~Biau, A.~Fischer, B.~Guedj, and J.~D. Malley.
\newblock Cobra: A combined regression strategy.
\newblock {\em Journal of Multivariate Analysis}, 146:18--28, 2016.

\bibitem{breiman1996bagging}
L.~Breiman.
\newblock Bagging predictors.
\newblock {\em Machine learning}, 24(2):123--140, 1996.

\bibitem{brown2005diversity}
G.~Brown, J.~Wyatt, R.~Harris, and X.~Yao.
\newblock Diversity creation methods: a survey and categorisation.
\newblock {\em Information Fusion}, 6(1):5--20, 2005.

\bibitem{de1999supremum}
G.~De~Cooman and D.~Aeyels.
\newblock Supremum preserving upper probabilities.
\newblock {\em Information Sciences}, 118(1-4):173--212, 1999.

\bibitem{deerwester1990indexing}
S.~Deerwester, S.~T. Dumais, G.~W. Furnas, T.~K. Landauer, and R.~Harshman.
\newblock Indexing by latent semantic analysis.
\newblock {\em Journal of the American society for information science},
  41(6):391--407, 1990.

\bibitem{destercke2008possibilistic}
S.~Destercke, D.~Dubois, and E.~Chojnacki.
\newblock Possibilistic information fusion using maximal coherent subsets.
\newblock {\em IEEE Transactions on Fuzzy Systems}, 17(1):79--92, 2008.

\bibitem{dubois2004probaposs}
D.~Dubois, L.~Foulloy, G.~Mauris, and H.~Prade.
\newblock Probability-possibility transformations, triangular fuzzy sets, and
  probabilistic inequalities.
\newblock {\em Reliable computing}, 10(4):273--297, 2004.

\bibitem{dubois1994automated}
D.~Dubois, J.~Lang, and H.~Prade.
\newblock Automated reasoning using possibilistic logic: Semantics, belief
  revision, and variable certainty weights.
\newblock {\em IEEE transactions on knowledge and data engineering},
  6(1):64--71, 1994.

\bibitem{Dub82}
D.~Dubois and H.~Prade.
\newblock On several representations of an uncertain body of evidence.
\newblock {\em Fuzzy Information and Decision Processes}, pages 161--181, 1982.

\bibitem{dubois1988possibility}
D.~Dubois and H.~Prade.
\newblock {\em Possibility theory: An approach to the computerized processing
  of information}.
\newblock Plenum Press, New York, 1988.

\bibitem{dubois1992upper}
D.~Dubois and H.~Prade.
\newblock When upper probabilities are possibility measures.
\newblock {\em Fuzzy sets and systems}, 49(1):65--74, 1992.

\bibitem{dubois2015possibility}
D.~Dubois and H.~Prade.
\newblock Possibility theory and its applications: Where do we stand?
\newblock In {\em Springer Handbook of Computational Intelligence}, pages
  31--60. Springer, 2015.

\bibitem{dubois2017generalized}
D.~Dubois, H.~Prade, and S.~Schockaert.
\newblock Generalized possibilistic logic: foundations and applications to
  qualitative reasoning about uncertainty.
\newblock {\em Artificial Intelligence}, 252:139--174, 2017.

\bibitem{GILBOA198765}
I.~Gilboa.
\newblock Expected utility with purely subjective non-additive probabilities.
\newblock {\em Journal of Mathematical Economics}, 16(1):65 -- 88, 1987.

\bibitem{goodman1982fuzzy}
I.~R. Goodman.
\newblock Fuzzy sets as equivalence classes of random sets.
\newblock In R.~Yager, editor, {\em Fuzzy Sets and Possibility Theory}, pages
  pp. 327--343. Pergamon Press, Oxford, 1982.

\bibitem{guo2019ifusion}
K.~Guo, T.~Xu, X.~Kui, R.~Zhang, and T.~Chi.
\newblock ifusion: Towards efficient intelligence fusion for deep learning from
  real-time and heterogeneous data.
\newblock {\em Information Fusion}, 51:215--223, 2019.

\bibitem{hoang2019collective}
M.~Hoang, N.~Hoang, B.~K.~H. Low, and C.~Kingsford.
\newblock Collective model fusion for multiple black-box experts.
\newblock In {\em International Conference on Machine Learning}, pages
  2742--2750, 2019.

\bibitem{huang1995method}
Y.~S. Huang and C.~Y. Suen.
\newblock A method of combining multiple experts for the recognition of
  unconstrained handwritten numerals.
\newblock {\em IEEE transactions on pattern analysis and machine intelligence},
  17(1):90--94, 1995.

\bibitem{hullermeier2002possibilistic}
E.~H{\"u}llermeier.
\newblock Possibilistic induction in decision-tree learning.
\newblock In {\em European Conference on Machine Learning}, pages 173--184.
  Springer, 2002.

\bibitem{ijcai2019-363}
J.~Ji, X.~Chen, Q.~Wang, L.~Yu, and P.~Li.
\newblock Learning to learn gradient aggregation by gradient descent.
\newblock In {\em Proceedings of the Twenty-Eighth International Joint
  Conference on Artificial Intelligence, {IJCAI-19}}, pages 2614--2620.
  International Joint Conferences on Artificial Intelligence Organization, 7
  2019.

\bibitem{Jou12}
A.-L. Jousselme and P.~Maupin.
\newblock Distances in evidence theory: Comprehensive survey and
  generalizations.
\newblock {\em International Journal of Approximate Reasoning}, 53(2):118 --
  145, 2012.
\newblock Theory of Belief Functions (BELIEF 2010).

\bibitem{de1982interpretation}
J.~Kamp{\'e}~de F{\'e}riet.
\newblock Interpretation of membership functions of fuzzy sets in terms of
  plausibility and belief.
\newblock In E.~S. Madan M.~Gupta, editor, {\em Fuzzy Information and Decision
  Processes}, pages 93--98. North-Holland, Amsterdam, 1982.

\bibitem{kim2012bayesian}
H.-C. Kim and Z.~Ghahramani.
\newblock Bayesian classifier combination.
\newblock In {\em Artificial Intelligence and Statistics}, pages 619--627,
  2012.

\bibitem{NIPS2019_8728}
T.~Kim and J.~Ghosh.
\newblock On single source robustness in deep fusion models.
\newblock In H.~Wallach, H.~Larochelle, A.~Beygelzimer, F.~d'~Alch\'{e}-Buc,
  E.~Fox, and R.~Garnett, editors, {\em Advances in Neural Information
  Processing Systems 32}, pages 4815--4826. Curran Associates, Inc., 2019.

\bibitem{lacoste2014agnostic}
A.~Lacoste, M.~Marchand, F.~Laviolette, and H.~Larochelle.
\newblock Agnostic bayesian learning of ensembles.
\newblock In {\em International Conference on Machine Learning}, pages
  611--619, 2014.

\bibitem{li2019rsa}
L.~Li, W.~Xu, T.~Chen, G.~B. Giannakis, and Q.~Ling.
\newblock Rsa: Byzantine-robust stochastic aggregation methods for distributed
  learning from heterogeneous datasets.
\newblock In {\em Proceedings of the AAAI Conference on Artificial
  Intelligence}, volume~33, pages 1544--1551, 2019.

\bibitem{li2019exploiting}
Y.~Li, B.~Rubinstein, and T.~Cohn.
\newblock Exploiting worker correlation for label aggregation in crowdsourcing.
\newblock In {\em International Conference on Machine Learning}, pages
  3886--3895, 2019.

\bibitem{liu2019advancing}
H.~Liu and L.~Zhang.
\newblock Advancing ensemble learning performance through data transformation
  and classifiers fusion in granular computing context.
\newblock {\em Expert Systems with Applications}, 131:20--29, 2019.

\bibitem{loustau2008aggregation}
S.~Loustau.
\newblock Aggregation of svm classifiers using sobolev spaces.
\newblock {\em Journal of Machine Learning Research}, 9(Jul):1559--1582, 2008.

\bibitem{ma2019secure}
X.~Ma, C.~Ji, X.~Zhang, J.~Wang, J.~Li, and K.-C. Li.
\newblock Secure multiparty learning from aggregation of locally trained
  models.
\newblock In {\em International Conference on Machine Learning for Cyber
  Security}, pages 173--182. Springer, 2019.

\bibitem{madry2018}
A.~Madry, A.~Makelov, L.~Schmidt, D.~Tsipras, and A.~Vladu.
\newblock Towards deep learning models resistant to adversarial attacks.
\newblock In {\em International Conference on Learning Representations}, 2018.

\bibitem{menon2019online}
A.~K. Menon, A.~Rajagopalan, B.~Sumengen, G.~Citovsky, Q.~Cao, and S.~Kumar.
\newblock Online hierarchical clustering approximations.
\newblock {\em arXiv preprint arXiv:1909.09667}, 2019.

\bibitem{pei1982treating}
W.~Pei-Zhuang and E.~Sanchez.
\newblock Treating a fuzzy subset as a projectable random subset.
\newblock In M.~Gupta and E.~Sanchez, editors, {\em Fuzzy Information and
  Decision Processes}, pages 213--220. North-Holland, Amsterdam, 1982.

\bibitem{prade1991possibilistic}
D.~D.-H. Prade.
\newblock Possibilistic logic, preferential models, non-monotonicity and
  related issues.
\newblock In {\em Proc. of IJCAI}, volume~91, pages 419--424, 1991.

\bibitem{rigollet2012sparse}
P.~Rigollet and A.~B. Tsybakov.
\newblock Sparse estimation by exponential weighting.
\newblock {\em Statistical Science}, pages 558--575, 2012.

\bibitem{sales1982fuzzy}
T.~Sales.
\newblock Fuzzy sets as set classes.
\newblock {\em Stochastica}, 6(3):249--264, 1982.

\bibitem{savage1954foundations}
L.~J. Savage.
\newblock The foundations of statistics.
\newblock {\em NY, John Wiley}, pages 188--190, 1954.

\bibitem{schapire1990strength}
R.~E. Schapire.
\newblock The strength of weak learnability.
\newblock {\em Machine learning}, 5(2):197--227, 1990.

\bibitem{serrurier2015entropy}
M.~Serrurier and H.~Prade.
\newblock Entropy evaluation based on confidence intervals of frequency
  estimates: Application to the learning of decision trees.
\newblock In {\em International Conference on Machine Learning}, pages
  1576--1584, 2015.

\bibitem{Sha76}
G.~Shafer.
\newblock {\em A Mathematical Theory of Evidence}.
\newblock Princeton University press, Princeton (NJ), USA, 1976.

\bibitem{shenoy1992using}
P.~P. Shenoy.
\newblock Using possibility theory in expert systems.
\newblock {\em Fuzzy Sets and Systems}, 52(2):129--142, 1992.

\bibitem{tulyakov2008review}
S.~Tulyakov, S.~Jaeger, V.~Govindaraju, and D.~Doermann.
\newblock Review of classifier combination methods.
\newblock In {\em Machine learning in document analysis and recognition}, pages
  361--386. Springer, 2008.

\bibitem{ward1963hierarchical}
J.~H. Ward~Jr.
\newblock Hierarchical grouping to optimize an objective function.
\newblock {\em Journal of the American statistical association},
  58(301):236--244, 1963.

\bibitem{wolpert1992stacked}
D.~H. Wolpert.
\newblock Stacked generalization.
\newblock {\em Neural networks}, 5(2):241--259, 1992.

\bibitem{woz14}
M.~Wozniak, M.~Grana, and E.~Corchado.
\newblock A survey of multiple classifier systems as hybrid systems.
\newblock {\em Information Fusion}, 16:3 -- 17, 2014.
\newblock Special Issue on Information Fusion in Hybrid Intelligent Fusion
  Systems.

\bibitem{yao2019federated}
X.~Yao, T.~Huang, R.-X. Zhang, R.~Li, and L.~Sun.
\newblock Federated learning with unbiased gradient aggregation and
  controllable meta updating.
\newblock In {\em NeuRIPS Workshop on Federated Learning}, page to appear,
  2019.

\bibitem{Zad78}
L.~A. Zadeh.
\newblock Fuzzy sets as a basis for a theory of possibility.
\newblock {\em Fuzzy Sets and Systems}, 1:3--28, 1978.

\end{thebibliography}

\end{document}